\documentclass[11pt]{article}
\usepackage[final]{acl}

\usepackage{times}
\usepackage{latexsym}
\usepackage[T1]{fontenc}
\usepackage[utf8]{inputenc}
\usepackage{microtype}
\usepackage{inconsolata}
\usepackage{graphicx}
\usepackage{booktabs}
\usepackage{array}
\usepackage{calc}
\usepackage{amsmath,amssymb}
\usepackage{xcolor}

\graphicspath{{figures/}}

\providecommand{\tightlist}{%
  \setlength{\itemsep}{0pt}\setlength{\parskip}{0pt}}

\DeclareUnicodeCharacter{00A7}{\S}
\DeclareUnicodeCharacter{00B1}{\ensuremath{\pm}}
\DeclareUnicodeCharacter{00B2}{\ensuremath{{}^{2}}}
\DeclareUnicodeCharacter{00B7}{\ensuremath{\cdot}}
\DeclareUnicodeCharacter{00D7}{\ensuremath{\times}}
\DeclareUnicodeCharacter{0394}{\ensuremath{\Delta}}
\DeclareUnicodeCharacter{03C1}{\ensuremath{\rho}}
\DeclareUnicodeCharacter{03B1}{\ensuremath{\alpha}}
\DeclareUnicodeCharacter{03B4}{\ensuremath{\delta}}
\DeclareUnicodeCharacter{03B8}{\ensuremath{\theta}}
\DeclareUnicodeCharacter{03C4}{\ensuremath{\tau}}
\DeclareUnicodeCharacter{2013}{--}
\DeclareUnicodeCharacter{2014}{---}
\DeclareUnicodeCharacter{2026}{\ldots}
\DeclareUnicodeCharacter{2190}{\ensuremath{\leftarrow}}
\DeclareUnicodeCharacter{2192}{\ensuremath{\rightarrow}}
\DeclareUnicodeCharacter{21D2}{\ensuremath{\Rightarrow}}
\DeclareUnicodeCharacter{2212}{\ensuremath{-}}
\DeclareUnicodeCharacter{2216}{\ensuremath{\setminus}}
\DeclareUnicodeCharacter{221E}{\ensuremath{\infty}}
\DeclareUnicodeCharacter{2248}{\ensuremath{\approx}}
\DeclareUnicodeCharacter{2264}{\ensuremath{\leq}}
\DeclareUnicodeCharacter{2265}{\ensuremath{\geq}}
\DeclareUnicodeCharacter{2284}{\ensuremath{\not\subset}}
\DeclareUnicodeCharacter{2070}{\ensuremath{{}^{0}}}
\DeclareUnicodeCharacter{2074}{\ensuremath{{}^{4}}}
\DeclareUnicodeCharacter{2075}{\ensuremath{{}^{5}}}
\DeclareUnicodeCharacter{2076}{\ensuremath{{}^{6}}}
\DeclareUnicodeCharacter{2077}{\ensuremath{{}^{7}}}
\DeclareUnicodeCharacter{207B}{\ensuremath{{}^{-}}}

\ifdefined
\fi

\title{Routing Is Least Learnable Where It Is Most Valuable:\\
  Bounds on Representation Routing for Web Agents}

\author{Jiaming Wei \\ University College London / Holistic AI \\ \texttt{jiaming.wei.25@ucl.ac.uk} \And
  Zekun Wu \\ Holistic AI / University College London \\ \texttt{zekun.wu.19@ucl.ac.uk} \AND
  Adriano Koshiyama \\ Holistic AI \\ \texttt{adriano.koshiyama@holisticai.com} \And
  Maria Perez-Ortiz \\ UCL Centre for Artificial Intelligence \\ \texttt{maria.perez@ucl.ac.uk}}

\begin{document}
\maketitle

\begin{abstract}
Web agents observe a browser through text, pixels, or both, and the choice is usually fixed once for all tasks. We measure six observation modes across eight site--model combinations (cells) on VisualWebArena and WebArena and ask what choosing per task would buy. The modes are complementary: each solves tasks the others miss, they fail in structurally different ways, and the best choice reverses between task sets. The obvious prize, an oracle that picks a winning mode for every task, looks large but is inflated by run-to-run noise: rerunning the same mode on the same tasks changes 12--14\% of outcomes, so a second run of a mode already in hand gains about as much as adding a new one. What survives is a cost bound: sending only the tasks no mode solves to the cheapest mode cuts cost by 9.5--30.6\% in 8 of 8 cells at unchanged success. We then test five routing policies (picking the mode, deciding when to spend on the strong mode, a zero-cost rule read off the task text, a confidence cascade, and pooled cost tiers), and none robustly beats simply fixing one well-chosen mode; the one exception is a fragile result in our sparsest cell. The central obstruction is that routing supervision is produced at the agent's success rate: the weaker the agent, the fewer labels a router gets, exactly where routing would be most valuable. This limit belongs to today's agents rather than to routing itself. Label supply and routing opportunity rise together (correlation 0.95 across cells), so a stronger agent can overturn the result, and we report the rerun noise bands and the full measurement protocol.

\end{abstract}

\section{Introduction}\label{sec:intro}

\begin{figure*}[t]
\centering
\includegraphics[width=0.90\textwidth]{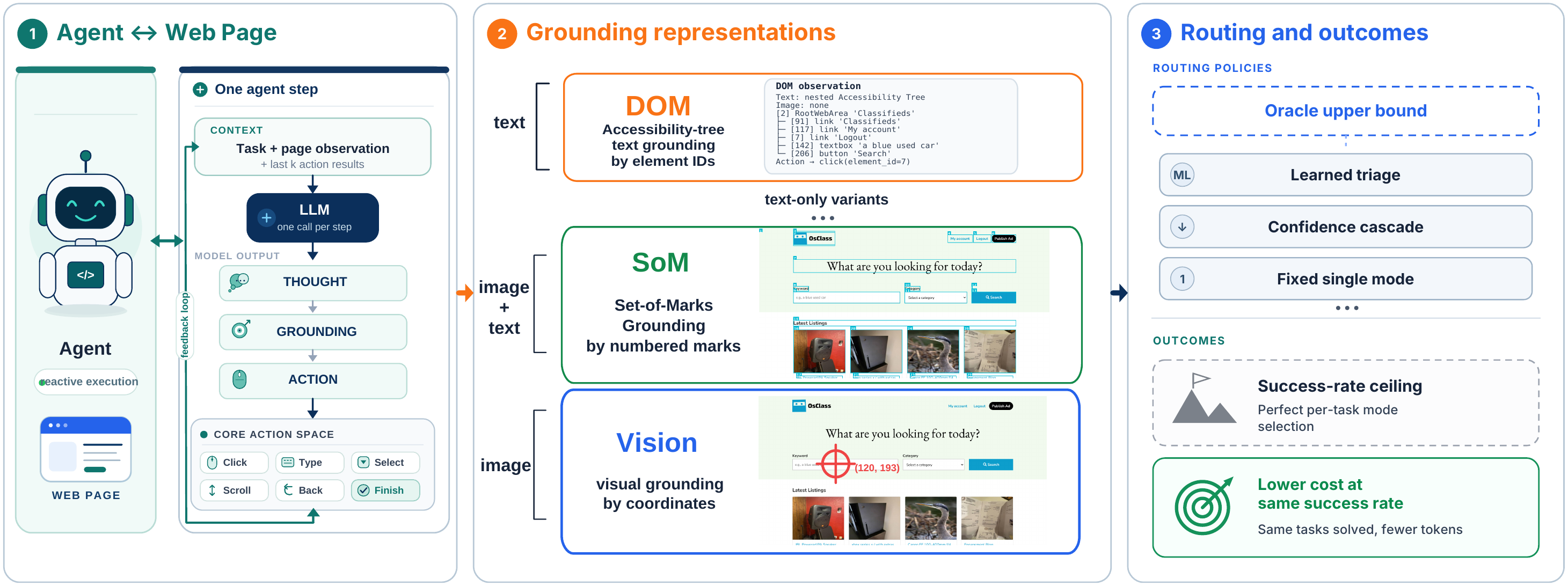}
\caption{What this paper measures. \textbf{(1)} One model call per step emits a thought, a grounding and an action, with the last eight action outcomes as context. \textbf{(2)} That step is grounded by element id, by numbered mark, or by coordinate; screenshot-free variants separate the text payload from the prompt family (\S\ref{sec:setup}). \textbf{(3)} Three of our five routing policies, bracketed by an outcome oracle above and a fixed single mode below (the pooled-tier router and the zero-token rule are omitted for space; \S\ref{sec:lowerbound}), and the two ceilings a perfect per-task choice could reach. Only the cost ceiling survives the rerun control of \S\ref{sec:noise}, because it adds no arm. \textbf{Caution:} the policy marked as exceeding a fixed single mode does so on a lossless-saving criterion under a label-shuffle null, and the one setting surviving correction is \texttt{red$\cdot$B2}, whose triage signal is below chance (AUROC 0.483; Appendix Table~\ref{tab:pb05}). Under the stricter test no cell Pareto-dominates always-cheapest (Appendix Table~\ref{tab:pb06}).}
\label{fig:overview}
\end{figure*}

A web agent has to be shown the page before it can act on it. In practice it is shown one of three things: the accessibility tree as text, a screenshot as pixels, or a set-of-marks fusion in which the screenshot is annotated with the tree's element ids \citep{yang2023som}. Which of these an agent receives is the grounding decision that \citet{zheng2024seeact} identify as the bottleneck for a capable backbone, and that a line of work on visual grounding for GUI agents attacks directly \citep{zheng2024uground}. Benchmarks such as VisualWebArena \citep{koh2024visualwebarena}, WebArena \citep{zhou2024webarena} and Mind2Web \citep{deng2023mind2web} report agents that make this choice once, at build time, and hold it for every task. The choice is treated as a design decision about the system rather than as a decision about the task in front of it.

\looseness=-1 It need not be. We hold one scaffold, one prompt budget and one action space fixed, and compare the six representations across eight (site $\times$ backbone) cells. They are not redundant: 44 of the 48 mode--cell pairs solve at least one task no other mode in that cell solved, and every mode does so in some cell. Where exactly one channel succeeds, the losing channel fails in a structurally different way. \emph{Which} representation wins also reverses between our two task sets, so the winner is a property of the deployment and not of the modality (\S\ref{sec:complementarity}). A per-task choice therefore has something to choose between, which raises the obvious next question: what would it be worth?

\looseness=-1 That question is easy to answer badly. An outcome oracle over six modes (it sees every outcome and picks a winning mode per task) reaches 7.1--51.9\% against a best single mode of 2.2--35.6\%, which reads as enormous. It is also a six-arm union quoted against one arm: a task counts as solved there if \emph{any} of six runs solved it, and an \emph{arm} is one mode run in one cell. The comparison it needs is a rerun. Rerunning one condition changes 12--14\% of task outcomes, and 22\% of tasks flip in at least one of three replicated arms; at that rate a second copy of an arm already in hand raises the union ceiling by as much as a different representation does. The union survives as a measured bound; what does not is the attribution of its headroom to representation diversity. What holds up without adding an arm is narrower: routing only the never-solved tasks to the cheapest mode cuts cost 9.5--30.6\% in every cell at unchanged success (\S\ref{sec:upperbound}).

Routing itself is not new. Systems route a query to one of several \emph{models} to buy accuracy at lower cost \citep{chen2023frugalgpt,ong2025routellm}, including for web and computer-use agents specifically \citep{webrouter2025,liu2026adaptive}; \citet{moslem2026routingsurvey} survey the area. What those route over is a capability gradient: a larger model really is more accurate, so difficulty predicts which arm to use. We ask whether the same holds when the arms are representations of one browser state under one model, where no such gradient exists by construction.

The lower bound is where the paper turns. We build five routing policies: choosing which mode to use, a learned triage of spend, a zero-token rule read off the task text, a confidence cascade, and pooled cost tiers. None yields a robust improvement over simply fixing the best or cheapest arm, the one exception being descriptive rather than confirmatory (\S\ref{sec:lowerbound}). That would be uninformative if they all failed the same way, and they do not. One dies of label supply. One has ample labels and buys nothing. One has a perfect free signal and still loses, because the expensive arm does not hurt on the tasks the signal excludes (\S\ref{sec:lowerbound}).

The gap between those bounds is what this paper can offer without overstating it. What it comes down to is a circularity: \textbf{routing supervision is produced at the success rate, so routing is least learnable exactly where it would be most valuable}. Those look like two obstructions and are one: the labels a router can learn from and the tasks where routing can pay both track the agent's success rate, which indexes the negative result to the current capability regime and says what would overturn it (\S\ref{sec:gap}).

Underneath all of it, every comparison is read against a measured same-condition rerun band rather than against zero (\S\ref{sec:noise}); \citet{hajimiri2026budgetmatched} make the same move on a different axis, showing that a token-matched vanilla baseline erases the reported gains of several agent memory and skill modules. Every quantity is read from its producing artefact at render time (\S\ref{sec:threats}). On workloads at these success rates the resolution of the instrument is comparable to the effects being reported, and a paper that does not say so is reporting its instrument.

\paragraph{Contributions.} (i) The six observation representations are genuinely complementary: each solves tasks the others do not, their failures differ structurally, and which one wins is a property of the deployment (\S\ref{sec:complementarity}). (ii) Most of the apparent oracle headroom comes from rerun variance rather than from representation diversity; the bound that survives the rerun control is the cost ceiling, 9.5--30.6\% at unchanged success (\S\ref{sec:upperbound}). (iii) The benchmarks the field measures on cannot produce routing supervision: labels arrive at the success rate, so five routing constructions all land at or below trivial fixed policies, and the obstruction is the supervision, not the estimator (\S\ref{sec:lowerbound}--\S\ref{sec:gap}).

\section{Setup}\label{sec:setup}

We evaluate on VisualWebArena \citep{koh2024visualwebarena} and WebArena \citep{zhou2024webarena}.

Throughout, a \emph{cell} is one (site $\times$ backbone) pair, the unit everything is measured in: sites \texttt{cls}/\texttt{red} are VisualWebArena classifieds/reddit and \texttt{WA} is WebArena reddit; backbones are B0 = Qwen3-VL-235B (API-served), B1 = Qwen3-VL-4B and B2 = Gemma-3-4B (both local). Eight cells exist; \texttt{WA$\cdot$B2} was never run. A \emph{condition} is one mode run in one cell (36 in total), called an \emph{arm} when it is an option a router could pick; \emph{mode} and \emph{representation} are used interchangeably.

Every number in this paper is read from its producing artefact at render time; none is typed by hand (\S\ref{sec:threats}).

Each table's multiplicity status (confirmatory / descriptive / selection-derived) is stated in the appendix preamble.

\paragraph{A deliberately plain agent.} One model call per step emits one JSON object: \texttt{thought} (free text; the backbones run in non-thinking mode, so this is the only reasoning the system produces), a self-reported \texttt{confidence}, an \texttt{action\_type} and its arguments. The prompt is the instruction, the mode's system prompt, the last eight steps and the current observation; there is no planner, memory, reflection or retry policy above the step. The history records \emph{outcomes} rather than reasoning: action, success, whether the page changed, resulting URL. Holding this fixed is what makes a difference between conditions a difference of representation and not of scaffold.

\paragraph{Six observation modes: a $2\times2$ plus two corners.} Four modes carry no screenshot and differ only in which \emph{text payload} the agent receives and which \emph{prompt family} tells it what to expect (Figure~\ref{fig:overview}, panel 2). The payload is either the full accessibility tree or the flattened \texttt{[SOM\_MARKS]} list that annotates a set-of-marks screenshot; crossing the two gives DOM, DOM+sprompt (tree under the SoM prompt), DOM+stext (marks under the DOM prompt) and SoM-image (marks under the SoM prompt and, despite the name, no screenshot). The corners add or remove the image: SoM is SoM-image plus the annotated screenshot, Vision is the raw screenshot with an \emph{empty} payload. The payload also fixes the element-id regime and therefore which modes inherit id churn (\S\ref{sec:noise}): \texttt{[SOM\_MARKS]} is keyed $1\ldots K$ by position, the tree keeps native browser node ids. A Vision episode whose screenshot failed to capture aborts rather than degrading to text, so its ``screenshot only'' contract is auditable rather than assumed. The full mode grid, with per-mode payload and prompt detail, is Appendix~\ref{supporting-tables}.

\paragraph{Deployment classes.} We group the six into the three shapes web agents are deployed in, \emph{no-image} (DOM and the three screenshot-free variants), \emph{hybrid} (SoM) and \emph{vision-only} (Vision), then compare at one arm per class, since a maximum over the four no-image arms against one is not a comparison. The grouping is a deployment convention rather than a claim the data licenses: the behavioural evidence separates Vision and SoM from the rest but cannot pool the four text modes (Appendix Table~\ref{tab:t05}). Appendix~\ref{app:classes} has the full comparison and the ablation (Table~\ref{tab:t04}).

\paragraph{Task universes.} Rates are formed over a \emph{scored} set of 224 classifieds tasks, 203 reddit, and the 104 WebArena tasks all six modes ran. That set is what we collected minus protocol exclusions. We wrote rules about task configuration rather than picking tasks, so how many each rule removes is its output, not our choice. Classifieds loses none; reddit loses two of 205. One is excluded on its configuration alone: its evaluator only asserts that three subreddits are \emph{absent} from the sidebar and never checks the subscription the task asks for, so doing nothing scores~1. The other is a cross-site task the model can answer from memory: it accounts for 9 of the 11 cross-site successes on reddit, and none of those 9 ever loaded the Wikipedia host the task points to. Every artefact records which task set it was built on, so an analysis using the older one fails a check rather than silently mixing denominators. Both exclusions are reported with and without.

\paragraph{Binary scoring.} Over 7{,}686 scored episodes in 36 conditions the evaluator emits exactly two values, none missing and none non-numeric (Appendix Table~\ref{tab:t35}). There is no graded target: a partly completed task is indistinguishable from one never begun. That is a property of the benchmarks rather than of this pipeline, and a precondition of every routing negative below.

\paragraph{Efficiency estimands.} An \emph{estimand} is what a number is actually measuring, as opposed to what it is called. Cost here is per-episode billed cost, but B0's per-token API bill and B1/B2's electricity estimate are not the same kind of quantity, so they can be ordered and not divided. Within a cell, mode-to-mode ratios can therefore be read as ratios; across deployment classes only the ordering carries; the two are never pooled. Switching the denominator from per-attempt to per-success changes which mode is cheapest in four of the six cells with enough successes to divide by. A latency figure is mostly environment: the model call is 22--67\% of a step. No bare cross-site latency number appears in this paper (Appendix~\ref{app:efficiency}, Tables~\ref{tab:t11} and \ref{tab:t15}; the per-mode efficiency matrix is Table~\ref{tab:t09}).

\section{Different representations fail differently}\label{sec:complementarity}

\looseness=-1 \paragraph{Unique solves.} The case that routing could be worth anything starts here: the modes are not redundant. 44 of the 48 mode--cell pairs solve at least one task no other mode in that cell solved, and every mode does so in some cell (the \texttt{unique} column of Table~\ref{tab:t06}). On the tasks where exactly one channel (the text-only or the image-bearing half of the modes) succeeds, the losing channel's failures are not generic: the paired attribution (Table~\ref{tab:t41}) and the vocabulary-free probes that confirm the residual is not a ruleset artefact (Table~\ref{tab:t26}) are in Appendix~\ref{app:failures}, with the text/prompt axis decomposition and the hallucinated-reference counts (Tables~\ref{tab:t30}, \ref{tab:t32}). Part of the difference is structural rather than behavioural: Vision is on the coordinate dispatch path by construction, and that path delivers a click far less often than the element-id path (Table~\ref{tab:t13}, Appendix~\ref{app:audits}).

\looseness=-1 \paragraph{Divergence at the first step.} The difference is present from the start: on the tasks whose outcome two modes disagree about, 72--100\% of VisualWebArena trajectories have already diverged by step 3 and at most 3\% diverge at step 10 or later (Appendix Table~\ref{tab:t43}). Accumulated drift would share a prefix and part late, so what the representation changes is the first decision rather than a compounding error. The effect is markedly weaker on WebArena (50--90\% early, up to 22\% late), so the two sites are quoted separately and never pooled. The shape holds on both, though only the VisualWebArena cells make it categorical.

\looseness=-1 \paragraph{Behavioural separation.} The modes differ in how they act, not only in what they solve, and the difference clears the noise floor of \S\ref{sec:noise}: on the replicated cell, 22 of 25 live behavioural metrics separate the modes by more than a rerun of a single arm moves them, several by 5--22$\times$ (Appendix Table~\ref{tab:t14}). The signatures are legible rather than statistical: on \texttt{red$\cdot$B2}, Vision scrolls 8$\times$ as often as SoM and clicks half as often (Appendix~\ref{app:matrices}, Tables~\ref{tab:t07}--\ref{tab:t08}). And the divergence is not only in the macro action mix: per-step decision quality moves more than macro action frequency in 7 of 8 cells, at 1.34--4.07$\times$ on the six VisualWebArena cells with the single exception \texttt{WA$\cdot$B0} at 0.97$\times$ (Appendix Table~\ref{tab:t31}), so what the representation changes is which action is chosen, not merely how often each type appears. The three that do not separate are the two latency metrics (0.87$\times$, 0.84$\times$, which the latency decomposition reaches independently) and parse-fail rate, near zero under every mode.

\begin{figure}[tbp]
\centering
\includegraphics[width=\columnwidth]{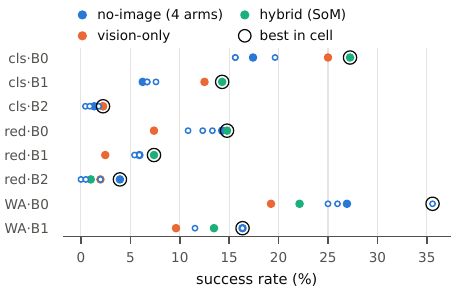}
\caption{Success rate of all six modes in all eight cells (Appendix Table~\ref{tab:t01}), coloured by deployment class; rings mark the best mode: hybrid or vision-bearing on classifieds, no-image on WebArena.}
\label{fig:sr}
\end{figure}

\looseness=-1 \paragraph{No portable winner.} Which representation wins is not portable either. Figure~\ref{fig:sr} plots the raw success rate of every mode in every cell (Appendix Table~\ref{tab:t01}); Appendix Table~\ref{tab:t03} groups the modes into the three deployment classes of \S\ref{sec:setup}, at one arm per class. \emph{hybrid} is the best class in four cells and \emph{no-image} in three, with one tie, and which class wins reverses between task sets: hybrid leads on classifieds by 8.04--9.82pp while no-image leads on WebArena by +4.81pp at one arm per class. Those three gaps clear the rerun threshold of \S\ref{sec:noise}; the remaining five cells sit at 0.49--2.96pp, within noise, and are read as directions.

\section{The noise floor}\label{sec:noise}

Appendix Table~\ref{tab:t27} is the supervision problem underneath every routing analysis in this paper: the rows a router would learn from are the contested ones, and those are exactly the rows that flip between two runs of the same condition. On \texttt{cls$\cdot$B0}, rerunning one condition changes the outcome of 12.1--14.3\% of tasks, and 49 of 224 tasks flip in at least one of the three replicated arms. Those flips are not spread evenly. They concentrate in the rows where more than one mode succeeds: 48--52\% of those flip, against 2.9\% of the rest.

\paragraph{What was replicated.} (full tables: Appendix~\ref{app:noise}) Three arms of \texttt{cls$\cdot$B0} (DOM, Vision, SoM) at $n=224$, and five modes of \texttt{wa\_red$\cdot$B1} on a registered ten-task draw. Nothing else: no VWA-reddit cell and no B2 cell carries a replicate, so every band below is imported into those cells rather than measured in them. On the three full arms, re-running one condition moves the set of solved tasks by 4.91--7.59pp in each direction (12.1--14.3\% of tasks change outcome); on the local backbone the same quantity is 2.00--4.00pp (6.0\% of tasks).

\looseness=-1 \paragraph{Band versus threshold.} Two quantities could each be called a noise floor, and only one of them is one. The \emph{observed band} is how much mean success rate actually moved across those pairs: 0.89--2.23pp. Each of those is \emph{one draw} of a random quantity rather than a bound on it, so reading a 2.2pp effect against a 2.23pp ``measured floor'' compares a draw to a draw. The \emph{resolution threshold} asks a different question: how large must an effect be before a single rerun would be unlikely to produce it unaided? Under the exchangeability null in which each of the $d$ discordant tasks falls either way with probability one half, $\mathrm{SD}(\Delta\mathrm{SR}) = \sqrt{d}/n$ over the $n$ scored tasks, giving 2.32--2.53pp, the same size as the observed band; an effect has to reach roughly 3.8--4.2pp to clear it. We use the threshold throughout, and quote the observed band only as what repetition happened to deliver. Neither is a significance test. A rerun band states the resolution of this instrument, meaning how small a difference it cannot separate. Where this paper needs an inferential claim it uses a paired interval on the contrast itself (Appendix Tables~\ref{tab:t16} and \ref{tab:t17}). The null behind the threshold models only exchangeability of the two runs, not environment drift; and site state accumulates rather than resets, so drift pushes in one direction and adds to the movement the null already allows. So 3.8--4.2pp is itself a lower bound on what repetition could deliver.

\paragraph{Mean differences versus set differences.} Two runs of one condition also differ as \emph{sets}: $|\{a\text{ solves}\} \setminus \{b\text{ solves}\}|$ runs 4.91--7.59pp on the three full arms. A union ceiling moves by that functional, so an added arm is judged against it in \S\ref{sec:upperbound} rather than against the threshold above. The two must never be compared to each other. A mean-difference threshold applied to a set-difference gain would be arithmetic across estimands, which is how the same replicate can appear to license opposite conclusions.

\looseness=-1 \paragraph{Sources of run-to-run movement.} Sampling is not the answer: every condition in this paper decodes at temperature $0$ with a fixed seed. Inference is nonetheless not bit-reproducible in general \citep{he2025nondeterminism,yuan2025numerical}, and language models are separately sensitive to formatting choices that carry no information \citep{sclar2024promptformat}. Two sources survive a source-level audit of the pipeline. The first is element-id churn. The accessibility-tree payload is keyed by native browser node ids that are reassigned per snapshot; the model is sensitive to those id tokens; and a replay that shuffles ids while holding the page fixed changes which element is chosen on 20.0\% of B1 steps and 12.5\% of B0 steps. Because churn is a property of the text payload and not of the prompt, it reaches exactly the two modes carrying an accessibility tree, DOM and DOM+sprompt. The modes carrying \texttt{[SOM\_MARKS]} are keyed $1\ldots K$ by position and are unaffected. The second applies only to B0. A hosted mixture-of-experts endpoint routes and batches server-side, and no client-side setting controls that. B1 and B2 are served locally and decode greedily, and their steps replay bit-identically (133/133 steps). Their \emph{episodes} still move, because an episode also depends on site state, wall-clock and session lifetime. What every band here names is therefore run-to-run movement including environment drift, never decoding stochasticity. We report no split of the total across these sources. An earlier linear decomposition was withdrawn as a category error, and the per-channel figures above are what each channel moves on its own rather than shares of one budget.

\paragraph{Consequences for the rest of the paper.} Every comparison after it is read against these bands, whether between modes, between deployment classes, or between a representation and a rerun. An arm added for its representation therefore has to be priced against an arm added by repetition (\S\ref{sec:upperbound}), and the tasks a router would be trained on are the least stable rows in the dataset (\S\ref{sec:lowerbound}).

\section{A rerun-corrected upper bound}\label{sec:upperbound}

\looseness=-1 Figure~\ref{fig:ceilings} shows the two ceilings a perfect per-task choice could reach (per-cell values in Appendix Table~\ref{tab:t42}), and only one survives its own control. The \emph{success-rate ceiling} (any mode solves it) runs 7.1--51.9\% against a best single mode of 2.2--35.6\%; but it is a six-arm union quoted against a one-arm baseline, and the arm-matched columns in the same table show that adding the best distinct arm buys +1.97 to +7.14pp while re-running an arm already in hand buys 4.91--7.59pp wherever a replicate exists. The two are directly comparable: the same quantity, computed at the same arm count in the same cell. The \emph{cost ceiling at matched success} adds no arm and is untouched by that objection: keeping the best mode everywhere and sending only never-solved tasks to the cheapest arm leaves success unchanged by construction and cuts cost 9.5--30.6\% in 8 of 8 cells. Its paper-B precursors are Appendix Tables~\ref{tab:pb01}--\ref{tab:pb02}.

The same rerun correction applies to fusion as a representation purchase: Appendix Table~\ref{tab:t18} puts a distinct arm and a rerun side by side at the same arm count, and against the workload-matched single channel the fused mode's premium clears the rerun threshold of \S\ref{sec:noise} in 0 of 8 cells (Appendix Figure~\ref{fig:fusion} and Table~\ref{tab:t16}).

\looseness=-1 The number worth aiming at is therefore not the six-arm union. It is the gap between the fixed policies of \S\ref{sec:lowerbound} and the \emph{cost} ceiling: 9.5--30.6\% in 8 of 8 cells at unchanged success, reachable in principle with one bit per task rather than a mode identity. That target survives the rerun objection for the reason the success-rate ceiling does not: it adds no arm, being the same arms spent differently.

\begin{figure}[tbp]
\centering
\includegraphics[width=\columnwidth]{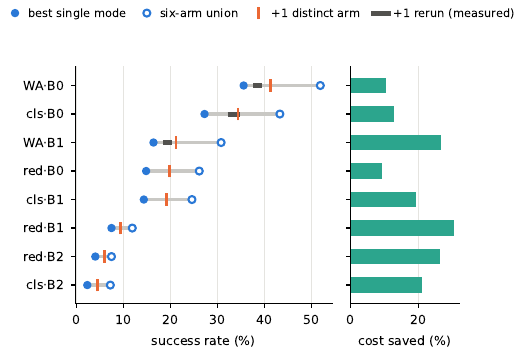}
\caption{The two ceilings, per cell (full table: Appendix Table~\ref{tab:t42}). Left: the six-run union ceiling against the best single mode; the tick is what one genuinely different arm adds, and the dark segment is what one rerun of an arm already in hand adds, where measured. Most of the apparent headroom is not attributable to representation diversity. Right: the cost ceiling at matched success, the bound that survives, because it adds no arm.}
\label{fig:ceilings}
\end{figure}

\section{The lower bound: five routing constructions, five obstructions}\label{sec:lowerbound}

No implementable policy we could construct yields a robust improvement over the trivial fixed policies (per-policy tables in Appendix~\ref{app:routing}). One exception is descriptive rather than confirmatory: on \texttt{red$\cdot$B2}, our sparsest cell, the zero-token rule exceeds always-DOM on success, cost and latency at once (Appendix Table~\ref{tab:t19}), but that cell is the one whose verdict the leakage correction of \S\ref{sec:threats} flips, and its triage signal is below chance. The obstructions differ, which is what makes the bound informative rather than an artefact of one weak router.

\textbf{Choosing which mode dies of label supply.} The natural label, the cheapest mode that solved the task, exists only where something succeeded, so it is produced at the success rate: 15--97 labels per cell over six classes (Appendix Table~\ref{tab:pb03}), and under a minimum-of-ten-rows-per-class filter four of six cells admit no classifier at all (Appendix Table~\ref{tab:pb04}).

The scarcity is worse than a label count suggests, because the \emph{question} is scarce too: the tasks where more than one mode succeeds, the only rows on which a per-task choice even exists, are 1.5--34.6\% of a cell, and on two cells they number 3 and 4 (Appendix Table~\ref{tab:t42}).

\textbf{Choosing whether to spend has the labels and still buys nothing.} The triage label is defined everywhere and predictable in five of the six cells (AUROC 0.651--0.717; the sixth, \texttt{red$\cdot$B2}, is below chance at 0.483; Appendix Table~\ref{tab:pb05}), yet under fully nested cross-validation no cell's learned triage Pareto-dominates always taking the cheapest mode (Appendix Table~\ref{tab:pb06}).

\textbf{A free signal read straight off the task text exists, and routing on it still loses.} A regex over the task intent flags tasks where the screenshot is worth +22.54pp against +0.65pp elsewhere (Figure~\ref{fig:partition}; per-cell intervals in Appendix Table~\ref{tab:t17}). The policy built on it still loses to always-Vision, because the screenshot does not hurt on the unflagged tasks either (Appendix Table~\ref{tab:t19}).

\textbf{The cascade has a usable ranking and still buys nothing.} Running the cheap mode first and escalating when the model's self-reported confidence is low, in the manner of \citet{gupta2024cascades} on the signal of \citet{kadavath2022know}, beats always-rich at no operating point in any comparable cell --- although against a random-escalation control that ranking is informative (Appendix Tables~\ref{tab:t20}--\ref{tab:t21}).

\textbf{Pooling backbones restores supply but makes labels ambiguous}, because different backbones disagree about the same task (Table~\ref{tab:t33}; the one surviving cell is taken apart in Appendix~\ref{the-reddit-b2-saving-in-detail}). Two further features fail before they start: the benchmark's own difficulty annotation adds $\Delta$AUROC +0.0024 (Table~\ref{tab:t22}), and whether the task supplies a reference image, the feature a practitioner would reach for first, looks like it should help and does not (Table~\ref{tab:t23}).

\looseness=-1 The pattern is not `no signal'. It is that the arm the router would route to is already the right arm to route everything to.

\begin{figure}[tbp]
\centering
\includegraphics[width=\columnwidth]{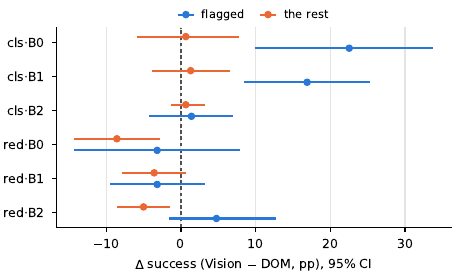}
\caption{The zero-token partition (Appendix Table~\ref{tab:t17}): the screenshot's worth (Vision $-$ DOM) on regex-flagged tasks versus the rest. It is large and significant only on the capable classifieds backbones, and null on reddit. The predicate flags 71 of 224 classifieds tasks and 63 of 203 reddit tasks.}
\label{fig:partition}
\end{figure}

\section{The gap, and what would close it}\label{sec:gap}

\looseness=-1 Between the rerun-corrected upper bound of \S\ref{sec:upperbound} and the lower bound of \S\ref{sec:lowerbound} sits a gap no method in this paper crosses, and the closed routes fail for three separate reasons, supply, estimand and value, rather than one obstruction a better router might get around (Appendix Table~\ref{tab:pb07}; the route-by-route derivations are Appendix~\ref{derivations-for-the-four-relabelling-routes}). The sharpest statement of the bind: routing supervision is produced at the success rate, so routing is least learnable exactly where it would be most valuable; and the rows a router must learn from are the contested ones, which are the rows that flip between identical reruns (Appendix Table~\ref{tab:t27}).

What would change the answer, in order of leverage:

\begin{itemize}
\tightlist
\item
  \textbf{Graded evaluators.} The benchmark emits two values (Appendix Table~\ref{tab:t35}); a graded per-task score would make every episode a training signal regardless of success, attacking the supply obstruction directly.
\item
  \textbf{Replicates as a reporting norm.} Single-run reporting remains the norm in this literature; at the flip rates we measure, a mode-to-mode difference below roughly 4pp is not resolvable by one run against another (\S\ref{sec:noise}). Rerun floors should be reported next to mode deltas.
\item
  \textbf{A third workload, and cross-family coverage on the second benchmark}, so the moderator of the winner-reversal (\S\ref{sec:complementarity}) becomes identifiable.
\item
  \textbf{Online cascade infrastructure.} Every escalation number here is an offline splice; what a real cascade does after the cheap arm has acted on a stateful site is unobserved in this project.
\end{itemize}

\looseness=-1 \paragraph{Falsifiability.} It would be easy to read the obstructions above as separate walls: too few labels, and too few tasks where a choice even exists. They are one wall. The rows a which-mode router can be trained on exist only where something succeeded, and the rows where a per-task choice exists at all are those with more than one solver. Both are subsets of the tasks the agent can do, and across our eight cells they track the best single mode's success rate at $\rho = 0.952$, mean gap 1.65pp (Appendix Table~\ref{tab:t44}). Part of that coupling is structural, both being functionals of one solve matrix. What is not structural is that the routable share grows roughly \emph{linearly} in the per-mode rate, where mutually independent modes would make it grow near-quadratically at these success levels, since two of them would then both solve a task at roughly the square of the per-mode rate. The negative result here is therefore a statement about routing at 2--36\% success rather than about routing, and it predicts its own reversal: run this measurement on an agent that solves most of these tasks and the label supply and the contested set grow together. We would rather be refuted that way than cited as evidence that representation routing does not work. Until then the number worth aiming at is the cost ceiling of \S\ref{sec:upperbound}: 9.5--30.6\% in every cell, reachable with one bit per task rather than a mode identity, and reached by none of the five policies here.

\section{Threats to validity}\label{sec:threats}

\looseness=-1 \paragraph{What the benchmark hands us.} The evaluator emits two values, so there is no graded target and every routing negative inherits that (Table~\ref{tab:t35}); rule-based trajectory scoring is known to disagree with human judgement often enough to matter \citep{lu2025agentrewardbench}. WebArena has since been re-audited task by task \citep{elhattami2025webarenaverified} and the field's reported progress reassessed on live sites \citep{xue2025illusion}, but the VisualWebArena audit below is our own.

\looseness=-1 \paragraph{What the harness adds.} An arm's measured ceiling is partly this harness. Vision is on the coordinate path by construction, and that path succeeds far less often than the element-id one (Table~\ref{tab:t13}). Reddit episodes leave the benchmark for the public internet on 1.05--2.13\% of steps against 0.16\% on classifieds, and reddit's container is slower than classifieds' before any agent behaviour enters (Table~\ref{tab:t29}); Table~\ref{tab:t34} corrects a page-change false positive.

\paragraph{What we did ourselves.} Six successes were credited by accumulated site state, and zeroing them flips one verdict (Appendix Table~\ref{tab:t28}; the full earned/leaked audit, now including WebArena at 0 leaked of 37 audited episodes, a lower bound, is Table~\ref{tab:t40}).

\paragraph{Provenance.} \looseness=-1 The threat we can report on rather than only declare is our own. An audit of the analysis code on 2026-08-03 found five conclusions hardcoded in the producers that generate the tables stating the same quantity, in four scripts. Four were stale denominators: an annotation column reading $6/6$ beside a count column reading $8/8$, and an interval described as no longer excluding zero while its own table showed that it did. The fifth was wrong on the fact: the confidence-cascade producer asserted that Vision is the cheapest mode in every cell, and on \texttt{WA$\cdot$B0} it is DOM. The sentence was true when written and never re-derived as the data grew. The fix is mechanical rather than editorial: every number in this paper and its appendix is now read from a product JSON when the table is rendered, so a stale claim cannot survive a regeneration. Two further instances surfaced while writing this draft: a caption describing a replicate inventory smaller than the one on disk, and a table whose body silently emitted zero rows because the producer asked for keys the product no longer had. We report the count as a lower bound: the sweep that found the five matched one textual shape ($n/6$), and hardcoded mode names, ratios and directions have not been swept.

\section{Discussion}\label{sec:discussion}

A builder of web agents can take four things from this without adopting our conclusion.

\looseness=-1 \textbf{Measure your own cell.} Which representation wins is a property of the deployment, not of the modality: it reverses between our two task sets and it is not stable under rerun noise in the low-success cells. A number read off someone else's benchmark does not transfer to your site, and the eight cells here disagree with each other more than any of them disagrees with a plausible prior.

\looseness=-1 \textbf{Price a rerun before pricing a representation.} The cheapest way to raise a union ceiling is to run an arm you already have a second time. Until that is measured, a gain from adding a representation cannot be attributed to the representation. Our own headline ceiling survived as a measurement but not as an attribution; the cost ceiling survived as both, because it adds no arm.

\textbf{Treat a difference smaller than the rerun band as unresolved, not as small.} At the flip rate we observe, a two-point difference between two modes is what a single repetition delivers on its own. Reporting it as an effect is reporting the resolution of the instrument.

\looseness=-1 \textbf{State the denominator, and do not pool a bill with an energy estimate.} Per-attempt and per-success cost disagree about which mode is cheapest in most of our cells; an API bill and an electricity estimate are not one quantity; and a latency figure is mostly browser and container. Each silently reorders the modes, so an efficiency claim without its estimand is ambiguous rather than weak: the trouble is not that the effect is small, it is that one cannot tell what was measured.

\looseness=-1 None of this settles whether representation routing can be made to work. It bounds the question: what a perfect choice would buy, what the available supervision delivers, and how far apart the two sit on the workloads the field measures.

\label{content-end}

\section*{Limitations}
\begin{itemize}
\tightlist
\item
  \textbf{No third workload.} The planned shopping workload never launched, so every claim in this paper rests on classifieds and reddit task sets alone.
\item
  \textbf{No cross-family control on WebArena.} Both WA cells are Qwen; the cross-family control exists only on VWA. This data cannot hold ``across benchmarks'' and ``across families'' simultaneously.
\item
  \textbf{The two benchmarks share one application.} WA reddit \emph{is} the \texttt{vwa-reddit} container: same image, same port, same account. On the reddit axis ``two benchmarks'' means one application with two task sets.
\item
  \textbf{The cascade outcome is an offline splice.} No run observes what a real cascade does after the cheap arm has already acted on a stateful site.
\item
  \textbf{Energy is uncalibrated} (psutil at \textasciitilde66W on a device rated several times that) and the local per-token cost constant was derived for a different accelerator.
\item
  \textbf{The rerun band is measured in two cells and imported into six.} Three arms of \texttt{cls$\cdot$B0} and five modes of \texttt{wa\_red$\cdot$B1} carry a replicate; no VWA-reddit cell and no B2 cell carries one. Every ``clears the noise floor'' statement about those six assumes the floor transfers, and it is not known to be mode- or cell-invariant.
\item
  \textbf{The WebArena leakage zero is a lower bound, not a clean bill.} No leaked success was found in 37 audited episodes; that bounds the rate rather than establishing absence.
\item
  \textbf{Two workloads cannot identify what moderates the reversal.} Modality, task set and benchmark all change together at the VWA/WA boundary; with two levels no design separates them.
\item
  \textbf{The B2 cells are near the floor.} They carry 1--8 successes per arm and 3--4 routable tasks, so a ratio computed on them is one or two tasks wide. They are reported for coverage and should not carry a comparison.
\item
  \textbf{The lower bound covers five simple constructions.} Logistic triage, a regex rule, a confidence cascade, pooled tiers and which-mode selection; no LLM-based router, reinforcement learning, contextual bandit or online policy was tested. The supply obstruction is estimator-independent, but the value of richer hypothesis classes on this data is unmeasured.
\end{itemize}

\label{refs-start}
\bibliography{paper}
\makeatletter
\@ifundefined{acl@finalcopytrue}{\bibliographystyle{acl_natbib}}{}
\makeatother

\clearpage
\appendix

\paragraph{Multiplicity status.} The tables in this paper and its appendix are not one inferential family and are not corrected as one. They divide into three classes: confirmatory (hypotheses fixed before the data and corrected within their own family, currently the 2$\times$2 axis-independence set of 64 conjunction hypotheses under BH/Holm \citep{holm1979sequentially} on $\max(p_1, p_2)$); descriptive (intervals, no test, no correction claimed: the full behavioural matrices, the efficiency tables, the failure-attribution counts); and selection-derived (a maximum over layers, signals, thresholds or comparators is part of the statistic, so these bound what a selection could deliver and cannot be read as effect estimates \citep{cawley2010overfitting}; each such table says so in its caption). No family-wise statement covers the set as a whole, and none is claimed.

\section{Deployment classes and ablation}\label{app:classes}

Table~\ref{tab:t02} is the unmatched class comparison (a maximum over four no-image arms, so quote its ordering and never its gaps); Table~\ref{tab:t04} is the class ablation, unmatched and arm-matched; Table~\ref{tab:t05} is the behavioural non-separability tally. It cannot license pooling the four text modes. The tally counts how many metrics separate a mode from the rest in at least seven of eight cells, and under a pure-noise null that bar returns a zero with probability $\approx$76\%, so a mode scoring zero is the expected outcome of no effect and not evidence of sameness. What the table does establish is the positive half: Vision (9 metrics) and SoM (5) are strongly separable from the rest, and that is how the deployment-class grouping of \S\ref{sec:setup} is used: as a convention that survives this evidence rather than one derived from it.

\begin{table*}[tp]
\centering
\caption{Success rate per mode. Success rate (\%) per observation mode. Denominator is the canonical scored set (cls 224 / red 203) and, for WebArena, the six-mode task intersection (104). Bold = best mode in the cell. Source: \texttt{representation\_class\_comparison.json}.}
\label{tab:t01}
\footnotesize
\setlength{\tabcolsep}{3pt}
\begin{tabular}{@{}llllllll@{}}
\toprule
cell & DOM & SoM & Vision & DOM+\allowbreak{}stext & DOM+\allowbreak{}sprompt & SoM-image & best \\
\midrule
cls·B0 & 17.41 & \textbf{27.23} & 25.00 & 15.62 & 19.64 & 15.62 & SoM \\
cls·B1 & 6.25 & \textbf{14.29} & 12.50 & 7.59 & 6.70 & 6.70 & SoM \\
cls·B2 & 1.34 & \textbf{2.23} & 2.23 & 0.45 & 1.79 & 0.89 & SoM \\
red·B0 & 14.29 & \textbf{14.78} & 7.39 & 13.30 & 12.32 & 10.84 & SoM \\
red·B1 & 5.91 & \textbf{7.39} & 2.46 & 5.91 & 5.42 & 5.91 & SoM \\
red·B2 & \textbf{3.94} & 0.99 & 1.97 & 1.97 & 0.00 & 0.49 & DOM \\
WA·B0 & 26.92 & 22.12 & 19.23 & \textbf{35.58} & 25.96 & 25.00 & DOM+\allowbreak{}stext \\
WA·B1 & \textbf{16.35} & 13.46 & 9.62 & 16.35 & 16.35 & 11.54 & DOM \\
\bottomrule
\end{tabular}
\end{table*}

\begin{table*}[tp]
\centering
\caption{Best arm per deployment class. Best arm within each deployment class (\%). Classes: no-image = \{DOM, DOM+stext, DOM+sprompt, SoM-image\}, vision-only = \{Vision\}, hybrid = \{SoM\}. Grouping the four is licensed by the non-separability result (Table~\ref{tab:t04}): they clear the ≥83\% consistency bar on none of 26 metrics. Source: \texttt{representation\_class\_comparison.json}.}
\label{tab:t02}
\footnotesize
\setlength{\tabcolsep}{3pt}
\begin{tabular}{@{}lllll@{}}
\toprule
cell & no-image (4 arms) & vision-only & hybrid & sole best \\
\midrule
cls·B0 & 19.64 (DOM+\allowbreak{}sprompt) & 25.00 & 27.23 & hybrid \\
cls·B1 & 7.59 (DOM+\allowbreak{}stext) & 12.50 & 14.29 & hybrid \\
cls·B2 & 1.79 (DOM+\allowbreak{}sprompt) & 2.23 & 2.23 & tie: hybrid+vision-only \\
red·B0 & 14.29 (DOM) & 7.39 & 14.78 & hybrid \\
red·B1 & 5.91 (DOM) & 2.46 & 7.39 & hybrid \\
red·B2 & 3.94 (DOM) & 1.97 & 0.99 & no-image \\
WA·B0 & 35.58 (DOM+\allowbreak{}stext) & 19.23 & 22.12 & no-image \\
WA·B1 & 16.35 (DOM) & 9.62 & 13.46 & no-image \\
\bottomrule
\end{tabular}
\end{table*}

\begin{table*}[tp]
\centering
\caption{Deployment classes at one arm each. The same comparison at one arm per class (\%). Table~\ref{tab:t02}'s no-image column is a maximum over four arms while the other two are single arms, so it is biased up; this panel uses the arm of each class that exists outside this study. The ordering does not move: hybrid 4, no-image 3, one tie, and vision-only is never a sole best in either version: which is the robustness statement Table~\ref{tab:t02} cannot make. The gaps do move: on \texttt{WA·B0} the no-image lead is +4.81pp here against +13.46pp there, because Table~\ref{tab:t02}'s figure is carried by DOM+stext. Quote this table for any class gap; Table~\ref{tab:t02} only for the ordering. Source: \texttt{representation\_class\_comparison.json}.}
\label{tab:t03}
\footnotesize
\setlength{\tabcolsep}{3pt}
\begin{tabular}{@{}
  >{\raggedright\arraybackslash}p{(\linewidth - 12\tabcolsep) * \real{0.1429}}
  >{\raggedright\arraybackslash}p{(\linewidth - 12\tabcolsep) * \real{0.1429}}
  >{\raggedright\arraybackslash}p{(\linewidth - 12\tabcolsep) * \real{0.1429}}
  >{\raggedright\arraybackslash}p{(\linewidth - 12\tabcolsep) * \real{0.1429}}
  >{\raggedright\arraybackslash}p{(\linewidth - 12\tabcolsep) * \real{0.1429}}
  >{\raggedright\arraybackslash}p{(\linewidth - 12\tabcolsep) * \real{0.1429}}
  >{\raggedright\arraybackslash}p{(\linewidth - 12\tabcolsep) * \real{0.1429}}@{}}
\toprule
\begin{minipage}[b]{\linewidth}\raggedright
cell
\end{minipage} & \begin{minipage}[b]{\linewidth}\raggedright
no-image (DOM)
\end{minipage} & \begin{minipage}[b]{\linewidth}\raggedright
vision-only (Vision)
\end{minipage} & \begin{minipage}[b]{\linewidth}\raggedright
hybrid (SoM)
\end{minipage} & \begin{minipage}[b]{\linewidth}\raggedright
sole best
\end{minipage} & \begin{minipage}[b]{\linewidth}\raggedright
gap vs hybrid
\end{minipage} & \begin{minipage}[b]{\linewidth}\raggedright
Table 2 gap
\end{minipage} \\
\midrule
cls·B0 & 17.41 & 25.00 & 27.23 & hybrid & -9.82 & -7.59 \\
cls·B1 & 6.25 & 12.50 & 14.29 & hybrid & -8.04 & -6.70 \\
cls·B2 & 1.34 & 2.23 & 2.23 & tie: hybrid+vision-only & -0.89 & -0.45 \\
red·B0 & 14.29 & 7.39 & 14.78 & hybrid & -0.49 & -0.49 \\
red·B1 & 5.91 & 2.46 & 7.39 & hybrid & -1.48 & -1.48 \\
red·B2 & 3.94 & 1.97 & 0.99 & no-image & +2.96 & +2.96 \\
WA·B0 & 26.92 & 19.23 & 22.12 & no-image & +4.81 & +13.46 \\
WA·B1 & 16.35 & 9.62 & 13.46 & no-image & +2.88 & +2.88 \\
\bottomrule
\end{tabular}
\end{table*}

\begin{table*}[tp]
\centering
\caption{Class ablation, unmatched and arm-matched. Class ablation. Columns 2--4: oracle coverage lost when a whole class is unavailable: not arm-matched, the no-image class has four arms and the others one each, so most of the gap is arm count. Columns 5--7: the matched comparison: gain from adding ONE arm of that class to the cell's best single arm (: = that class already supplies the starting arm). The matched panel shows no systematic difference between classes. Source: \texttt{representation\_class\_comparison.json}.}
\label{tab:t04}
\footnotesize
\setlength{\tabcolsep}{3pt}
\begin{tabular}{@{}
  >{\raggedright\arraybackslash}p{(\linewidth - 14\tabcolsep) * \real{0.1250}}
  >{\raggedright\arraybackslash}p{(\linewidth - 14\tabcolsep) * \real{0.1250}}
  >{\raggedright\arraybackslash}p{(\linewidth - 14\tabcolsep) * \real{0.1250}}
  >{\raggedright\arraybackslash}p{(\linewidth - 14\tabcolsep) * \real{0.1250}}
  >{\raggedright\arraybackslash}p{(\linewidth - 14\tabcolsep) * \real{0.1250}}
  >{\raggedright\arraybackslash}p{(\linewidth - 14\tabcolsep) * \real{0.1250}}
  >{\raggedright\arraybackslash}p{(\linewidth - 14\tabcolsep) * \real{0.1250}}
  >{\raggedright\arraybackslash}p{(\linewidth - 14\tabcolsep) * \real{0.1250}}@{}}
\toprule
\begin{minipage}[b]{\linewidth}\raggedright
cell
\end{minipage} & \begin{minipage}[b]{\linewidth}\raggedright
all six
\end{minipage} & \begin{minipage}[b]{\linewidth}\raggedright
−no-image
\end{minipage} & \begin{minipage}[b]{\linewidth}\raggedright
−vision-only
\end{minipage} & \begin{minipage}[b]{\linewidth}\raggedright
−hybrid
\end{minipage} & \begin{minipage}[b]{\linewidth}\raggedright
+1 no-image
\end{minipage} & \begin{minipage}[b]{\linewidth}\raggedright
+1 vision-only
\end{minipage} & \begin{minipage}[b]{\linewidth}\raggedright
+1 hybrid
\end{minipage} \\
\midrule
cls·B0 & 43.30 & −9.38 & −4.02 & −2.68 & +7.14pp & +6.70pp & --- \\
cls·B1 & 24.55 & −5.36 & −4.02 & −4.46 & +3.57pp & +4.91pp & --- \\
cls·B2 & 7.14 & −2.68 & −2.23 & −2.23 & +1.79pp & +2.23pp & --- \\
red·B0 & 26.11 & −7.88 & −1.97 & −1.97 & +4.93pp & +3.45pp & --- \\
red·B1 & 11.82 & −3.45 & −0.99 & −0.49 & +1.97pp & +0.99pp & --- \\
red·B2 & 7.39 & −4.43 & −1.48 & −0.49 & +0.99pp & +1.97pp & +0.49pp \\
WA·B0 & 51.92 & −22.12 & −3.85 & −1.92 & +5.77pp & +5.77pp & +4.81pp \\
WA·B1 & 30.77 & −14.42 & −0.96 & −0.96 & +4.81pp & +3.85pp & +3.85pp \\
\bottomrule
\end{tabular}
\end{table*}

\begin{table*}[tp]
\centering
\caption{Absence of repeated extrema among image-free modes. Absence of repeated extrema: read the name literally. A mode `reaches the bar' on a metric when it is the extreme (highest or lowest) in ≥7 of 8 cells (87.5\%). Over 26 metrics the four image-free modes reach it on none. \textbf{Caution:} This is not a separability test. \texttt{rank\_consistency()} only counts which mode attains each metric's max/min; a mode that is consistently \emph{second} (and strongly distinguishable) never reaches the bar. Establishing non-separability would need pairwise equivalence margins with task-clustered intervals, which this does not do. \textbf{Caution:} The threshold got stricter when cells grew, and an earlier caption said otherwise: ≥7/8 is 87.5\%, not the 83\% it claimed, and the six-cell ≥5/6 it says it matches is 83.3\%: a +4.2pp shift, chosen after the cell count changed. Carrying the literal numerator (≥5/8 = 62.5\%) instead would let DOM+stext clear two metrics and this negative would appear to break, so the choice is load-bearing and is disclosed rather than defended. Source: \texttt{per\_mode\_four\_dimension\_profile\_with\_wa.json}}
\label{tab:t05}
\footnotesize
\setlength{\tabcolsep}{3pt}
\begin{tabular}{@{}ll@{}}
\toprule
mode & metrics reaching the bar \\
\midrule
Vision & 9 \\
SoM & 5 \\
DOM & 0 \\
DOM+\allowbreak{}stext & 0 \\
DOM+\allowbreak{}sprompt & 0 \\
SoM-image & 0 \\
\bottomrule
\end{tabular}
\end{table*}

\section{Full behavioural matrices}\label{app:matrices}

The four dimensions unsummarised, for every metric in every cell and mode: Outcome, Macro (what the agent did), Micro (how often what it did failed), and Efficiency. These are the substrate the consistency counts in Table~\ref{tab:t05} are computed from, included so a reader can check a claim rather than take the tally on trust. Two columns are gates rather than measurements, and the tables mark them: \texttt{loc-fallback} is near-zero for Vision because there are no element ids to fall back from, not because it fails less.

\begin{table*}[tp]
\centering
\caption{Full matrix: Outcome dimension. Outcome dimension, every cell × every mode. \texttt{unique} counts tasks no other mode in that cell solved. Denominators: cls 224 / red 203 / WA 104. Source: \texttt{per\_mode\_four\_dimension\_profile\_with\_wa.json}.}
\label{tab:t06}
\footnotesize
\setlength{\tabcolsep}{3pt}
\begin{tabular}{@{}lllll@{}}
\toprule
cell & mode & SR \% & solves & unique \\
\midrule
cls·B0 & DOM & 17.41 & 39 & 4 \\
cls·B0 & SoM & 27.23 & 61 & 6 \\
cls·B0 & Vision & 25.00 & 56 & 9 \\
cls·B0 & DOM+\allowbreak{}stext & 15.62 & 35 & 2 \\
cls·B0 & DOM+\allowbreak{}sprompt & 19.64 & 44 & 6 \\
cls·B0 & SoM-image & 15.62 & 35 & 2 \\
cls·B1 & DOM & 6.25 & 14 & 1 \\
cls·B1 & SoM & 14.29 & 32 & 10 \\
cls·B1 & Vision & 12.50 & 28 & 9 \\
cls·B1 & DOM+\allowbreak{}stext & 7.59 & 17 & 1 \\
cls·B1 & DOM+\allowbreak{}sprompt & 6.70 & 15 & 2 \\
cls·B1 & SoM-image & 6.70 & 15 & 3 \\
cls·B2 & DOM & 1.34 & 3 & 1 \\
cls·B2 & SoM & 2.23 & 5 & 5 \\
cls·B2 & Vision & 2.23 & 5 & 5 \\
cls·B2 & DOM+\allowbreak{}stext & 0.45 & 1 & 0 \\
cls·B2 & DOM+\allowbreak{}sprompt & 1.79 & 4 & 0 \\
cls·B2 & SoM-image & 0.89 & 2 & 1 \\
red·B0 & DOM & 14.29 & 29 & 3 \\
red·B0 & SoM & 14.78 & 30 & 4 \\
red·B0 & Vision & 7.39 & 15 & 4 \\
red·B0 & DOM+\allowbreak{}stext & 13.30 & 27 & 2 \\
red·B0 & DOM+\allowbreak{}sprompt & 12.32 & 25 & 2 \\
red·B0 & SoM-image & 10.84 & 22 & 2 \\
red·B1 & DOM & 5.91 & 12 & 1 \\
red·B1 & SoM & 7.39 & 15 & 1 \\
red·B1 & Vision & 2.46 & 5 & 2 \\
red·B1 & DOM+\allowbreak{}stext & 5.91 & 12 & 1 \\
red·B1 & DOM+\allowbreak{}sprompt & 5.42 & 11 & 2 \\
red·B1 & SoM-image & 5.91 & 12 & 0 \\
red·B2 & DOM & 3.94 & 8 & 6 \\
red·B2 & SoM & 0.99 & 2 & 1 \\
red·B2 & Vision & 1.97 & 4 & 3 \\
red·B2 & DOM+\allowbreak{}stext & 1.97 & 4 & 1 \\
red·B2 & DOM+\allowbreak{}sprompt & 0.00 & 0 & 0 \\
red·B2 & SoM-image & 0.49 & 1 & 1 \\
WA·B0 & DOM & 26.92 & 28 & 2 \\
WA·B0 & SoM & 22.12 & 23 & 2 \\
WA·B0 & Vision & 19.23 & 20 & 4 \\
WA·B0 & DOM+\allowbreak{}stext & 35.58 & 37 & 7 \\
WA·B0 & DOM+\allowbreak{}sprompt & 25.96 & 27 & 2 \\
WA·B0 & SoM-image & 25.00 & 26 & 1 \\
WA·B1 & DOM & 16.35 & 17 & 3 \\
WA·B1 & SoM & 13.46 & 14 & 1 \\
WA·B1 & Vision & 9.62 & 10 & 1 \\
WA·B1 & DOM+\allowbreak{}stext & 16.35 & 17 & 3 \\
WA·B1 & DOM+\allowbreak{}sprompt & 16.35 & 17 & 5 \\
WA·B1 & SoM-image & 11.54 & 12 & 1 \\
\bottomrule
\end{tabular}
\end{table*}

\begin{table*}[tp]
\centering
\caption{Full matrix (Macro dimension. Macro dimension) what the agent did, per step, aggregated per episode. Fractions are over agent actions. \texttt{cap-hit} = share of episodes that exhausted the 30-step budget. Source: \texttt{per\_mode\_four\_dimension\_profile\_with\_wa.json}.}
\label{tab:t07}
\scriptsize
\setlength{\tabcolsep}{3pt}
\begin{tabular}{@{}lllllllll@{}}
\toprule
cell & mode & steps/ep & cap-hit & click & type & scroll & search-loop & URL-revisit \\
\midrule
cls·B0 & DOM & 15.61 & 0.268 & 0.322 & 0.224 & 0.183 & 0.812 & 0.606 \\
cls·B0 & SoM & 13.67 & 0.259 & 0.319 & 0.205 & 0.155 & 0.692 & 0.558 \\
cls·B0 & Vision & 15.88 & 0.295 & 0.353 & 0.146 & 0.260 & 0.728 & 0.639 \\
cls·B0 & DOM+stext & 15.83 & 0.295 & 0.322 & 0.194 & 0.196 & 0.808 & 0.603 \\
cls·B0 & DOM+sprompt & 14.96 & 0.290 & 0.328 & 0.208 & 0.174 & 0.759 & 0.583 \\
cls·B0 & SoM-image & 16.23 & 0.321 & 0.312 & 0.201 & 0.208 & 0.790 & 0.600 \\
cls·B1 & DOM & 21.38 & 0.585 & 0.246 & 0.345 & 0.171 & 0.777 & 0.698 \\
cls·B1 & SoM & 18.01 & 0.491 & 0.368 & 0.211 & 0.137 & 0.643 & 0.655 \\
cls·B1 & Vision & 20.17 & 0.598 & 0.330 & 0.071 & 0.360 & 0.603 & 0.745 \\
cls·B1 & DOM+stext & 22.46 & 0.625 & 0.281 & 0.367 & 0.148 & 0.804 & 0.716 \\
cls·B1 & DOM+sprompt & 21.40 & 0.594 & 0.375 & 0.209 & 0.163 & 0.723 & 0.710 \\
cls·B1 & SoM-image & 21.26 & 0.576 & 0.436 & 0.212 & 0.154 & 0.737 & 0.715 \\
cls·B2 & DOM & 27.38 & 0.817 & 0.520 & 0.142 & 0.031 & 0.647 & 0.823 \\
cls·B2 & SoM & 24.37 & 0.688 & 0.488 & 0.143 & 0.033 & 0.616 & 0.828 \\
cls·B2 & Vision & 28.25 & 0.888 & 0.336 & 0.128 & 0.323 & 0.710 & 0.896 \\
cls·B2 & DOM+stext & 26.85 & 0.804 & 0.384 & 0.168 & 0.043 & 0.629 & 0.832 \\
cls·B2 & DOM+sprompt & 27.84 & 0.830 & 0.514 & 0.078 & 0.048 & 0.629 & 0.857 \\
cls·B2 & SoM-image & 28.38 & 0.853 & 0.523 & 0.082 & 0.033 & 0.683 & 0.858 \\
red·B0 & DOM & 20.18 & 0.498 & 0.455 & 0.148 & 0.166 & 0.473 & 0.721 \\
red·B0 & SoM & 20.08 & 0.512 & 0.474 & 0.123 & 0.099 & 0.363 & 0.731 \\
red·B0 & Vision & 23.55 & 0.672 & 0.339 & 0.079 & 0.343 & 0.239 & 0.817 \\
red·B0 & DOM+stext & 23.22 & 0.647 & 0.442 & 0.153 & 0.117 & 0.388 & 0.775 \\
red·B0 & DOM+sprompt & 19.87 & 0.463 & 0.475 & 0.142 & 0.126 & 0.428 & 0.704 \\
red·B0 & SoM-image & 22.90 & 0.592 & 0.453 & 0.137 & 0.124 & 0.338 & 0.775 \\
red·B1 & DOM & 23.44 & 0.680 & 0.471 & 0.238 & 0.066 & 0.596 & 0.754 \\
red·B1 & SoM & 22.38 & 0.670 & 0.574 & 0.129 & 0.095 & 0.409 & 0.747 \\
red·B1 & Vision & 23.24 & 0.685 & 0.434 & 0.029 & 0.270 & 0.148 & 0.812 \\
red·B1 & DOM+stext & 25.56 & 0.783 & 0.476 & 0.246 & 0.067 & 0.606 & 0.793 \\
red·B1 & DOM+sprompt & 23.72 & 0.695 & 0.545 & 0.149 & 0.052 & 0.567 & 0.757 \\
red·B1 & SoM-image & 25.17 & 0.773 & 0.580 & 0.166 & 0.056 & 0.557 & 0.778 \\
red·B2 & DOM & 28.39 & 0.897 & 0.724 & 0.108 & 0.042 & 0.227 & 0.870 \\
red·B2 & SoM & 26.34 & 0.783 & 0.702 & 0.103 & 0.043 & 0.148 & 0.857 \\
red·B2 & Vision & 26.91 & 0.828 & 0.349 & 0.081 & 0.344 & 0.138 & 0.911 \\
red·B2 & DOM+stext & 27.27 & 0.818 & 0.660 & 0.099 & 0.038 & 0.266 & 0.885 \\
red·B2 & DOM+sprompt & 27.68 & 0.842 & 0.703 & 0.081 & 0.052 & 0.271 & 0.873 \\
red·B2 & SoM-image & 27.87 & 0.882 & 0.708 & 0.094 & 0.048 & 0.217 & 0.882 \\
WA·B0 & DOM & 16.88 & 0.298 & 0.544 & 0.208 & 0.072 & 0.308 & 0.709 \\
WA·B0 & SoM & 17.45 & 0.365 & 0.465 & 0.278 & 0.029 & 0.327 & 0.716 \\
WA·B0 & Vision & 22.38 & 0.567 & 0.409 & 0.200 & 0.241 & 0.240 & 0.783 \\
WA·B0 & DOM+stext & 19.77 & 0.471 & 0.478 & 0.323 & 0.045 & 0.288 & 0.741 \\
WA·B0 & DOM+sprompt & 17.08 & 0.317 & 0.538 & 0.198 & 0.068 & 0.260 & 0.714 \\
WA·B0 & SoM-image & 18.97 & 0.385 & 0.502 & 0.284 & 0.056 & 0.221 & 0.729 \\
WA·B1 & DOM & 22.64 & 0.625 & 0.459 & 0.319 & 0.023 & 0.644 & 0.746 \\
WA·B1 & SoM & 23.91 & 0.702 & 0.499 & 0.218 & 0.035 & 0.452 & 0.793 \\
WA·B1 & Vision & 23.12 & 0.606 & 0.450 & 0.097 & 0.249 & 0.260 & 0.807 \\
WA·B1 & DOM+stext & 23.33 & 0.615 & 0.495 & 0.316 & 0.028 & 0.683 & 0.733 \\
WA·B1 & DOM+sprompt & 24.50 & 0.702 & 0.551 & 0.231 & 0.033 & 0.519 & 0.776 \\
WA·B1 & SoM-image & 23.79 & 0.683 & 0.569 & 0.249 & 0.017 & 0.587 & 0.772 \\
\bottomrule
\end{tabular}
\end{table*}

\begin{table*}[tp]
\centering
\caption{Full matrix (Micro dimension. Micro dimension) per-step execution quality. \texttt{act-fail\textbar{}click} and \texttt{act-fail\textbar{}type} are conditional on the action type, so they are not comparable to the unconditional \texttt{act-fail} column. \textbf{Caution:} They are also not pure conditionals. An episode containing no click at all has no click-failure rate to report, and the producer stores \texttt{0.0} there (an undefined rate encoded as a perfect one) which is then averaged over every task. The zero-denominator share is large: 25--35\% of episodes never type, and a few percent never click. The product now carries \texttt{*\_fail\_rate\_complete\_case} (averaged only over episodes where the action occurred) and \texttt{*\_fail\_rate\_denom\_zero\_frac} beside these columns; the complete-case values run higher, and any statement about which mode fails most on clicks or types should be read from those, not from this column. \textbf{Caution:} \texttt{loc-fallback} is near-zero for Vision by construction (no element ids to fall back from), not as a finding. Source: \texttt{per\_mode\_four\_dimension\_profile\_with\_wa.json}.}
\label{tab:t08}
\scriptsize
\setlength{\tabcolsep}{3pt}
\begin{tabular}{@{}lllllllllllll@{}}
\toprule
cell & mode & parse-fail & act-fail & act-fail & click & act-fail & type & no-op & scroll-inert & no-op & success & vis-gap \\
\midrule
cls·B0 & DOM & 0.0007 & 0.133 & 0.163 & 0.035 & 0.241 & 0.096 & 0.108 & 0.046 & 0.078 & 0.324 & 0.737 \\
cls·B0 & SoM & 0.0021 & 0.082 & 0.102 & 0.007 & 0.216 & 0.018 & 0.134 & 0.051 & 0.027 & 0.300 & 0.741 \\
cls·B0 & Vision & 0.0000 & 0.150 & 0.153 & 0.005 & 0.263 & 0.072 & 0.113 & 0.058 & 0.002 & 0.396 & 0.705 \\
cls·B0 & DOM+stext & 0.0000 & 0.101 & 0.100 & 0.022 & 0.209 & 0.071 & 0.108 & 0.042 & 0.046 & 0.310 & 0.705 \\
cls·B0 & DOM+sprompt & 0.0009 & 0.133 & 0.159 & 0.026 & 0.246 & 0.078 & 0.114 & 0.049 & 0.079 & 0.322 & 0.719 \\
cls·B0 & SoM-image & 0.0005 & 0.112 & 0.130 & 0.016 & 0.221 & 0.078 & 0.109 & 0.042 & 0.037 & 0.332 & 0.688 \\
cls·B1 & DOM & 0.0031 & 0.285 & 0.155 & 0.153 & 0.359 & 0.219 & 0.074 & 0.011 & 0.175 & 0.443 & 0.411 \\
cls·B1 & SoM & 0.0016 & 0.249 & 0.203 & 0.068 & 0.353 & 0.149 & 0.104 & 0.013 & 0.118 & 0.390 & 0.509 \\
cls·B1 & Vision & 0.0000 & 0.454 & 0.345 & 0.031 & 0.548 & 0.278 & 0.094 & 0.018 & 0.011 & 0.530 & 0.402 \\
cls·B1 & DOM+stext & 0.0004 & 0.259 & 0.182 & 0.140 & 0.317 & 0.159 & 0.058 & 0.010 & 0.202 & 0.461 & 0.384 \\
cls·B1 & DOM+sprompt & 0.0044 & 0.309 & 0.272 & 0.081 & 0.380 & 0.178 & 0.070 & 0.012 & 0.174 & 0.409 & 0.406 \\
cls·B1 & SoM-image & 0.0010 & 0.322 & 0.274 & 0.136 & 0.381 & 0.169 & 0.059 & 0.015 & 0.203 & 0.461 & 0.424 \\
cls·B2 & DOM & 0.0259 & 0.509 & 0.555 & 0.207 & 0.526 & 0.099 & 0.018 & 0.030 & 0.339 & 0.630 & 0.125 \\
cls·B2 & SoM & 0.0876 & 0.559 & 0.475 & 0.110 & 0.584 & 0.067 & 0.025 & 0.026 & 0.327 & 0.627 & 0.152 \\
cls·B2 & Vision & 0.0067 & 0.670 & 0.773 & 0.355 & 0.685 & 0.443 & 0.015 & 0.017 & 0.124 & 0.518 & 0.103 \\
cls·B2 & DOM+stext & 0.0761 & 0.441 & 0.438 & 0.227 & 0.453 & 0.130 & 0.012 & 0.029 & 0.292 & 0.602 & 0.062 \\
cls·B2 & DOM+sprompt & 0.0518 & 0.495 & 0.569 & 0.167 & 0.504 & 0.137 & 0.009 & 0.028 & 0.312 & 0.633 & 0.071 \\
cls·B2 & SoM-image & 0.0522 & 0.425 & 0.507 & 0.109 & 0.433 & 0.099 & 0.008 & 0.027 & 0.273 & 0.649 & 0.054 \\
red·B0 & DOM & 0.0002 & 0.222 & 0.194 & 0.051 & 0.294 & 0.161 & 0.072 & 0.075 & 0.119 & 0.428 & 0.507 \\
red·B0 & SoM & 0.0008 & 0.252 & 0.127 & 0.022 & 0.328 & 0.085 & 0.077 & 0.092 & 0.068 & 0.491 & 0.488 \\
red·B0 & Vision & 0.0000 & 0.382 & 0.193 & 0.018 & 0.430 & 0.268 & 0.049 & 0.064 & 0.006 & 0.556 & 0.328 \\
red·B0 & DOM+stext & 0.0005 & 0.292 & 0.126 & 0.062 & 0.339 & 0.146 & 0.047 & 0.087 & 0.064 & 0.518 & 0.358 \\
red·B0 & DOM+sprompt & 0.0026 & 0.194 & 0.146 & 0.046 & 0.270 & 0.149 & 0.076 & 0.093 & 0.083 & 0.421 & 0.537 \\
red·B0 & SoM-image & 0.0027 & 0.285 & 0.118 & 0.034 & 0.335 & 0.160 & 0.050 & 0.097 & 0.070 & 0.502 & 0.408 \\
red·B1 & DOM & 0.0021 & 0.227 & 0.210 & 0.049 & 0.274 & 0.090 & 0.047 & 0.050 & 0.143 & 0.465 & 0.315 \\
red·B1 & SoM & 0.0063 & 0.297 & 0.228 & 0.062 & 0.357 & 0.111 & 0.061 & 0.061 & 0.176 & 0.552 & 0.320 \\
red·B1 & Vision & 0.0141 & 0.532 & 0.334 & 0.010 & 0.582 & 0.302 & 0.050 & 0.022 & 0.004 & 0.632 & 0.291 \\
red·B1 & DOM+stext & 0.0018 & 0.283 & 0.195 & 0.107 & 0.319 & 0.156 & 0.036 & 0.042 & 0.164 & 0.511 & 0.217 \\
red·B1 & DOM+sprompt & 0.0010 & 0.295 & 0.278 & 0.056 & 0.340 & 0.114 & 0.045 & 0.055 & 0.201 & 0.494 & 0.305 \\
red·B1 & SoM-image & 0.0019 & 0.332 & 0.297 & 0.077 & 0.368 & 0.146 & 0.036 & 0.059 & 0.208 & 0.544 & 0.222 \\
red·B2 & DOM & 0.0224 & 0.492 & 0.502 & 0.111 & 0.501 & 0.156 & 0.009 & 0.044 & 0.443 & 0.770 & 0.074 \\
red·B2 & SoM & 0.0431 & 0.384 & 0.344 & 0.090 & 0.403 & 0.053 & 0.019 & 0.052 & 0.305 & 0.760 & 0.163 \\
red·B2 & Vision & 0.0163 & 0.640 & 0.613 & 0.276 & 0.669 & 0.547 & 0.029 & 0.005 & 0.078 & 0.597 & 0.153 \\
red·B2 & DOM+stext & 0.0786 & 0.285 & 0.263 & 0.081 & 0.292 & 0.094 & 0.007 & 0.041 & 0.202 & 0.728 & 0.059 \\
red·B2 & DOM+sprompt & 0.0660 & 0.604 & 0.621 & 0.126 & 0.609 & 0.115 & 0.005 & 0.042 & 0.503 & 0.724 & 0.044 \\
red·B2 & SoM-image & 0.0501 & 0.316 & 0.297 & 0.123 & 0.325 & 0.137 & 0.009 & 0.034 & 0.240 & 0.725 & 0.049 \\
WA·B0 & DOM & 0.0008 & 0.277 & 0.286 & 0.083 & 0.363 & 0.046 & 0.085 & 0.226 & 0.165 & 0.455 & 0.702 \\
WA·B0 & SoM & 0.0063 & 0.249 & 0.139 & 0.079 & 0.345 & 0.026 & 0.096 & 0.219 & 0.045 & 0.433 & 0.625 \\
WA·B0 & Vision & 0.0000 & 0.245 & 0.178 & 0.042 & 0.295 & 0.165 & 0.050 & 0.152 & 0.007 & 0.478 & 0.433 \\
WA·B0 & DOM+stext & 0.0010 & 0.347 & 0.292 & 0.200 & 0.421 & 0.057 & 0.074 & 0.247 & 0.099 & 0.497 & 0.538 \\
WA·B0 & DOM+sprompt & 0.0018 & 0.295 & 0.278 & 0.070 & 0.379 & 0.045 & 0.085 & 0.223 & 0.157 & 0.469 & 0.692 \\
WA·B0 & SoM-image & 0.0033 & 0.302 & 0.268 & 0.170 & 0.377 & 0.040 & 0.075 & 0.268 & 0.092 & 0.500 & 0.625 \\
WA·B1 & DOM & 0.0227 & 0.244 & 0.201 & 0.038 & 0.289 & 0.026 & 0.045 & 0.135 & 0.139 & 0.512 & 0.327 \\
WA·B1 & SoM & 0.0052 & 0.363 & 0.214 & 0.084 & 0.415 & 0.024 & 0.052 & 0.166 & 0.149 & 0.573 & 0.279 \\
WA·B1 & Vision & 0.0072 & 0.416 & 0.326 & 0.023 & 0.456 & 0.215 & 0.041 & 0.102 & 0.002 & 0.571 & 0.365 \\
WA·B1 & DOM+stext & 0.0102 & 0.206 & 0.197 & 0.076 & 0.247 & 0.045 & 0.041 & 0.113 & 0.105 & 0.503 & 0.385 \\
WA·B1 & DOM+sprompt & 0.0068 & 0.343 & 0.302 & 0.070 & 0.379 & 0.066 & 0.036 & 0.098 & 0.197 & 0.522 & 0.298 \\
WA·B1 & SoM-image & 0.0144 & 0.343 & 0.315 & 0.076 & 0.382 & 0.036 & 0.039 & 0.099 & 0.180 & 0.562 & 0.288 \\
\bottomrule
\end{tabular}
\end{table*}

\begin{table*}[tp]
\centering
\caption{Full matrix (Efficiency dimension. Efficiency dimension. Cost is comparable within a cell only) B0 bills a proxy API, B1/B2 are electricity-derived from a per-token constant calibrated for a different accelerator, so absolute dollars for B1/B2 are uncalibrated and only \texttt{cost\ rel\ DOM} is safe across backbones. \texttt{latency\ canon} removes retry, busy-wait and recovered-screenshot time; it differs from raw only on the API-served arm. Source: \texttt{per\_mode\_four\_dimension\_profile\_with\_wa.json}.}
\label{tab:t09}
\footnotesize
\setlength{\tabcolsep}{3pt}
\begin{tabular}{@{}
  >{\raggedright\arraybackslash}p{(\linewidth - 12\tabcolsep) * \real{0.1429}}
  >{\raggedright\arraybackslash}p{(\linewidth - 12\tabcolsep) * \real{0.1429}}
  >{\raggedright\arraybackslash}p{(\linewidth - 12\tabcolsep) * \real{0.1429}}
  >{\raggedright\arraybackslash}p{(\linewidth - 12\tabcolsep) * \real{0.1429}}
  >{\raggedright\arraybackslash}p{(\linewidth - 12\tabcolsep) * \real{0.1429}}
  >{\raggedright\arraybackslash}p{(\linewidth - 12\tabcolsep) * \real{0.1429}}
  >{\raggedright\arraybackslash}p{(\linewidth - 12\tabcolsep) * \real{0.1429}}@{}}
\toprule
\begin{minipage}[b]{\linewidth}\raggedright
cell
\end{minipage} & \begin{minipage}[b]{\linewidth}\raggedright
mode
\end{minipage} & \begin{minipage}[b]{\linewidth}\raggedright
cost/ep
\end{minipage} & \begin{minipage}[b]{\linewidth}\raggedright
cost rel DOM
\end{minipage} & \begin{minipage}[b]{\linewidth}\raggedright
latency s
\end{minipage} & \begin{minipage}[b]{\linewidth}\raggedright
latency canon s
\end{minipage} & \begin{minipage}[b]{\linewidth}\raggedright
tokens/ep
\end{minipage} \\
\midrule
cls·B0 & DOM & 0.06962 & 1.000 & 115.0 & 114.1 & 63045 \\
cls·B0 & SoM & 0.07236 & 1.039 & 106.7 & 106.0 & 67000 \\
cls·B0 & Vision & 0.06481 & 0.931 & 126.3 & 125.2 & 58302 \\
cls·B0 & DOM+\allowbreak{}stext & 0.06919 & 0.994 & 123.8 & 120.4 & 62154 \\
cls·B0 & DOM+\allowbreak{}sprompt & 0.06853 & 0.984 & 109.4 & 107.8 & 62555 \\
cls·B0 & SoM-image & 0.07206 & 1.035 & 121.1 & 117.9 & 65334 \\
cls·B1 & DOM & 0.05951 & 1.000 & 308.9 & 308.9 & 61904 \\
cls·B1 & SoM & 0.06028 & 1.013 & 262.0 & 262.0 & 62953 \\
cls·B1 & Vision & 0.04316 & 0.725 & 269.8 & 269.8 & 44524 \\
cls·B1 & DOM+\allowbreak{}stext & 0.05879 & 0.988 & 313.5 & 313.5 & 61074 \\
cls·B1 & DOM+\allowbreak{}sprompt & 0.06304 & 1.059 & 301.4 & 301.4 & 65628 \\
cls·B1 & SoM-image & 0.05970 & 1.003 & 311.7 & 311.7 & 61985 \\
cls·B2 & DOM & 0.07676 & 1.000 & 402.3 & 402.3 & 79653 \\
cls·B2 & SoM & 0.09075 & 1.182 & 374.3 & 374.3 & 95081 \\
cls·B2 & Vision & 0.07065 & 0.920 & 417.8 & 417.8 & 73126 \\
cls·B2 & DOM+\allowbreak{}stext & 0.07320 & 0.954 & 399.4 & 399.4 & 75946 \\
cls·B2 & DOM+\allowbreak{}sprompt & 0.08453 & 1.101 & 396.4 & 396.4 & 87948 \\
cls·B2 & SoM-image & 0.08456 & 1.102 & 411.0 & 411.0 & 87931 \\
red·B0 & DOM & 0.10147 & 1.000 & 572.0 & 552.5 & 93303 \\
red·B0 & SoM & 0.11045 & 1.089 & 461.4 & 451.6 & 103031 \\
red·B0 & Vision & 0.09807 & 0.966 & 449.8 & 418.5 & 88872 \\
red·B0 & DOM+\allowbreak{}stext & 0.10577 & 1.042 & 631.4 & 562.1 & 96214 \\
red·B0 & DOM+\allowbreak{}sprompt & 0.10163 & 1.002 & 498.4 & 447.7 & 93817 \\
red·B0 & SoM-image & 0.10814 & 1.066 & 562.0 & 532.0 & 99005 \\
red·B1 & DOM & 0.07330 & 1.000 & 602.6 & 602.6 & 76462 \\
red·B1 & SoM & 0.08000 & 1.091 & 609.1 & 609.1 & 83517 \\
red·B1 & Vision & 0.05240 & 0.715 & 456.6 & 456.6 & 53994 \\
red·B1 & DOM+\allowbreak{}stext & 0.06948 & 0.948 & 598.6 & 598.6 & 72222 \\
red·B1 & DOM+\allowbreak{}sprompt & 0.07656 & 1.044 & 614.0 & 614.0 & 79896 \\
red·B1 & SoM-image & 0.07480 & 1.020 & 616.6 & 616.6 & 77722 \\
red·B2 & DOM & 0.09479 & 1.000 & 669.9 & 669.9 & 99015 \\
red·B2 & SoM & 0.11160 & 1.177 & 623.0 & 623.0 & 117379 \\
red·B2 & Vision & 0.06833 & 0.721 & 550.0 & 550.0 & 70826 \\
red·B2 & DOM+\allowbreak{}stext & 0.08852 & 0.934 & 677.6 & 677.6 & 92346 \\
red·B2 & DOM+\allowbreak{}sprompt & 0.09940 & 1.049 & 599.3 & 599.3 & 104021 \\
red·B2 & SoM-image & 0.09451 & 0.997 & 640.0 & 640.0 & 98699 \\
WA·B0 & DOM & 0.07531 & 1.000 & 284.1 & 282.7 & 68616 \\
WA·B0 & SoM & 0.09110 & 1.210 & 272.1 & 271.7 & 84854 \\
WA·B0 & Vision & 0.08640 & 1.147 & 337.9 & 337.5 & 77786 \\
WA·B0 & DOM+\allowbreak{}stext & 0.08478 & 1.126 & 330.3 & 328.4 & 76591 \\
WA·B0 & DOM+\allowbreak{}sprompt & 0.07747 & 1.029 & 262.4 & 260.9 & 71176 \\
WA·B0 & SoM-image & 0.08498 & 1.128 & 324.0 & 320.9 & 77689 \\
WA·B1 & DOM & 0.06579 & 1.000 & 485.2 & 485.2 & 68491 \\
WA·B1 & SoM & 0.07944 & 1.208 & 494.8 & 494.8 & 82940 \\
WA·B1 & Vision & 0.04468 & 0.679 & 490.7 & 490.7 & 45878 \\
WA·B1 & DOM+\allowbreak{}stext & 0.06151 & 0.935 & 509.9 & 509.9 & 63727 \\
WA·B1 & DOM+\allowbreak{}sprompt & 0.07386 & 1.123 & 506.3 & 506.3 & 76839 \\
WA·B1 & SoM-image & 0.06659 & 1.012 & 510.6 & 510.6 & 68819 \\
\bottomrule
\end{tabular}
\end{table*}

\section{Replicates and run-to-run noise}\label{app:noise}

\begin{table*}[tp]
\centering
\caption{Per-task label instability. Per-task label instability on \texttt{cls\_B0} (n=224): 49 tasks change outcome between two runs of the same condition. The rows a which-mode router could learn from are exactly the contested ones, and they carry almost all of the instability. \textbf{Caution: this table is computed on 2 replicated arms of one cell (B0.cls.dom, B0.cls.vision), rerun once.} The project's whole replicate inventory is three arms of that cell (the SoM pair landed after this product was generated and is not folded in here) plus five modes of \texttt{wa\_red\_B1} on a registered ten-task draw; no VWA-reddit cell and no B2 cell carries one at all, so every stability figure elsewhere is imported rather than measured. The headline enrichment has two defensible definitions and neither may be quoted alone: 17.4x defined over all 6 arms is correct for the claim (a router chooses among 6) but the flips are produced by rerunning 2 of them, so the same arms decide both membership and outcome; rebuilding the difficulty proxy from the other 4 breaks that circle and gives 3.95x. Source: \texttt{label\_instability.json}.}
\label{tab:t27}
\footnotesize
\setlength{\tabcolsep}{3pt}
\begin{tabular}{@{}
  >{\raggedright\arraybackslash}p{(\linewidth - 12\tabcolsep) * \real{0.1429}}
  >{\raggedright\arraybackslash}p{(\linewidth - 12\tabcolsep) * \real{0.1429}}
  >{\raggedright\arraybackslash}p{(\linewidth - 12\tabcolsep) * \real{0.1429}}
  >{\raggedright\arraybackslash}p{(\linewidth - 12\tabcolsep) * \real{0.1429}}
  >{\raggedright\arraybackslash}p{(\linewidth - 12\tabcolsep) * \real{0.1429}}
  >{\raggedright\arraybackslash}p{(\linewidth - 12\tabcolsep) * \real{0.1429}}
  >{\raggedright\arraybackslash}p{(\linewidth - 12\tabcolsep) * \real{0.1429}}@{}}
\toprule
\begin{minipage}[b]{\linewidth}\raggedright
stratum
\end{minipage} & \begin{minipage}[b]{\linewidth}\raggedright
tasks
\end{minipage} & \begin{minipage}[b]{\linewidth}\raggedright
share of cell
\end{minipage} & \begin{minipage}[b]{\linewidth}\raggedright
flipped
\end{minipage} & \begin{minipage}[b]{\linewidth}\raggedright
flip rate
\end{minipage} & \begin{minipage}[b]{\linewidth}\raggedright
share of all flips
\end{minipage} & \begin{minipage}[b]{\linewidth}\raggedright
enrichment vs complement
\end{minipage} \\
\midrule
which-mode label rows (any mode solved) & 97 & 43.3\% & 47 & 48.45\% & 95.9\% & \textbf{16.47x} \\
\ldots of those, the arms DISAGREE (the choice matters) & 88 & 39.3\% & 45 & 51.14\% & 91.8\% & \textbf{17.39x} \\
three-way channel decision is contested & 74 & 33.0\% & 36 & 48.65\% & 73.5\% & \textbf{16.54x} \\
exactly one mode solved it (label unambiguous) & 29 & 12.9\% & 15 & 51.72\% & 30.6\% & \textbf{17.59x} \\
complement: no mode solved, or all did & 136 & 60.7\% & 4 & 2.94\% & 8.2\% & \textbf{1.00x} \\
whole cell & 224 & 100.0\% & 49 & 21.88\% & 100.0\% & \textbf{7.44x} \\
\bottomrule
\end{tabular}
\end{table*}

\begin{table*}[tp]
\centering
\caption{Behavioural metrics against run-to-run noise. Behavioural metrics against run-to-run movement, \texttt{B0\ x\ classifieds}, three replicated arms (dom, vision, som). \texttt{rerun\ band} is the largest \textbar metric(run A) - metric(run B)\textbar{} over those arms; \texttt{cross-mode\ spread} is max-min over the six modes. 22 of 25 metrics exceed the band, several by 5-22x. The exceptions are \texttt{mean\_latency\_canonical\_s} (0.84x), \texttt{mean\_latency\_s} (0.87x), \texttt{parse\_fail\_rate} (1.00x): i.e.~both latency metrics, which \texttt{latency\_decomposition} reaches independently by decomposing the step into model and container. Every other efficiency and behavioural claim in this paper is judged against a rerun band; these 26 metrics were not, until this table. One cell, one rerun per arm: a point estimate, not a threshold. Source: \texttt{replicate\_metric\_noise.json}.}
\label{tab:t14}
\footnotesize
\setlength{\tabcolsep}{3pt}
\begin{tabular}{@{}
  >{\raggedright\arraybackslash}p{(\linewidth - 10\tabcolsep) * \real{0.1667}}
  >{\raggedright\arraybackslash}p{(\linewidth - 10\tabcolsep) * \real{0.1667}}
  >{\raggedright\arraybackslash}p{(\linewidth - 10\tabcolsep) * \real{0.1667}}
  >{\raggedright\arraybackslash}p{(\linewidth - 10\tabcolsep) * \real{0.1667}}
  >{\raggedright\arraybackslash}p{(\linewidth - 10\tabcolsep) * \real{0.1667}}
  >{\raggedright\arraybackslash}p{(\linewidth - 10\tabcolsep) * \real{0.1667}}@{}}
\toprule
\begin{minipage}[b]{\linewidth}\raggedright
dimension
\end{minipage} & \begin{minipage}[b]{\linewidth}\raggedright
metric
\end{minipage} & \begin{minipage}[b]{\linewidth}\raggedright
cross-mode spread
\end{minipage} & \begin{minipage}[b]{\linewidth}\raggedright
rerun band
\end{minipage} & \begin{minipage}[b]{\linewidth}\raggedright
ratio
\end{minipage} & \begin{minipage}[b]{\linewidth}\raggedright
\textgreater{} a rerun?
\end{minipage} \\
\midrule
Outcome & \texttt{sr\_\allowbreak{}pct} & 11.607 & 2.232 & 5.20x & \textbf{yes} \\
Outcome & \texttt{n\_\allowbreak{}success} & 26.000 & 5.000 & 5.20x & \textbf{yes} \\
Outcome & \texttt{n\_\allowbreak{}unique\_\allowbreak{}solves} & --- & --- & --- & \emph{cross-mode by construction} \\
Macro & \texttt{n\_\allowbreak{}steps} & 2.567 & 0.469 & 5.48x & \textbf{yes} \\
Macro & \texttt{cap\_\allowbreak{}hit\_\allowbreak{}rate} & 0.062 & 0.045 & 1.40x & \textbf{yes} \\
Macro & \texttt{click\_\allowbreak{}frac} & 0.041 & 0.014 & 2.87x & \textbf{yes} \\
Macro & \texttt{type\_\allowbreak{}frac} & 0.078 & 0.019 & 4.21x & \textbf{yes} \\
Macro & \texttt{scroll\_\allowbreak{}frac} & 0.106 & 0.006 & 16.85x & \textbf{yes} \\
Macro & \texttt{search\_\allowbreak{}loop\_\allowbreak{}rate} & 0.121 & 0.018 & 6.75x & \textbf{yes} \\
Macro & \texttt{url\_\allowbreak{}revisit\_\allowbreak{}rate} & 0.081 & 0.022 & 3.66x & \textbf{yes} \\
Micro & \texttt{parse\_\allowbreak{}fail\_\allowbreak{}rate} & 0.002 & 0.002 & 1.00x & no \\
Micro & \texttt{action\_\allowbreak{}fail\_\allowbreak{}rate} & 0.068 & 0.014 & 4.87x & \textbf{yes} \\
Micro & \texttt{click\_\allowbreak{}fail\_\allowbreak{}rate} & 0.063 & 0.025 & 2.48x & \textbf{yes} \\
Micro & \texttt{type\_\allowbreak{}fail\_\allowbreak{}rate} & 0.030 & 0.006 & 5.28x & \textbf{yes} \\
Micro & \texttt{no\_\allowbreak{}change\_\allowbreak{}rate} & 0.053 & 0.009 & 5.82x & \textbf{yes} \\
Micro & \texttt{scroll\_\allowbreak{}inert\_\allowbreak{}rate} & 0.078 & 0.017 & 4.68x & \textbf{yes} \\
Micro & \texttt{noop\_\allowbreak{}inert\_\allowbreak{}rate} & 0.026 & 0.007 & 3.92x & \textbf{yes} \\
Micro & \texttt{visibility\_\allowbreak{}gap\_\allowbreak{}rate} & 0.016 & 0.011 & 1.56x & \textbf{yes} \\
Micro & \texttt{locator\_\allowbreak{}fallback\_\allowbreak{}rate} & 0.076 & 0.006 & 11.81x & \textbf{yes} \\
Micro & \texttt{action\_\allowbreak{}repeat\_\allowbreak{}frac} & 0.096 & 0.004 & 21.88x & \textbf{yes} \\
Micro & \texttt{finish\_\allowbreak{}rate} & 0.054 & 0.036 & 1.50x & \textbf{yes} \\
Efficiency & \texttt{mean\_\allowbreak{}cost\_\allowbreak{}usd} & 0.008 & 0.002 & 3.24x & \textbf{yes} \\
Efficiency & \texttt{cost\_\allowbreak{}rel\_\allowbreak{}dom} & 0.108 & 0.026 & 4.22x & \textbf{yes} \\
Efficiency & \texttt{mean\_\allowbreak{}latency\_\allowbreak{}s} & 19.562 & 22.488 & 0.87x & no \\
Efficiency & \texttt{mean\_\allowbreak{}latency\_\allowbreak{}canonical\_\allowbreak{}s} & 19.204 & 22.846 & 0.84x & no \\
Efficiency & \texttt{mean\_\allowbreak{}tokens} & 8697.978 & 2186.054 & 3.98x & \textbf{yes} \\
\bottomrule
\end{tabular}
\end{table*}

\section{Efficiency and its estimands}\label{app:efficiency}

Two tables carry the efficiency point. Table~\ref{tab:t11} decomposes a latency figure: the model call is 22--67\% of a step, the rest is browser and container, and stripping the environment share changes which mode is fastest in 4 of 8 cells. Table~\ref{tab:t15} switches the denominator from per-attempt to per-success and the cheapest mode changes in 4 of the 6 cells with enough successes to divide by, with every pairwise interval overlapping. Both support the same statement: an efficiency claim without its estimand is ambiguous, and the estimand is usually left implicit.

\begin{table*}[tp]
\centering
\caption{What a latency number contains. What a latency number contains. Every latency figure elsewhere in this paper is the whole step; \texttt{backend\_infer} isolates the model call and was read by no analysis script until 2026-08-03. The model is 22--67\% of the measured time; the rest is the browser and the container, which \texttt{offsite\_navigation\_audit} measures at 1.69× between the two sites. Removing it changes which mode is fastest in 4 of 8 cells, and not at random: 4 of 5 reddit-family cells flip against 0 of 3 classifieds cells: the flips land where the container is slowest. A sentence naming the fastest mode is therefore partly a sentence about this deployment. What survives estimand choice is only that the two orderings disagree. Source: \texttt{latency\_decomposition.json}.}
\label{tab:t11}
\footnotesize
\setlength{\tabcolsep}{3pt}
\begin{tabular}{@{}
  >{\raggedright\arraybackslash}p{(\linewidth - 12\tabcolsep) * \real{0.1429}}
  >{\raggedright\arraybackslash}p{(\linewidth - 12\tabcolsep) * \real{0.1429}}
  >{\raggedright\arraybackslash}p{(\linewidth - 12\tabcolsep) * \real{0.1429}}
  >{\raggedright\arraybackslash}p{(\linewidth - 12\tabcolsep) * \real{0.1429}}
  >{\raggedright\arraybackslash}p{(\linewidth - 12\tabcolsep) * \real{0.1429}}
  >{\raggedright\arraybackslash}p{(\linewidth - 12\tabcolsep) * \real{0.1429}}
  >{\raggedright\arraybackslash}p{(\linewidth - 12\tabcolsep) * \real{0.1429}}@{}}
\toprule
\begin{minipage}[b]{\linewidth}\raggedright
cell
\end{minipage} & \begin{minipage}[b]{\linewidth}\raggedright
mean step (ms)
\end{minipage} & \begin{minipage}[b]{\linewidth}\raggedright
model call (ms)
\end{minipage} & \begin{minipage}[b]{\linewidth}\raggedright
model share
\end{minipage} & \begin{minipage}[b]{\linewidth}\raggedright
fastest by total
\end{minipage} & \begin{minipage}[b]{\linewidth}\raggedright
fastest by model only
\end{minipage} & \begin{minipage}[b]{\linewidth}\raggedright
same?
\end{minipage} \\
\midrule
B0·classifieds & 7,622 & 2,140 & \textbf{28.1\%} & DOM+\allowbreak{}sprompt & DOM+\allowbreak{}sprompt & yes \\
B0·reddit & 24,601 & 5,603 & \textbf{22.8\%} & Vision & SoM-image & \textbf{no} \\
B0·wa\_\allowbreak{}reddit & 16,112 & 3,551 & \textbf{22.0\%} & Vision & DOM+\allowbreak{}sprompt & \textbf{no} \\
B1·classifieds & 14,180 & 8,213 & \textbf{57.9\%} & Vision & Vision & yes \\
B1·reddit & 24,376 & 7,916 & \textbf{32.5\%} & Vision & DOM+\allowbreak{}stext & \textbf{no} \\
B1·wa\_\allowbreak{}reddit & 21,221 & 7,747 & \textbf{36.5\%} & DOM+\allowbreak{}sprompt & Vision & \textbf{no} \\
B2·classifieds & 14,739 & 9,908 & \textbf{67.2\%} & DOM+\allowbreak{}sprompt & DOM+\allowbreak{}sprompt & yes \\
B2·reddit & 22,863 & 9,131 & \textbf{39.9\%} & Vision & Vision & yes \\
\bottomrule
\end{tabular}
\end{table*}

\begin{table*}[tp]
\centering
\caption{Per-attempt versus per-success. Per-attempt versus per-success denominators. \texttt{content?\ =\ no} marks cells whose best mode has fewer than 10 successes: their ratios are directions at best and both B2 cells are in that state. Among the 6 cells with content the cheapest mode changes under the success denominator in 4 and the fastest in 3. \textbf{Caution:} Every pairwise CI overlaps, so this supports the methodological point that the denominator must be declared, not any ranking. Source: \texttt{outcome\_efficiency.json}.}
\label{tab:t15}
\footnotesize
\setlength{\tabcolsep}{3pt}
\begin{tabular}{@{}
  >{\raggedright\arraybackslash}p{(\linewidth - 12\tabcolsep) * \real{0.1429}}
  >{\raggedright\arraybackslash}p{(\linewidth - 12\tabcolsep) * \real{0.1429}}
  >{\raggedright\arraybackslash}p{(\linewidth - 12\tabcolsep) * \real{0.1429}}
  >{\raggedright\arraybackslash}p{(\linewidth - 12\tabcolsep) * \real{0.1429}}
  >{\raggedright\arraybackslash}p{(\linewidth - 12\tabcolsep) * \real{0.1429}}
  >{\raggedright\arraybackslash}p{(\linewidth - 12\tabcolsep) * \real{0.1429}}
  >{\raggedright\arraybackslash}p{(\linewidth - 12\tabcolsep) * \real{0.1429}}@{}}
\toprule
\begin{minipage}[b]{\linewidth}\raggedright
cell
\end{minipage} & \begin{minipage}[b]{\linewidth}\raggedright
content?
\end{minipage} & \begin{minipage}[b]{\linewidth}\raggedright
cheapest/attempt
\end{minipage} & \begin{minipage}[b]{\linewidth}\raggedright
cheapest/success
\end{minipage} & \begin{minipage}[b]{\linewidth}\raggedright
fastest/attempt
\end{minipage} & \begin{minipage}[b]{\linewidth}\raggedright
fastest/success
\end{minipage} & \begin{minipage}[b]{\linewidth}\raggedright
max solves
\end{minipage} \\
\midrule
cls·B0 & yes & Vision & Vision & SoM & SoM & 61 \\
red·B0 & yes & Vision & DOM & Vision & SoM & 30 \\
cls·B1 & yes & Vision & Vision & SoM & SoM & 32 \\
red·B1 & yes & Vision & SoM & Vision & SoM & 15 \\
cls·B2 & \textbf{no} & Vision & Vision & SoM & SoM & 5 \\
red·B2 & \textbf{no} & Vision & DOM & Vision & DOM & 8 \\
WA·B1 & yes & Vision & DOM+\allowbreak{}stext & DOM & DOM & 17 \\
WA·B0 & yes & DOM & DOM+\allowbreak{}stext & DOM+\allowbreak{}sprompt & DOM+\allowbreak{}stext & 37 \\
\bottomrule
\end{tabular}
\end{table*}

\section{Routing, triage, and cascade}\label{app:routing}

Table~\ref{tab:pb05} is the triage-label predictability that \S\ref{sec:lowerbound} builds on. Table~\ref{tab:t44} asks whether the two obstructions that section reports, too few labels and too few contested tasks, are two walls or one: both sets are subsets of the tasks something solves, and across the eight cells they track the best single mode's success rate at $\rho = 0.952$. Part of that association is structural rather than empirical, and the caption separates the two; what it licenses is the framing of \S\ref{sec:gap}, not a new effect.

\begin{table*}[tp]
\centering
\caption{Predictability of the triage label. AUROC is cross-validated; the comparison column is the best single covariate used alone, so the margin isolates what the multivariate model adds. The reddit · B2 row is below chance and is the subject of \S\ref{sec:lowerbound}.}
\label{tab:pb05}
\footnotesize
\setlength{\tabcolsep}{3pt}
\begin{tabular}{@{}
  >{\raggedright\arraybackslash}p{(\linewidth - 10\tabcolsep) * \real{0.1667}}
  >{\raggedright\arraybackslash}p{(\linewidth - 10\tabcolsep) * \real{0.1667}}
  >{\raggedright\arraybackslash}p{(\linewidth - 10\tabcolsep) * \real{0.1667}}
  >{\raggedright\arraybackslash}p{(\linewidth - 10\tabcolsep) * \real{0.1667}}
  >{\raggedright\arraybackslash}p{(\linewidth - 10\tabcolsep) * \real{0.1667}}
  >{\raggedright\arraybackslash}p{(\linewidth - 10\tabcolsep) * \real{0.1667}}@{}}
\toprule
\begin{minipage}[b]{\linewidth}\raggedright
cell
\end{minipage} & \begin{minipage}[b]{\linewidth}\raggedright
tasks
\end{minipage} & \begin{minipage}[b]{\linewidth}\raggedright
solvable
\end{minipage} & \begin{minipage}[b]{\linewidth}\raggedright
AUROC (logistic)
\end{minipage} & \begin{minipage}[b]{\linewidth}\raggedright
best single covariate
\end{minipage} & \begin{minipage}[b]{\linewidth}\raggedright
margin
\end{minipage} \\
\midrule
classifieds · B0 & 224 & 43.3\% & 0.676 & 0.607 & +0.069 \\
reddit · B0 & 203 & 26.1\% & 0.666 & 0.612 & +0.054 \\
classifieds · B1 & 224 & 24.6\% & 0.717 & 0.627 & +0.090 \\
reddit · B1 & 203 & 11.8\% & 0.685 & 0.637 & +0.048 \\
classifieds · B2 & 224 & 7.1\% & 0.651 & 0.655 & −0.005 \\
reddit · B2 & 203 & 7.4\% & 0.483 & 0.711 & −0.228 \\
\bottomrule
\end{tabular}
\end{table*}

\begin{table*}[tp]
\centering
\caption{Label supply and routing value are one quantity. Label supply and routing value are the same set. The rows a which-mode router can be trained on exist only where something succeeded; the rows where a per-task choice exists at all are those with more than one solver. Both are subsets of the solvable set, and across all 8 cells they track the best single mode's success rate with Spearman rho = 0.952 (exact permutation p = 0.0011), mean \textbar best SR - routable share\textbar{} = 1.65pp. So the two obstructions the lower bound reports are not independent walls but one wall, whose height is set by how much the agent can do. \textbf{Caution: part of this association is structural, not empirical}: both columns are functionals of one solve matrix. The non-structural content is the near-linearity: under mode independence the routable share would grow near-quadratically in the per-mode rate at these success levels, and it does not, which says task difficulty dominates mode-task matching. \textbf{Caution: the ratio is not constant}: 0.53-0.71 in the 6 cells above the floor, but half that (\texttt{red\_B2}, \texttt{cls\_B2}) on 3 and 4 routable tasks, where it is one task wide. \textbf{Caution: the cells are not independent}: they share three sites and three backbones, so the permutation null asks whether the pairing is arbitrary, not whether eight systems were sampled. Source: \texttt{supply\_value\_coupling.json}.}
\label{tab:t44}
\footnotesize
\setlength{\tabcolsep}{3pt}
\begin{tabular}{@{}
  >{\raggedright\arraybackslash}p{(\linewidth - 12\tabcolsep) * \real{0.1429}}
  >{\raggedright\arraybackslash}p{(\linewidth - 12\tabcolsep) * \real{0.1429}}
  >{\raggedright\arraybackslash}p{(\linewidth - 12\tabcolsep) * \real{0.1429}}
  >{\raggedright\arraybackslash}p{(\linewidth - 12\tabcolsep) * \real{0.1429}}
  >{\raggedright\arraybackslash}p{(\linewidth - 12\tabcolsep) * \real{0.1429}}
  >{\raggedright\arraybackslash}p{(\linewidth - 12\tabcolsep) * \real{0.1429}}
  >{\raggedright\arraybackslash}p{(\linewidth - 12\tabcolsep) * \real{0.1429}}@{}}
\toprule
\begin{minipage}[b]{\linewidth}\raggedright
cell
\end{minipage} & \begin{minipage}[b]{\linewidth}\raggedright
n
\end{minipage} & \begin{minipage}[b]{\linewidth}\raggedright
best single mode
\end{minipage} & \begin{minipage}[b]{\linewidth}\raggedright
best SR
\end{minipage} & \begin{minipage}[b]{\linewidth}\raggedright
routable set (\textgreater1 solver)
\end{minipage} & \begin{minipage}[b]{\linewidth}\raggedright
solvable (any mode)
\end{minipage} & \begin{minipage}[b]{\linewidth}\raggedright
routable / solvable
\end{minipage} \\
\midrule
\texttt{wa\_red\_B0} & 104 & DOM+stext & 35.58\% & 34.6\% (36) & 51.9\% & 0.67 \\
\texttt{cls\_B0} & 224 & SoM & 27.23\% & 30.4\% (68) & 43.3\% & 0.70 \\
\texttt{wa\_red\_B1} & 104 & DOM & 16.35\% & 17.3\% (18) & 30.8\% & 0.56 \\
\texttt{red\_B0} & 203 & SoM & 14.78\% & 17.7\% (36) & 26.1\% & 0.68 \\
\texttt{cls\_B1} & 224 & SoM & 14.29\% & 12.9\% (29) & 24.6\% & 0.53 \\
\texttt{red\_B1} & 203 & SoM & 7.39\% & 8.4\% (17) & 11.8\% & 0.71 \\
\texttt{red\_B2} & 203 & DOM & 3.94\% & 1.5\% (3) & 7.4\% & 0.20 \\
\texttt{cls\_B2} & 224 & SoM & 2.23\% & 1.8\% (4) & 7.1\% & 0.25 \\
\bottomrule
\end{tabular}
\end{table*}


Table~\ref{tab:t19} turns the ex-ante partition (Table~\ref{tab:t17}) into a policy. Table~\ref{tab:t20}'s cascade beats always-rich at no operating point in any comparable cell, and Table~\ref{tab:t21} is the table that says what that means: against the \emph{same budget spent at random} the confidence ranking wins nearly everywhere, so the signal is informative and still not enough. Table~\ref{tab:t22} adds the benchmark's own difficulty annotation for a mean $\Delta$AUROC of +0.0024. Table~\ref{tab:t23} shows the feature a practitioner would reach for first looks like it should help and does not. Table~\ref{tab:t33} pools the backbones and routes by cost tier instead of by task.

\begin{figure*}[tp]
\centering
\includegraphics[width=\columnwidth]{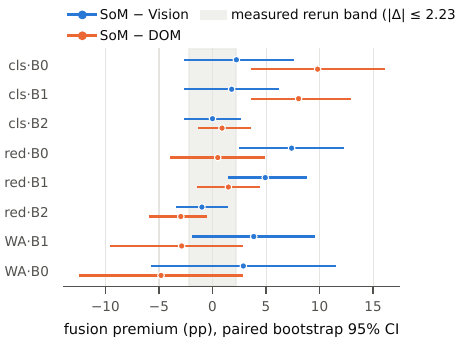}
\caption{Fusion premium per cell (Appendix Table~\ref{tab:t16}), paired bootstrap 95\% CIs. The shaded region is the rerun threshold of \S\ref{sec:noise}: 3.8--4.2pp, the level a mean difference must reach before a single repetition of one arm would be unlikely to produce it. Against the workload-matched channel no cell clears it (the largest such premium is +2.23pp); SoM's decisive wins are all against the channel mismatched to that workload.}
\label{fig:fusion}
\end{figure*}

The bound the rest of this section does not reach is Table~\ref{tab:t42} in \S\ref{sec:upperbound}. Its paper-B precursors (the oracle stated against the best single mode, and its triage/route decomposition) are below, followed by the rerun comparator (Table~\ref{tab:t18}), the per-cell fusion-premium intervals behind Figure~\ref{fig:fusion} (Table~\ref{tab:t16}), the which-mode label supply (Table~\ref{tab:pb03}), and the per-cell partition intervals behind Figure~\ref{fig:partition} (Table~\ref{tab:t17}).

\begin{table*}[tbp]
\centering
\caption{What a perfect per-task choice could buy. Two ceilings, and only one of them survives its own control. \emph{Ceiling: any mode solves it} is what a perfect per-task choice could reach; it runs 7.1-51.9\% against a best single mode of 2.2-35.6\%. \textbf{Caution:} That column is a six-arm union against a one-arm baseline. The arm-matched comparison is the last two columns: adding the single best distinct arm buys +1.97 to +7.14pp, and rerunning an arm already in hand buys a draw in the same range wherever a replicate exists -- so the headroom cannot be attributed to representation diversity rather than to resampling. The union of \emph{five} reruns has never been measured, so the split at higher arm counts is unknown, not estimated. \emph{Same tasks, lower cost} keeps the best mode everywhere and sends only the tasks no mode solves to the cheapest one: success is unchanged by construction and cost falls 9.5-30.6\% in 8 of 8 cells. That ceiling is immune to the arm-count objection because it adds no arms. Why both are hard to reach: no mode solves 48.1-92.9\% of tasks, and the set where a per-task choice even exists is 1.5-34.6\% of the cell. Leaked successes are kept here, as in every other table; 6 scored successes on reddit were credited without the episode ever visiting the forum the evaluator reads, and zeroing them (denominator unchanged) is reported as a sensitivity analysis rather than folded in: the detection criterion is a lower bound, so it cannot license a corrected point estimate. Both sets of figures are in the product. Source: \texttt{routing\_ceiling.json}}
\label{tab:t42}
\footnotesize
\setlength{\tabcolsep}{4pt}
\begin{tabular}{@{}
  >{\raggedright\arraybackslash}p{(\linewidth - 14\tabcolsep) * \real{0.1250}}
  >{\raggedright\arraybackslash}p{(\linewidth - 14\tabcolsep) * \real{0.1250}}
  >{\raggedright\arraybackslash}p{(\linewidth - 14\tabcolsep) * \real{0.1250}}
  >{\raggedright\arraybackslash}p{(\linewidth - 14\tabcolsep) * \real{0.1250}}
  >{\raggedright\arraybackslash}p{(\linewidth - 14\tabcolsep) * \real{0.1250}}
  >{\raggedright\arraybackslash}p{(\linewidth - 14\tabcolsep) * \real{0.1250}}
  >{\raggedright\arraybackslash}p{(\linewidth - 14\tabcolsep) * \real{0.1250}}
  >{\raggedright\arraybackslash}p{(\linewidth - 14\tabcolsep) * \real{0.1250}}@{}}
\toprule
\begin{minipage}[b]{\linewidth}\raggedright
cell
\end{minipage} & \begin{minipage}[b]{\linewidth}\raggedright
n
\end{minipage} & \begin{minipage}[b]{\linewidth}\raggedright
best single mode
\end{minipage} & \begin{minipage}[b]{\linewidth}\raggedright
ceiling: any mode solves
\end{minipage} & \begin{minipage}[b]{\linewidth}\raggedright
headroom
\end{minipage} & \begin{minipage}[b]{\linewidth}\raggedright
same tasks, lower cost
\end{minipage} & \begin{minipage}[b]{\linewidth}\raggedright
+1 arm
\end{minipage} & \begin{minipage}[b]{\linewidth}\raggedright
rerun once
\end{minipage} \\
\midrule
WA·B0 & 104 & DOM+\allowbreak{}stext 35.58\% & \textbf{51.92\%} & +16.35pp & \textbf{-10.7\%} & +5.77 & 2.00-4.00 \\
cls·B0 & 224 & SoM 27.23\% & \textbf{43.30\%} & +16.07pp & \textbf{-12.8\%} & +7.14 & 4.91-7.59 \\
WA·B1 & 104 & DOM 16.35\% & \textbf{30.77\%} & +14.42pp & \textbf{-26.8\%} & +4.81 & 2.00-4.00 \\
red·B0 & 203 & SoM 14.78\% & \textbf{26.11\%} & +11.33pp & \textbf{-9.5\%} & +4.93 & -- \\
cls·B1 & 224 & SoM 14.29\% & \textbf{24.55\%} & +10.27pp & \textbf{-19.4\%} & +4.91 & -- \\
red·B1 & 203 & SoM 7.39\% & \textbf{11.82\%} & +4.43pp & \textbf{-30.6\%} & +1.97 & -- \\
red·B2 & 203 & DOM 3.94\% & \textbf{7.39\%} & +3.45pp & \textbf{-26.4\%} & +1.97 & -- \\
cls·B2 & 224 & SoM 2.23\% & \textbf{7.14\%} & +4.91pp & \textbf{-21.3\%} & +2.23 & -- \\
\bottomrule
\end{tabular}
\end{table*}

\begin{table*}[tp]
\centering
\caption{The routing ceiling, over classifieds (cls) and reddit (red). The last three columns give success rate (\%) and mean per-episode billed cost (USD). The oracle picks the cheapest mode that solved each task, so its success rate equals the solvable column by construction; ``cheapest'' is the lowest-mean-cost mode applied everywhere. Cost is never compared across backbones.}
\label{tab:pb01}
\footnotesize
\setlength{\tabcolsep}{3pt}
\begin{tabular}{@{}
  >{\raggedright\arraybackslash}p{(\linewidth - 10\tabcolsep) * \real{0.1667}}
  >{\raggedright\arraybackslash}p{(\linewidth - 10\tabcolsep) * \real{0.1667}}
  >{\raggedright\arraybackslash}p{(\linewidth - 10\tabcolsep) * \real{0.1667}}
  >{\raggedright\arraybackslash}p{(\linewidth - 10\tabcolsep) * \real{0.1667}}
  >{\raggedright\arraybackslash}p{(\linewidth - 10\tabcolsep) * \real{0.1667}}
  >{\raggedright\arraybackslash}p{(\linewidth - 10\tabcolsep) * \real{0.1667}}@{}}
\toprule
\begin{minipage}[b]{\linewidth}\raggedright
cell
\end{minipage} & \begin{minipage}[b]{\linewidth}\raggedright
n
\end{minipage} & \begin{minipage}[b]{\linewidth}\raggedright
solvable
\end{minipage} & \begin{minipage}[b]{\linewidth}\raggedright
best single
\end{minipage} & \begin{minipage}[b]{\linewidth}\raggedright
cheapest
\end{minipage} & \begin{minipage}[b]{\linewidth}\raggedright
oracle
\end{minipage} \\
\midrule
cls · B0 & 224 & 43.3\% & 27.23 / 0.07236 & 25.00 / 0.06481 & 43.30 / 0.05777 \\
red · B0 & 203 & 26.1\% & 14.78 / 0.11045 & 7.39 / 0.09807 & 26.11 / 0.09534 \\
cls · B1 & 224 & 24.6\% & 14.29 / 0.06028 & 12.50 / 0.04316 & 24.55 / 0.04171 \\
red · B1 & 203 & 11.8\% & 7.39 / 0.08000 & 2.46 / 0.05240 & 11.82 / 0.05178 \\
cls · B2 & 224 & 7.1\% & 2.23 / 0.09075 & 2.23 / 0.07065 & 7.14 / 0.06953 \\
red · B2 & 203 & 7.4\% & 3.94 / 0.09479 & 1.97 / 0.06833 & 7.39 / 0.06958 \\
\bottomrule
\end{tabular}
\end{table*}

\begin{table*}[tp]
\centering
\caption{The ceiling decomposed. Both columns are deltas against the best single mode. The triage half is accuracy-neutral by construction and carries 9.5--30.6\% of cost saving; the route half carries 3.45--16.35pp of success rate and much less cost. The two WebArena rows were added 2026-08-04 and are the favourable end of both halves: \texttt{WA$\cdot$B0} carries the largest route-half headroom (+16.35pp) and the largest route-half cost saving (−16.2\%) anywhere in the study, so the six-cell version understated the ceiling on the cells where routing has most to work with. The two halves require different labels, and the rest of the paper fails on one obstruction per half.}
\label{tab:pb02}
\footnotesize
\setlength{\tabcolsep}{3pt}
\begin{tabular}{@{}lll@{}}
\toprule
cell & triage half: ΔSR / Δcost & route half: ΔSR / Δcost \\
\midrule
classifieds · B0 & ±0.00pp / −12.8\% & +16.07pp / −7.4\% \\
reddit · B0 & ±0.00pp / −9.5\% & +11.33pp / −4.2\% \\
classifieds · B1 & ±0.00pp / −19.4\% & +10.27pp / −11.4\% \\
reddit · B1 & ±0.00pp / −30.6\% & +4.43pp / −4.7\% \\
classifieds · B2 & ±0.00pp / −21.3\% & +4.91pp / −2.1\% \\
reddit · B2 & ±0.00pp / −26.4\% & +3.45pp / −0.2\% \\
WA reddit · B0 & ±0.00pp / −10.7\% & +16.35pp / −16.2\% \\
WA reddit · B1 & ±0.00pp / −26.8\% & +14.42pp / −6.7\% \\
\bottomrule
\end{tabular}
\end{table*}

\begin{table*}[tp]
\centering
\caption{New representation versus a rerun. Is a new representation worth more than a rerun? Both middle columns are the same functional at the same arm count (\texttt{\textbar{}\{added\}\ ∖\ \{baseline\}\textbar{}\ /\ n}) so they are directly comparable; only the \emph{source} of the extra arm differs. The band is 3 rerun pairs on one cell (\texttt{B0\ ×\ classifieds}, n=224), one each for DOM, SoM, Vision: so the rows without a band have no comparator at all, and the band itself is 3 draws rather than a bound. Since the SoM replicate landed 2026-08-03 the band is no longer extrapolated onto an unreplicated arm: both the fused mode this table's best-single column keeps selecting and the arm the comparison adds now carry their own measured floor, and adding the third pair left the band unmoved. Source: \texttt{noise\_floor\_inventory.json}}
\label{tab:t18}
\footnotesize
\setlength{\tabcolsep}{3pt}
\begin{tabular}{@{}
  >{\raggedright\arraybackslash}p{(\linewidth - 8\tabcolsep) * \real{0.2000}}
  >{\raggedright\arraybackslash}p{(\linewidth - 8\tabcolsep) * \real{0.2000}}
  >{\raggedright\arraybackslash}p{(\linewidth - 8\tabcolsep) * \real{0.2000}}
  >{\raggedright\arraybackslash}p{(\linewidth - 8\tabcolsep) * \real{0.2000}}
  >{\raggedright\arraybackslash}p{(\linewidth - 8\tabcolsep) * \real{0.2000}}@{}}
\toprule
\begin{minipage}[b]{\linewidth}\raggedright
cell
\end{minipage} & \begin{minipage}[b]{\linewidth}\raggedright
best single
\end{minipage} & \begin{minipage}[b]{\linewidth}\raggedright
+1 distinct arm
\end{minipage} & \begin{minipage}[b]{\linewidth}\raggedright
+1 rerun (measured floor)
\end{minipage} & \begin{minipage}[b]{\linewidth}\raggedright
verdict
\end{minipage} \\
\midrule
cls·B0 & SoM at 27.23 & +7.14 (DOM) & 4.91 -- 7.59pp & \textbf{inside the rerun band} \\
cls·B1 & SoM at 14.29 & +4.91 (Vision) & --- & no floor on this cell \\
cls·B2 & SoM at 2.23 & +2.23 (Vision) & --- & no floor on this cell \\
red·B0 & SoM at 14.78 & +4.93 (DOM) & --- & no floor on this cell \\
red·B1 & SoM at 7.39 & +1.97 (DOM+\allowbreak{}sprompt) & --- & no floor on this cell \\
red·B2 & DOM at 3.94 & +1.97 (Vision) & --- & no floor on this cell \\
WA·B0 & DOM+\allowbreak{}stext at 35.58 & +5.77 (DOM) & --- & no floor on this cell \\
WA·B1 & DOM at 16.35 & +4.81 (DOM+\allowbreak{}stext) & 0.00 -- 10.00pp \emph{(pooled would read 2.00--4.00)} & \textbf{inside the rerun band} \\
\bottomrule
\end{tabular}
\end{table*}

\begin{table*}[tp]
\centering
\caption{Fusion premium against the rerun band. Fusion premium (pp). Comparators are fixed a priori, not per-cell maxima. Paired bootstrap over tasks, 10,000 resamples. Read against the measured rerun band 0.89--2.23pp, not against zero: a premium must beat what repetition delivers for the same money. No cell clears the band; \texttt{cls\_B0}'s +2.23 \emph{equals} its upper edge (both are 5/224). Source: \texttt{fusion\_premium.json}}
\label{tab:t16}
\footnotesize
\setlength{\tabcolsep}{3pt}
\begin{tabular}{@{}llllll@{}}
\toprule
cell & n & SoM − Vision & 95\% CI & SoM − DOM & 95\% CI \\
\midrule
cls·B0 & 224 & +2.23 & {[}-2.68, +7.59{]} & +9.82 & {[}+3.57, +16.07{]} \\
cls·B1 & 224 & +1.79 & {[}-2.68, +6.25{]} & +8.04 & {[}+3.57, +12.95{]} \\
cls·B2 & 224 & +0.00 & {[}-2.68, +2.68{]} & +0.89 & {[}-1.34, +3.57{]} \\
red·B0 & 203 & +7.39 & {[}+2.46, +12.32{]} & +0.49 & {[}-3.94, +4.93{]} \\
red·B1 & 203 & +4.93 & {[}+1.48, +8.87{]} & +1.48 & {[}-1.48, +4.43{]} \\
red·B2 & 203 & -0.99 & {[}-3.45, +1.48{]} & -2.96 & {[}-5.91, -0.49{]} \\
WA·B1 & 104 & +3.85 & {[}-1.92, +9.62{]} & -2.88 & {[}-9.62, +2.88{]} \\
WA·B0 & 104 & +2.88 & {[}-5.77, +11.54{]} & -4.81 & {[}-12.50, +2.88{]} \\
\bottomrule
\end{tabular}
\end{table*}

\begin{table*}[tp]
\centering
\caption{Which-mode label supply. A label exists only where some mode solved the task, so the labelled column equals the number of solved tasks. Six classes have to be discriminated from between 15 and 97 examples.}
\label{tab:pb03}
\footnotesize
\setlength{\tabcolsep}{3pt}
\begin{tabular}{@{}lllll@{}}
\toprule
cell & scored tasks & labelled & solvable & classes present \\
\midrule
classifieds · B0 & 224 & 97 & 43.3\% & 6/6 \\
reddit · B0 & 203 & 53 & 26.1\% & 6/6 \\
classifieds · B1 & 224 & 55 & 24.6\% & 6/6 \\
reddit · B1 & 203 & 24 & 11.8\% & 6/6 \\
classifieds · B2 & 224 & 16 & 7.1\% & 5/6 \\
reddit · B2 & 203 & 15 & 7.4\% & 5/6 \\
\bottomrule
\end{tabular}
\end{table*}

\begin{table*}[tp]
\centering
\caption{Classes surviving a minimum of ten training rows per class in a five-fold split. Fewer than two surviving classes leaves nothing to discriminate between, which is the state of four of the six cells.}
\label{tab:pb04}
\footnotesize
\setlength{\tabcolsep}{3pt}
\begin{tabular}{@{}lllll@{}}
\toprule
cell & labels & classes & classes surviving the filter & trainable \\
\midrule
classifieds · B0 & 97 & 6 & 3 (DOM, DOM+\allowbreak{}sprompt, SoM) & yes \\
reddit · B0 & 53 & 6 & 1 (DOM) & no \\
classifieds · B1 & 55 & 6 & 2 (DOM, SoM) & yes \\
reddit · B1 & 24 & 6 & 0 & no \\
classifieds · B2 & 16 & 5 & 0 & no \\
reddit · B2 & 15 & 5 & 0 & no \\
\bottomrule
\end{tabular}
\end{table*}

\begin{table*}[tp]
\centering
\caption{Ex-ante visual-intent partition. Ex-ante partition. The predicate is a regex over the task intent plus `carries no reference image': both read the task config, so it costs no tokens and needs no episode. On classifieds the screenshot is worth an order of magnitude more on the flagged tasks than on the rest, and the flagged/rest split is significant/not respectively on the two capable backbones. \textbf{Caution:} WebArena is omitted: the predicate fires on only 5 of 104 tasks there and none is solved by any mode, so the cells are degenerate rather than null. Source: \texttt{visual\_intent\_routing.json}.}
\label{tab:t17}
\footnotesize
\setlength{\tabcolsep}{3pt}
\begin{tabular}{@{}
  >{\raggedright\arraybackslash}p{(\linewidth - 12\tabcolsep) * \real{0.1429}}
  >{\raggedright\arraybackslash}p{(\linewidth - 12\tabcolsep) * \real{0.1429}}
  >{\raggedright\arraybackslash}p{(\linewidth - 12\tabcolsep) * \real{0.1429}}
  >{\raggedright\arraybackslash}p{(\linewidth - 12\tabcolsep) * \real{0.1429}}
  >{\raggedright\arraybackslash}p{(\linewidth - 12\tabcolsep) * \real{0.1429}}
  >{\raggedright\arraybackslash}p{(\linewidth - 12\tabcolsep) * \real{0.1429}}
  >{\raggedright\arraybackslash}p{(\linewidth - 12\tabcolsep) * \real{0.1429}}@{}}
\toprule
\begin{minipage}[b]{\linewidth}\raggedright
cell
\end{minipage} & \begin{minipage}[b]{\linewidth}\raggedright
flagged
\end{minipage} & \begin{minipage}[b]{\linewidth}\raggedright
arm
\end{minipage} & \begin{minipage}[b]{\linewidth}\raggedright
Δ vs DOM on flagged
\end{minipage} & \begin{minipage}[b]{\linewidth}\raggedright
95\% CI
\end{minipage} & \begin{minipage}[b]{\linewidth}\raggedright
Δ on the rest
\end{minipage} & \begin{minipage}[b]{\linewidth}\raggedright
95\% CI
\end{minipage} \\
\midrule
cls·B0 & 71 & vision & +22.54 & {[}+9.86, +33.80{]} & +0.65 & {[}-5.88, +7.84{]} \\
cls·B0 & 71 & som & +19.72 & {[}+7.04, +32.39{]} & +5.23 & {[}-1.31, +12.42{]} \\
cls·B1 & 71 & vision & +16.90 & {[}+8.45, +25.35{]} & +1.31 & {[}-3.92, +6.54{]} \\
cls·B1 & 71 & som & +12.68 & {[}+5.63, +21.13{]} & +5.88 & {[}+0.00, +11.76{]} \\
cls·B2 & 71 & vision & +1.41 & {[}-4.23, +7.04{]} & +0.65 & {[}-1.31, +3.27{]} \\
cls·B2 & 71 & som & +0.00 & {[}-5.63, +5.63{]} & +1.31 & {[}-1.31, +3.92{]} \\
red·B0 & 63 & vision & -3.17 & {[}-14.29, +7.94{]} & -8.57 & {[}-14.29, -2.86{]} \\
red·B0 & 63 & som & +0.00 & {[}-9.52, +9.52{]} & +0.71 & {[}-4.29, +5.71{]} \\
red·B1 & 63 & vision & -3.17 & {[}-9.52, +3.17{]} & -3.57 & {[}-7.86, +0.71{]} \\
red·B1 & 63 & som & +1.59 & {[}-3.17, +7.94{]} & +1.43 & {[}-2.14, +5.00{]} \\
red·B2 & 63 & vision & +4.76 & {[}-1.59, +12.70{]} & -5.00 & {[}-8.57, -1.43{]} \\
red·B2 & 63 & som & -1.59 & {[}-4.76, +0.00{]} & -3.57 & {[}-7.14, +0.00{]} \\
\bottomrule
\end{tabular}
\end{table*}

\begin{table*}[tp]
\centering
\caption{Routing policies on the 3-axis frontier. Is routing worth it? \texttt{rule} policies send the ex-ante-flagged tasks to one arm and the rest to another; the partition is a regex over the task intent, so nothing is learned and there is no in-sample optimism. `On frontier' means nothing dominates it, not that it is preferable: on \texttt{cls·B0} all three rule policies sit between always-SoM and always-Vision, worse on every axis than one or the other. Cost/latency are per-attempt cell means, within-cell comparable only. Source: \texttt{rule\_routing\_pareto.json}.}
\label{tab:t19}
\footnotesize
\setlength{\tabcolsep}{3pt}
\begin{tabular}{@{}llllll@{}}
\toprule
cell & policy & SR & cost & latency & on frontier \\
\midrule
cls·B0 & always-SoM & 27.23 & 0.07236 & 106.0s & yes \\
cls·B0 & always-Vision & 25.00 & 0.06481 & 125.2s & yes \\
cls·B0 & rule: flag→Vision else DOM & 24.55 & 0.06809 & 117.6s & yes \\
cls·B0 & rule: flag→SoM else DOM+\allowbreak{}stext & 24.11 & 0.07019 & 115.8s & yes \\
cls·B0 & rule: flag→SoM else DOM & 23.66 & 0.07049 & 111.5s & yes \\
cls·B0 & always-DOM & 17.41 & 0.06962 & 114.1s & no \\
cls·B1 & always-SoM & 14.29 & 0.06028 & 262.0s & yes \\
cls·B1 & always-Vision & 12.50 & 0.04316 & 269.8s & yes \\
cls·B1 & rule: flag→Vision else DOM & 11.61 & 0.05433 & 296.5s & no \\
cls·B1 & rule: flag→SoM else DOM+\allowbreak{}stext & 11.16 & 0.05926 & 297.2s & no \\
cls·B1 & rule: flag→SoM else DOM & 10.27 & 0.05976 & 294.0s & no \\
cls·B1 & always-DOM & 6.25 & 0.05951 & 308.9s & no \\
cls·B2 & always-SoM & 2.23 & 0.09075 & 374.3s & yes \\
cls·B2 & always-Vision & 2.23 & 0.07065 & 417.8s & yes \\
cls·B2 & rule: flag→Vision else DOM & 1.79 & 0.07483 & 407.2s & yes \\
cls·B2 & always-DOM & 1.34 & 0.07676 & 402.3s & yes \\
cls·B2 & rule: flag→SoM else DOM & 1.34 & 0.08120 & 393.4s & no \\
cls·B2 & rule: flag→SoM else DOM+\allowbreak{}stext & 1.34 & 0.07876 & 391.5s & yes \\
red·B0 & always-SoM & 14.78 & 0.11045 & 451.6s & yes \\
red·B0 & always-DOM & 14.29 & 0.10147 & 552.5s & yes \\
red·B0 & rule: flag→SoM else DOM & 14.29 & 0.10426 & 521.2s & yes \\
red·B0 & rule: flag→Vision else DOM & 13.30 & 0.10041 & 510.9s & yes \\
red·B0 & rule: flag→SoM else DOM+\allowbreak{}stext & 13.30 & 0.10722 & 527.8s & no \\
red·B0 & always-Vision & 7.39 & 0.09807 & 418.5s & yes \\
red·B1 & always-SoM & 7.39 & 0.08000 & 609.1s & yes \\
red·B1 & rule: flag→SoM else DOM+\allowbreak{}stext & 6.90 & 0.07275 & 601.8s & yes \\
red·B1 & rule: flag→SoM else DOM & 6.40 & 0.07538 & 604.6s & no \\
red·B1 & always-DOM & 5.91 & 0.07330 & 602.6s & no \\
red·B1 & rule: flag→Vision else DOM & 4.93 & 0.06682 & 557.3s & yes \\
red·B1 & always-Vision & 2.46 & 0.05240 & 456.6s & yes \\
red·B2 & rule: flag→Vision else DOM & 5.42 & 0.08658 & 632.7s & yes \\
red·B2 & always-DOM & 3.94 & 0.09479 & 669.9s & no \\
red·B2 & rule: flag→SoM else DOM & 3.45 & 0.10001 & 655.4s & no \\
red·B2 & always-Vision & 1.97 & 0.06833 & 550.0s & yes \\
red·B2 & rule: flag→SoM else DOM+\allowbreak{}stext & 1.48 & 0.09568 & 660.7s & no \\
red·B2 & always-SoM & 0.99 & 0.11160 & 623.0s & no \\
\bottomrule
\end{tabular}
\end{table*}

\begin{table*}[tp]
\centering
\caption{Confidence-triggered cascade. Confidence-triggered cascade, vision → som. The escalation decision sees only the cheap run's own episode: no outcome, no rich-run information. In 0 of the 6 comparable cells does any operating point beat always-rich on success rate; the cells marked \emph{rich is worse} are excluded because the cascade's premise fails there, and that exclusion rule (\texttt{cheap\_sr\ \textgreater{}=\ rich\_sr}) is outcome-dependent. \textbf{Caution:} The two counts are split for a reason. A point also `wins' by matching always-rich's SR at lower cost, and every such point here is a tie artefact: on \texttt{WA·B1} all of them sit at exactly the rich arm's SR while 44-60 of 104 episodes are tied at the cutoff, so the ranking falls through to task id and the reachable SR spans 8.65-14.42. Counting the two together is what let this table print a WA win for months while the prose said the WA exception was withdrawn: the rows disagreed with the sentence and nobody could see it, because the WA rows were dropped before render until 2026-08-03. \textbf{Caution:} Every number is an offline splice: an escalated task takes its outcome from a standalone rich run, whereas a real cascade would start the rich episode after the cheap one had already acted on a stateful site. That sequential outcome is unobserved in this project. Source: \texttt{confidence\_cascade\_with\_wa.json}}
\label{tab:t20}
\footnotesize
\setlength{\tabcolsep}{3pt}
\begin{tabular}{@{}
  >{\raggedright\arraybackslash}p{(\linewidth - 16\tabcolsep) * \real{0.1111}}
  >{\raggedright\arraybackslash}p{(\linewidth - 16\tabcolsep) * \real{0.1111}}
  >{\raggedright\arraybackslash}p{(\linewidth - 16\tabcolsep) * \real{0.1111}}
  >{\raggedright\arraybackslash}p{(\linewidth - 16\tabcolsep) * \real{0.1111}}
  >{\raggedright\arraybackslash}p{(\linewidth - 16\tabcolsep) * \real{0.1111}}
  >{\raggedright\arraybackslash}p{(\linewidth - 16\tabcolsep) * \real{0.1111}}
  >{\raggedright\arraybackslash}p{(\linewidth - 16\tabcolsep) * \real{0.1111}}
  >{\raggedright\arraybackslash}p{(\linewidth - 16\tabcolsep) * \real{0.1111}}
  >{\raggedright\arraybackslash}p{(\linewidth - 16\tabcolsep) * \real{0.1111}}@{}}
\toprule
\begin{minipage}[b]{\linewidth}\raggedright
cell
\end{minipage} & \begin{minipage}[b]{\linewidth}\raggedright
n
\end{minipage} & \begin{minipage}[b]{\linewidth}\raggedright
cheap SR
\end{minipage} & \begin{minipage}[b]{\linewidth}\raggedright
rich SR
\end{minipage} & \begin{minipage}[b]{\linewidth}\raggedright
always-rich cost
\end{minipage} & \begin{minipage}[b]{\linewidth}\raggedright
oracle SR
\end{minipage} & \begin{minipage}[b]{\linewidth}\raggedright
comparable?
\end{minipage} & \begin{minipage}[b]{\linewidth}\raggedright
beats always-rich --- strictly / SR-tied
\end{minipage} & \begin{minipage}[b]{\linewidth}\raggedright
signals dropped
\end{minipage} \\
\midrule
cls·B0 & 224 & 25.00 & 27.23 & 1.116× & 33.93 & yes & 0 / 0 & 2 \\
cls·B1 & 224 & 12.50 & 14.29 & 1.397× & 19.20 & yes & 0 / 0 & 0 \\
cls·B2 & 224 & 2.23 & 2.23 & 1.284× & 4.46 & --- \emph{(rich is worse)} & \textbf{11} / 22 & 0 \\
red·B0 & 203 & 7.39 & 14.78 & 1.126× & 18.23 & yes & 0 / 0 & 0 \\
red·B1 & 203 & 2.46 & 7.39 & 1.527× & 8.37 & yes & 0 / 0 & 2 \\
red·B2 & 203 & 1.97 & 0.99 & 1.633× & 2.96 & --- \emph{(rich is worse)} & \textbf{41} / 5 & 1 \\
WA·B0 & 104 & 19.23 & 22.12 & 1.054× & 29.81 & yes & 0 / 0 & 0 \\
WA·B1 & 104 & 9.62 & 13.46 & 1.778× & 16.35 & yes & 0 / 3 & 1 \\
\bottomrule
\end{tabular}
\end{table*}

\begin{table*}[tp]
\centering
\caption{Cascade signal against a random-escalation control. Is the cascade's signal doing anything? Table~\ref{tab:t20} answers a deployment question: no operating point beats always-rich on success rate: which on its own cannot distinguish a signal that carries nothing from one that carries something insufficient. This is the comparator that separates them: the same escalation budget spent at random. \textbf{Caution:} Read the median column, not the best column. The best margin is positive in 24 of 24 (cell × fraction) combinations, but each of those is a maximum over that cell's 8--10 candidate signals against a constant (the random comparator does not depend on which signal is used) so positivity there is partly arithmetic. Over all 222 (signal × cell × fraction) points, 64.4\% are positive with a median of +0.327pp: that is the unselected statement, and it is the one to quote. On cls·B2, red·B0, WA·B1 the median is negative at one or more fractions while the best is positive: there the apparent win is entirely selection. The right-hand column is how much of the gap to a per-task oracle the best signal recovers, and inherits the same caveat. The offline-splice caveat on Table~\ref{tab:t20} applies unchanged. Source: \texttt{confidence\_cascade\_with\_wa.json}.}
\label{tab:t21}
\footnotesize
\setlength{\tabcolsep}{3pt}
\begin{tabular}{@{}
  >{\raggedright\arraybackslash}p{(\linewidth - 10\tabcolsep) * \real{0.1667}}
  >{\raggedright\arraybackslash}p{(\linewidth - 10\tabcolsep) * \real{0.1667}}
  >{\raggedright\arraybackslash}p{(\linewidth - 10\tabcolsep) * \real{0.1667}}
  >{\raggedright\arraybackslash}p{(\linewidth - 10\tabcolsep) * \real{0.1667}}
  >{\raggedright\arraybackslash}p{(\linewidth - 10\tabcolsep) * \real{0.1667}}
  >{\raggedright\arraybackslash}p{(\linewidth - 10\tabcolsep) * \real{0.1667}}@{}}
\toprule
\begin{minipage}[b]{\linewidth}\raggedright
cell
\end{minipage} & \begin{minipage}[b]{\linewidth}\raggedright
signals
\end{minipage} & \begin{minipage}[b]{\linewidth}\raggedright
\textbf{best} margin (10/20/30\%)
\end{minipage} & \begin{minipage}[b]{\linewidth}\raggedright
\textbf{median over all signals} (10/20/30\%)
\end{minipage} & \begin{minipage}[b]{\linewidth}\raggedright
signals \textgreater0
\end{minipage} & \begin{minipage}[b]{\linewidth}\raggedright
oracle headroom captured (10/20/30\%)
\end{minipage} \\
\midrule
cls·B0 & 8 & +1.12 / +1.78 / +2.01 & +1.12 / +1.11 / +0.67 & 19/24 & 15\% / 25\% / 30\% \\
cls·B1 & 10 & +1.16 / +1.43 / +2.14 & +0.27 / +0.53 / +0.81 & 19/30 & 20\% / 27\% / 40\% \\
cls·B2 & 10 & +0.45 / +0.89 / +0.45 & +0.00 / +0.00 / +0.00 & 11/30 & 20\% / 40\% / 20\% \\
red·B0 & 10 & +1.74 / +1.46 / +0.74 & +0.01 / +0.48 / -0.25 & 15/30 & 23\% / 27\% / 27\% \\
red·B1 & 8 & +1.49 / +1.96 / +1.48 & +0.50 / +1.22 / +1.23 & 23/24 & 33\% / 50\% / 50\% \\
red·B2 & 9 & +0.10 / +0.20 / +0.30 & +0.10 / +0.20 / +0.30 & 20/27 & 0\% / 0\% / 0\% \\
WA·B0 & 10 & +3.57 / +3.26 / +3.95 & +1.16 / +1.82 / +1.54 & 24/30 & 36\% / 36\% / 45\% \\
WA·B1 & 9 & +2.51 / +2.11 / +2.70 & -0.37 / -0.78 / -0.18 & 12/27 & 43\% / 43\% / 57\% \\
\bottomrule
\end{tabular}
\end{table*}

\begin{table*}[tp]
\centering
\caption{Triage learnability and the visual-difficulty feature. Can the triage label be predicted, and does the benchmark's own visual-difficulty annotation help? Task-held-out 5-fold CV, L2 logistic regression, seed 42. Mean ΔAUROC = +0.0024 over 6 cells, improving 1: inside fold-split noise. \texttt{solvable} is the positive count: the label exists for every task, unlike the which-mode label. \textbf{Caution:} Six cells, not eight, and the reason is the finding. \texttt{visual\_difficulty} is a VisualWebArena task-config annotation; WebArena's 106 reddit configs carry none of \texttt{visual\_difficulty}, \texttt{reasoning\_difficulty}, \texttt{overall\_difficulty} or \texttt{image}: the field simply is not there, so the contrast cannot be computed rather than computing to zero. That absence is what makes WA the clean test elsewhere: where a router can read the benchmark's own difficulty annotation it looks learnable, and WA is the setting where it cannot. Source: \texttt{visual\_difficulty\_router.json}.}
\label{tab:t22}
\footnotesize
\setlength{\tabcolsep}{3pt}
\begin{tabular}{@{}
  >{\raggedright\arraybackslash}p{(\linewidth - 12\tabcolsep) * \real{0.1429}}
  >{\raggedright\arraybackslash}p{(\linewidth - 12\tabcolsep) * \real{0.1429}}
  >{\raggedright\arraybackslash}p{(\linewidth - 12\tabcolsep) * \real{0.1429}}
  >{\raggedright\arraybackslash}p{(\linewidth - 12\tabcolsep) * \real{0.1429}}
  >{\raggedright\arraybackslash}p{(\linewidth - 12\tabcolsep) * \real{0.1429}}
  >{\raggedright\arraybackslash}p{(\linewidth - 12\tabcolsep) * \real{0.1429}}
  >{\raggedright\arraybackslash}p{(\linewidth - 12\tabcolsep) * \real{0.1429}}@{}}
\toprule
\begin{minipage}[b]{\linewidth}\raggedright
cell
\end{minipage} & \begin{minipage}[b]{\linewidth}\raggedright
n
\end{minipage} & \begin{minipage}[b]{\linewidth}\raggedright
solvable
\end{minipage} & \begin{minipage}[b]{\linewidth}\raggedright
AUROC without
\end{minipage} & \begin{minipage}[b]{\linewidth}\raggedright
AUROC with visual\_\allowbreak{}difficulty
\end{minipage} & \begin{minipage}[b]{\linewidth}\raggedright
Δ
\end{minipage} & \begin{minipage}[b]{\linewidth}\raggedright
visual\_\allowbreak{}difficulty alone
\end{minipage} \\
\midrule
cls·B0 & 224 & 97 & 0.726 & 0.726 & -0.000 & 0.534 \\
cls·B1 & 224 & 55 & 0.732 & 0.726 & -0.006 & 0.553 \\
cls·B2 & 224 & 16 & 0.642 & 0.630 & -0.013 & 0.570 \\
red·B0 & 203 & 53 & 0.780 & 0.776 & -0.005 & 0.630 \\
red·B1 & 203 & 24 & 0.864 & 0.863 & -0.002 & 0.620 \\
red·B2 & 203 & 15 & 0.615 & 0.655 & +0.040 & 0.718 \\
\bottomrule
\end{tabular}
\end{table*}

\begin{table*}[tp]
\centering
\caption{The intuitive routing feature. The intuitive routing feature. A task shipping a reference image ought to route to a mode that can see images: the table shows which mode is actually best on each side of that split. \textbf{Caution:} Reference images are delivered in every mode, so this feature does not separate what it appears to. WebArena cannot arbitrate: its 106 reddit configs carry no \texttt{image} field at all (nor any of the three difficulty annotations) so both of this table's stratifiers are undefined there. Six cells is the whole population for this question, not a coverage gap. Source: \texttt{routing\_feature\_diagnostics.json}.}
\label{tab:t23}
\footnotesize
\setlength{\tabcolsep}{3pt}
\begin{tabular}{@{}
  >{\raggedright\arraybackslash}p{(\linewidth - 12\tabcolsep) * \real{0.1429}}
  >{\raggedright\arraybackslash}p{(\linewidth - 12\tabcolsep) * \real{0.1429}}
  >{\raggedright\arraybackslash}p{(\linewidth - 12\tabcolsep) * \real{0.1429}}
  >{\raggedright\arraybackslash}p{(\linewidth - 12\tabcolsep) * \real{0.1429}}
  >{\raggedright\arraybackslash}p{(\linewidth - 12\tabcolsep) * \real{0.1429}}
  >{\raggedright\arraybackslash}p{(\linewidth - 12\tabcolsep) * \real{0.1429}}
  >{\raggedright\arraybackslash}p{(\linewidth - 12\tabcolsep) * \real{0.1429}}@{}}
\toprule
\begin{minipage}[b]{\linewidth}\raggedright
cell
\end{minipage} & \begin{minipage}[b]{\linewidth}\raggedright
tasks with ref image
\end{minipage} & \begin{minipage}[b]{\linewidth}\raggedright
without
\end{minipage} & \begin{minipage}[b]{\linewidth}\raggedright
best mode WITH image
\end{minipage} & \begin{minipage}[b]{\linewidth}\raggedright
SR
\end{minipage} & \begin{minipage}[b]{\linewidth}\raggedright
best mode WITHOUT
\end{minipage} & \begin{minipage}[b]{\linewidth}\raggedright
SR
\end{minipage} \\
\midrule
cls·B0 & 65 & 159 & SoM & 40.00 & Vision & 22.64 \\
cls·B1 & 65 & 159 & DOM+\allowbreak{}stext & 16.92 & SoM & 13.84 \\
cls·B2 & 65 & 159 & SoM & 4.62 & DOM & 1.89 \\
red·B0 & 79 & 124 & SoM & 26.58 & DOM & 7.26 \\
red·B1 & 79 & 124 & SoM & 13.92 & SoM & 3.23 \\
red·B2 & 79 & 124 & DOM & 8.86 & Vision & 3.23 \\
\bottomrule
\end{tabular}
\end{table*}

\begin{table*}[tp]
\centering
\caption{Pooled tier router. Pooling backbones restores label supply and does not buy a router. Labels are scarce per cell, so this arm pools backbones that share a task and routes by the coarsest label available (which of two cost tiers to spend) giving classifieds 152/reddit 77/wa\_reddit 86 labelled rows against the 15--97 a single cell affords. H-pool NOT supported: the same-family × cost-tier router dominates always-cheapest in 0/6 cells and is dominated by the fixed-mode menu in every cell. The most favourable corner (agreeing backbones, coarse label, highest ceiling) does not change the negative result. Δ SR is not the story: the cost column is. Where the router does raise success it also raises cost by 2.7--55.7\%, so it is buying success at a price rather than routing to a better arm, and it strictly dominates always-cheapest in 0 of 14 arms. \textbf{Caution: non-dominance is not dominance.} The \texttt{non-dominated} column is an admissibility criterion (a policy buying success for more money passes it) and it must never be quoted as a win rate. \textbf{Caution: ‡ marks rows that are identical by construction, not by result}: WebArena has no cross-family backbone, so its all-three pool and its same-family pool are the same pool, reported twice to keep the layout parallel. \textbf{Caution:} the pooled task overlap is 50 tasks on classifieds and 20 on reddit, so every per-cell contrast is underpowered; costs are never compared across backbones (an API bill against an electricity estimate); and this probe is post-hoc and exploratory: it is not the preregistered gate and must not be cited as one. Source: \texttt{router\_pooled\_tier\_learnability.json}.}
\label{tab:t33}
\footnotesize
\setlength{\tabcolsep}{3pt}
\begin{tabular}{@{}
  >{\raggedright\arraybackslash}p{(\linewidth - 12\tabcolsep) * \real{0.1429}}
  >{\raggedright\arraybackslash}p{(\linewidth - 12\tabcolsep) * \real{0.1429}}
  >{\raggedright\arraybackslash}p{(\linewidth - 12\tabcolsep) * \real{0.1429}}
  >{\raggedright\arraybackslash}p{(\linewidth - 12\tabcolsep) * \real{0.1429}}
  >{\raggedright\arraybackslash}p{(\linewidth - 12\tabcolsep) * \real{0.1429}}
  >{\raggedright\arraybackslash}p{(\linewidth - 12\tabcolsep) * \real{0.1429}}
  >{\raggedright\arraybackslash}p{(\linewidth - 12\tabcolsep) * \real{0.1429}}@{}}
\toprule
\begin{minipage}[b]{\linewidth}\raggedright
pool
\end{minipage} & \begin{minipage}[b]{\linewidth}\raggedright
cell
\end{minipage} & \begin{minipage}[b]{\linewidth}\raggedright
router SR
\end{minipage} & \begin{minipage}[b]{\linewidth}\raggedright
cheapest SR
\end{minipage} & \begin{minipage}[b]{\linewidth}\raggedright
Δ SR
\end{minipage} & \begin{minipage}[b]{\linewidth}\raggedright
Δ cost
\end{minipage} & \begin{minipage}[b]{\linewidth}\raggedright
non-dom.
\end{minipage} \\
\midrule
same-family & \texttt{cls·B0} & 25.00\% & 25.00\% & +0.00pp & +13.9\% & 46.8\% \\
same-family & \texttt{cls·B1} & 12.05\% & 12.50\% & -0.45pp & +33.0\% & 33.6\% \\
all-three & \texttt{cls·B0} & 24.11\% & 25.00\% & -0.89pp & +14.6\% & 34.5\% \\
all-three & \texttt{cls·B1} & 11.16\% & 12.50\% & -1.34pp & +36.6\% & 21.2\% \\
all-three & \texttt{cls·B2} & 1.79\% & 2.23\% & -0.45pp & +4.8\% & 18.0\% \\
same-family & \texttt{red·B0} & 14.29\% & 7.39\% & +6.90pp & +10.2\% & 99.2\% \\
same-family & \texttt{red·B1} & 4.43\% & 2.46\% & +1.97pp & +38.2\% & 86.8\% \\
all-three & \texttt{red·B0} & 13.30\% & 7.39\% & +5.91pp & +2.7\% & 98.2\% \\
all-three & \texttt{red·B1} & 4.43\% & 2.46\% & +1.97pp & +35.8\% & 86.8\% \\
all-three & \texttt{red·B2} & 1.97\% & 1.97\% & +0.00pp & +7.4\% & 42.5\% \\
same-family & \texttt{wa\_red·B0} & 30.77\% & 26.92\% & +3.85pp & +6.3\% & 83.3\% \\
same-family & \texttt{wa\_red·B1} & 15.38\% & 9.62\% & +5.77pp & +55.7\% & 94.1\% \\
all-three ‡ & \texttt{wa\_red·B0} & 30.77\% & 26.92\% & +3.85pp & +6.3\% & 83.3\% \\
all-three ‡ & \texttt{wa\_red·B1} & 15.38\% & 9.62\% & +5.77pp & +55.7\% & 94.1\% \\
\bottomrule
\end{tabular}
\end{table*}

\begin{table*}[tp]
\centering
\caption{The nested triage policy against the always-cheapest fixed policy. Pareto dominance requires no worse on both axes; no cell achieves it. The parenthetical gives the router's deltas relative to the fixed policy.}
\label{tab:pb06}
\footnotesize
\setlength{\tabcolsep}{3pt}
\begin{tabular}{@{}
  >{\raggedright\arraybackslash}p{(\linewidth - 6\tabcolsep) * \real{0.2500}}
  >{\raggedright\arraybackslash}p{(\linewidth - 6\tabcolsep) * \real{0.2500}}
  >{\raggedright\arraybackslash}p{(\linewidth - 6\tabcolsep) * \real{0.2500}}
  >{\raggedright\arraybackslash}p{(\linewidth - 6\tabcolsep) * \real{0.2500}}@{}}
\toprule
\begin{minipage}[b]{\linewidth}\raggedright
cell
\end{minipage} & \begin{minipage}[b]{\linewidth}\raggedright
nested SR / cost
\end{minipage} & \begin{minipage}[b]{\linewidth}\raggedright
always-cheapest SR / cost
\end{minipage} & \begin{minipage}[b]{\linewidth}\raggedright
Pareto-dominates?
\end{minipage} \\
\midrule
classifieds · B0 & 26.79\% / 0.07197 & 25.00\% / 0.06481 & no (+1.79pp SR, +11.0\% cost) \\
reddit · B0 & 12.81\% / 0.09836 & 7.39\% / 0.09807 & no (+5.42pp SR, +0.3\% cost) \\
classifieds · B1 & 14.29\% / 0.05757 & 12.50\% / 0.04316 & no (+1.79pp SR, +33.4\% cost) \\
reddit · B1 & 5.91\% / 0.06970 & 2.46\% / 0.05240 & no (+3.45pp SR, +33.0\% cost) \\
classifieds · B2 & 1.34\% / 0.07247 & 2.23\% / 0.07065 & no (−0.89pp SR, +2.6\% cost) \\
reddit · B2 & 3.94\% / 0.06964 & 1.97\% / 0.06833 & no (+1.97pp SR, +1.9\% cost) \\
\bottomrule
\end{tabular}
\end{table*}

\begin{table*}[tp]
\centering
\caption{The five supervision targets against the two requirements a trainable router needs. Each closed route is closed for a different reason, which is what makes the negative result closed rather than provisional. Supply and identifiability are judged on that target's own denominator, which differs between the last two rows.}
\label{tab:pb07}
\footnotesize
\setlength{\tabcolsep}{3pt}
\begin{tabular}{@{}
  >{\raggedright\arraybackslash}p{(\linewidth - 6\tabcolsep) * \real{0.2500}}
  >{\raggedright\arraybackslash}p{(\linewidth - 6\tabcolsep) * \real{0.2500}}
  >{\raggedright\arraybackslash}p{(\linewidth - 6\tabcolsep) * \real{0.2500}}
  >{\raggedright\arraybackslash}p{(\linewidth - 6\tabcolsep) * \real{0.2500}}@{}}
\toprule
\begin{minipage}[b]{\linewidth}\raggedright
route
\end{minipage} & \begin{minipage}[b]{\linewidth}\raggedright
supply
\end{minipage} & \begin{minipage}[b]{\linewidth}\raggedright
identifiability
\end{minipage} & \begin{minipage}[b]{\linewidth}\raggedright
outcome
\end{minipage} \\
\midrule
which-mode, per cell & fails (15--97 labels, 4/6 cells untrainable) & fine & closed by \S\ref{sec:lowerbound} \\
continuous target & would fix & fine & closed: benchmark score is binary \\
pooled which-mode & fixed (260) & fails (56--57\% conflict; in-sample modal agreement 79--84\%) & wrong estimand \\
triage (binary) & fine (203--224, every task) & fine & learnable, and beaten by a fixed policy (\S\ref{sec:lowerbound}) \\
screenshot tier & fine, but only on solved tasks & fine (68--88\% cross-backbone agreement, over solved tasks) & agreement measured; no tier classifier trained \\
\bottomrule
\end{tabular}
\end{table*}

\section{Failure attribution and mechanisms}\label{app:failures}

Table~\ref{tab:t43} locates \emph{when} two modes part on the tasks whose outcome they disagree about: at the first or second step, essentially never late, which is not the shape accumulated drift has. It is markedly weaker on WebArena, so it is reported per site rather than pooled. Table~\ref{tab:t41} is the paired cut: on tasks only one channel solved, how did the other fail? It counts rule hits, so Table~\ref{tab:t26} asks the same question without the rule vocabulary, using six probes read straight off the step records, and finds the losing channel failing \emph{more blandly} there than elsewhere, which is what closes the objection that the residual is an artefact of a VWA-shaped ruleset. Table~\ref{tab:t30} decomposes the DOM$\rightarrow$SoM-image transition into a text axis and a prompt axis, and Table~\ref{tab:t31} asks whether that text axis moves per-step decision quality more than it moves macro action frequencies. Table~\ref{tab:t32} counts hallucinated element references, which are inapplicable to Vision by construction, and the table marks them. \textbf{Caution:} a per-rule frequency is a distribution of symptoms, not of causes. The two largest rows in most cells are risk markers that causal verification did not confirm as death causes.

Each block of Table~\ref{tab:t41} states the largest enrichment among the rules it does \emph{not} list. Read each block against its own baseline column, never across blocks: the text channel is four arms and the image channel two, so the two blocks' task counts are not comparable to each other.

\begin{table*}[tp]
\centering
\caption{Vocabulary-free probes on the text-wins residual. Is the text-wins residual real, or an artefact of the rule vocabulary? Tables 24--25 count rule hits, and the ruleset was discovered on VisualWebArena: so an absent signature there could be a property of the vocabulary rather than of the world. Each probe here is computed from raw step fields and never from a rule hit, over 142 disagreement episodes against 2304 baseline failures. The largest enrichment is 1.15× and 5 of 6 sit \emph{below} 1: on the tasks the text channel uniquely solves, the image channel fails more blandly than it fails elsewhere: it did not arrive, rather than breaking somewhere nameable. \textbf{Caution:} Six candidates chosen by us, so this cannot show that no mechanism exists; what it closes is the specific objection that the residual is an artefact of a VWA-shaped vocabulary. Source: \texttt{conditional\_failure\_attribution.json}}
\label{tab:t26}
\footnotesize
\setlength{\tabcolsep}{3pt}
\begin{tabular}{@{}
  >{\raggedright\arraybackslash}p{(\linewidth - 6\tabcolsep) * \real{0.2500}}
  >{\raggedright\arraybackslash}p{(\linewidth - 6\tabcolsep) * \real{0.2500}}
  >{\raggedright\arraybackslash}p{(\linewidth - 6\tabcolsep) * \real{0.2500}}
  >{\raggedright\arraybackslash}p{(\linewidth - 6\tabcolsep) * \real{0.2500}}@{}}
\toprule
\begin{minipage}[b]{\linewidth}\raggedright
candidate mechanism
\end{minipage} & \begin{minipage}[b]{\linewidth}\raggedright
on the disagreement set
\end{minipage} & \begin{minipage}[b]{\linewidth}\raggedright
that channel's baseline
\end{minipage} & \begin{minipage}[b]{\linewidth}\raggedright
enrichment
\end{minipage} \\
\midrule
finished in five steps or fewer & 0.162 & 0.141 & \textbf{1.15×} \\
any parse failure & 0.092 & 0.109 & \textbf{0.84×} \\
never searched & 0.387 & 0.468 & \textbf{0.83×} \\
ran out of budget (\textgreater=30 steps) & 0.535 & 0.658 & \textbf{0.81×} \\
page unchanged on over half the steps & 0.331 & 0.431 & \textbf{0.77×} \\
action failure on over half the steps & 0.310 & 0.416 & \textbf{0.74×} \\
\bottomrule
\end{tabular}
\end{table*}

\begin{table*}[tbp]
\centering
\caption{Failure modes are asymmetric across channels. The two channels do not fail the same way. On tasks only one channel solved, how did the other fail? Pooled over 8 cells at ruleset v11. Enrichment = hit rate on the disagreement set over that channel's hit rate across all its failures, so about 1x means it failed there the way it fails everywhere. Rules with fewer than 8 pooled conditional hits are omitted, and the largest omitted enrichment is stated per block rather than left to the reader. Block A has named death causes (top 2.31x, 404 losing-channel episodes against 5388 of that channel's failures overall). Block B does not: its top row rests on 8 hits, exactly the reporting floor, and all 8 of them fall in the 2 WebArena cells, none in the other 6; every other rule in that block sits at or below the everywhere-baseline. On the tasks the text channel uniquely solves, the image channel did not break somewhere nameable -- it did not arrive. \textbf{Caution:} TEXT is four arms and IMAGE is two, so the two blocks' task counts are not comparable to each other; read each block against its own baseline column, never across blocks. \textbf{Caution:} a per-rule frequency is a distribution of symptoms, not of causes -- the largest rows in most cells are risk markers, not death causes, and only rules whose docstrings record a causal check are verified as such. Source: \texttt{conditional\_failure\_attribution.json}.}
\label{tab:t41}
\small
\setlength{\tabcolsep}{4pt}
\begin{tabular}{@{}
  >{\raggedright\arraybackslash}p{(\linewidth - 12\tabcolsep) * \real{0.1429}}
  >{\raggedright\arraybackslash}p{(\linewidth - 12\tabcolsep) * \real{0.1429}}
  >{\raggedright\arraybackslash}p{(\linewidth - 12\tabcolsep) * \real{0.1429}}
  >{\raggedright\arraybackslash}p{(\linewidth - 12\tabcolsep) * \real{0.1429}}
  >{\raggedright\arraybackslash}p{(\linewidth - 12\tabcolsep) * \real{0.1429}}
  >{\raggedright\arraybackslash}p{(\linewidth - 12\tabcolsep) * \real{0.1429}}
  >{\raggedright\arraybackslash}p{(\linewidth - 12\tabcolsep) * \real{0.1429}}@{}}
\toprule
\begin{minipage}[b]{\linewidth}\raggedright
\end{minipage} & \begin{minipage}[b]{\linewidth}\raggedright
rule
\end{minipage} & \begin{minipage}[b]{\linewidth}\raggedright
how it failed
\end{minipage} & \begin{minipage}[b]{\linewidth}\raggedright
on disagreement
\end{minipage} & \begin{minipage}[b]{\linewidth}\raggedright
baseline
\end{minipage} & \begin{minipage}[b]{\linewidth}\raggedright
enrichment
\end{minipage} & \begin{minipage}[b]{\linewidth}\raggedright
hits
\end{minipage} \\
\midrule
\textbf{A. image wins} & & \emph{how the text channel failed} & (101 tasks) & & & \\
& \texttt{P27} & gives up when not found & 3.2\% & 1.4\% & \textbf{2.31x} & 13 \\
& \texttt{P17} & click-back oscillation & 15.1\% & 6.7\% & \textbf{2.25x} & 61 \\
& \texttt{P16} & visual task, DOM cannot see & 6.2\% & 2.8\% & \textbf{2.24x} & 25 \\
& \texttt{P43} & page visual, no screenshot & 48.5\% & 29.3\% & \textbf{1.65x} & 196 \\
& & \emph{15 further rules} & & & all \textless= 1.55x & \\
\textbf{B. text wins} & & \emph{how the image channel failed} & (109 tasks) & & & \\
& \texttt{P49} & submit-page anchor misclick & 3.7\% & 1.0\% & \textbf{3.61x} & 8 \\
& \texttt{P17} & click-back oscillation & 4.6\% & 3.9\% & \textbf{1.17x} & 10 \\
& \texttt{P12} & never paginates & 13.8\% & 14.8\% & \textbf{0.93x} & 30 \\
& \texttt{P31} & budget exhausted, unfinished & 49.5\% & 54.2\% & \textbf{0.91x} & 108 \\
& & \emph{6 further rules} & & & all \textless= 0.87x & \\
\bottomrule
\end{tabular}
\end{table*}

\begin{table*}[tbp]
\centering
\caption{Where two representations part company. Where two representations part company. Restricted to the tasks whose outcome the two modes disagree on (exactly one of them succeeds) so these are the trajectories that produced the difference, not all trajectories. \texttt{first\ divergent\ step} is the first step at which the two URL signatures differ. Drift would look different: two runs that separate through accumulated perturbation share a prefix and part late, whereas on VisualWebArena 72--100\% of these tasks have already diverged by step 3 and at most 3.0\% diverge at step 10 or later. \textbf{Caution: this is markedly weaker on WebArena} (50--89\% early, up to 22.2\% late), which is why the table is per site and not pooled. \textbf{Caution:} each row rests on 13--46 tasks, and element ids are deliberately not used as the anchor: they are neither step-invariant nor mode-invariant, so URL signatures are the only stable comparison. Source: \texttt{mechanism\_per\_task.json} E2.}
\label{tab:t43}
\small
\setlength{\tabcolsep}{4pt}
\begin{tabular}{@{}
  >{\raggedright\arraybackslash}p{(\linewidth - 10\tabcolsep) * \real{0.1667}}
  >{\raggedright\arraybackslash}p{(\linewidth - 10\tabcolsep) * \real{0.1667}}
  >{\raggedright\arraybackslash}p{(\linewidth - 10\tabcolsep) * \real{0.1667}}
  >{\raggedright\arraybackslash}p{(\linewidth - 10\tabcolsep) * \real{0.1667}}
  >{\raggedright\arraybackslash}p{(\linewidth - 10\tabcolsep) * \real{0.1667}}
  >{\raggedright\arraybackslash}p{(\linewidth - 10\tabcolsep) * \real{0.1667}}@{}}
\toprule
\begin{minipage}[b]{\linewidth}\raggedright
site
\end{minipage} & \begin{minipage}[b]{\linewidth}\raggedright
contrast
\end{minipage} & \begin{minipage}[b]{\linewidth}\raggedright
tasks
\end{minipage} & \begin{minipage}[b]{\linewidth}\raggedright
median first divergent step
\end{minipage} & \begin{minipage}[b]{\linewidth}\raggedright
diverged by step 3
\end{minipage} & \begin{minipage}[b]{\linewidth}\raggedright
diverged at step 10+
\end{minipage} \\
\midrule
classifieds & DOM vs SoM-image & 22 & 1 & 95.5\% & 0.0\% \\
classifieds & DOM vs DOM+\allowbreak{}sprompt & 33 & 1 & 84.8\% & 3.0\% \\
classifieds & DOM vs DOM+\allowbreak{}stext & 22 & 1 & 95.5\% & 0.0\% \\
classifieds & SoM-image vs SoM & 46 & 1 & 93.5\% & 0.0\% \\
classifieds & DOM+\allowbreak{}sprompt vs SoM-image & 23 & 1 & 91.3\% & 0.0\% \\
classifieds & DOM+\allowbreak{}stext vs SoM-image & 18 & 2 & 72.2\% & 0.0\% \\
reddit & DOM vs SoM-image & 19 & 0 & 100.0\% & 0.0\% \\
reddit & DOM vs DOM+\allowbreak{}sprompt & 18 & 1 & 88.9\% & 0.0\% \\
reddit & DOM vs DOM+\allowbreak{}stext & 16 & 0 & 93.8\% & 0.0\% \\
reddit & SoM-image vs SoM & 20 & 0 & 94.7\% & 0.0\% \\
reddit & DOM+\allowbreak{}sprompt vs SoM-image & 17 & 0 & 100.0\% & 0.0\% \\
reddit & DOM+\allowbreak{}stext vs SoM-image & 19 & 0 & 100.0\% & 0.0\% \\
wa\_reddit & DOM vs SoM-image & 16 & 1 & 75.0\% & 6.2\% \\
wa\_reddit & DOM vs DOM+\allowbreak{}sprompt & 19 & 1 & 89.5\% & 5.3\% \\
wa\_reddit & DOM vs DOM+\allowbreak{}stext & 21 & 1 & 66.7\% & 9.5\% \\
wa\_reddit & SoM-image vs SoM & 19 & 1 & 73.7\% & 0.0\% \\
wa\_reddit & DOM+\allowbreak{}sprompt vs SoM-image & 13 & 0.5 & 83.3\% & 8.3\% \\
wa\_reddit & DOM+\allowbreak{}stext vs SoM-image & 19 & 4 & 50.0\% & 22.2\% \\
\bottomrule
\end{tabular}
\end{table*}

\begin{table*}[tp]
\centering
\caption{2x2 axis decomposition. 2×2 axis decomposition. Effect sizes are Cohen's h (binary metrics) or d\_z (paired continuous), signed right-minus-left. The compound DOM→SoM-image transition decomposes into a text-payload axis and a prompt-style axis; the image axis is SoM-image→SoM. \textbf{Caution:} On mean differences the two decomposition routes agreeing is an algebraic identity, so a zero residual is arithmetic and not evidence about an interaction. \texttt{B2\ ×\ wa\_reddit} is absent because B2 never ran WebArena. Source: \texttt{axis\_effect\_size\_with\_wa.json}.}
\label{tab:t30}
\scriptsize
\setlength{\tabcolsep}{3pt}
\begin{tabular}{@{}llllll@{}}
\toprule
cell & metric & text axis & prompt axis & image axis & DOM→SoM-image \\
\midrule
cls·B0 & search\_loop & -0.011 & -0.045 & -0.225 & -0.056 \\
cls·B0 & type\_frac & -0.169 & +0.049 & +0.030 & -0.140 \\
cls·B0 & scroll\_frac & +0.072 & +0.063 & -0.253 & +0.134 \\
cls·B0 & selfcorr\_count & +0.088 & -0.115 & -0.091 & -0.020 \\
cls·B0 & click\_frac & -0.001 & -0.059 & +0.035 & -0.056 \\
cls·B0 & finish\_rate & -0.070 & -0.039 & +0.119 & -0.109 \\
cls·B0 & n\_steps & +0.023 & +0.039 & -0.246 & +0.066 \\
cls·B0 & action\_repeat\_frac & -0.051 & +0.086 & -0.117 & +0.030 \\
cls·B1 & search\_loop & +0.066 & -0.160 & -0.203 & -0.094 \\
cls·B1 & type\_frac & +0.088 & -0.546 & -0.005 & -0.452 \\
cls·B1 & scroll\_frac & -0.097 & +0.025 & -0.062 & -0.065 \\
cls·B1 & selfcorr\_count & -0.015 & +0.007 & +0.149 & -0.007 \\
cls·B1 & click\_frac & +0.139 & +0.526 & -0.169 & +0.614 \\
cls·B1 & finish\_rate & -0.055 & +0.082 & +0.170 & +0.027 \\
cls·B1 & n\_steps & +0.101 & -0.108 & -0.253 & -0.011 \\
cls·B1 & action\_repeat\_frac & +0.053 & +0.002 & -0.174 & +0.048 \\
cls·B2 & search\_loop & -0.037 & +0.113 & -0.141 & +0.076 \\
cls·B2 & type\_frac & +0.116 & -0.410 & +0.297 & -0.290 \\
cls·B2 & scroll\_frac & +0.154 & -0.112 & +0.003 & +0.026 \\
cls·B2 & selfcorr\_count & +0.100 & -0.111 & -0.105 & -0.024 \\
cls·B2 & click\_frac & -0.476 & +0.485 & -0.097 & +0.011 \\
cls·B2 & finish\_rate & -0.217 & -0.038 & +0.333 & -0.256 \\
cls·B2 & n\_steps & -0.052 & +0.160 & -0.363 & +0.115 \\
cls·B2 & action\_repeat\_frac & -0.093 & +0.159 & -0.067 & +0.064 \\
red·B0 & search\_loop & -0.170 & -0.104 & +0.052 & -0.275 \\
red·B0 & type\_frac & +0.029 & -0.098 & -0.084 & -0.067 \\
red·B0 & scroll\_frac & -0.206 & +0.041 & -0.148 & -0.182 \\
red·B0 & selfcorr\_count & +0.279 & -0.065 & -0.258 & +0.223 \\
red·B0 & click\_frac & -0.057 & +0.053 & +0.081 & -0.009 \\
red·B0 & finish\_rate & -0.299 & +0.102 & +0.160 & -0.200 \\
red·B0 & n\_steps & +0.276 & -0.032 & -0.252 & +0.245 \\
red·B0 & action\_repeat\_frac & +0.226 & -0.056 & -0.033 & +0.202 \\
red·B1 & search\_loop & +0.020 & -0.100 & -0.297 & -0.080 \\
red·B1 & type\_frac & +0.032 & -0.329 & -0.159 & -0.301 \\
red·B1 & scroll\_frac & +0.008 & -0.071 & +0.157 & -0.074 \\
red·B1 & selfcorr\_count & -0.053 & -0.075 & -0.062 & -0.120 \\
red·B1 & click\_frac & +0.019 & +0.326 & -0.016 & +0.370 \\
red·B1 & finish\_rate & -0.224 & +0.012 & +0.223 & -0.212 \\
red·B1 & n\_steps & +0.202 & -0.044 & -0.231 & +0.165 \\
red·B1 & action\_repeat\_frac & +0.133 & +0.108 & +0.020 & +0.249 \\
red·B2 & search\_loop & +0.092 & -0.115 & -0.179 & -0.024 \\
red·B2 & type\_frac & -0.045 & -0.023 & +0.045 & -0.070 \\
red·B2 & scroll\_frac & -0.037 & +0.093 & -0.040 & +0.054 \\
red·B2 & selfcorr\_count & +0.052 & -0.026 & +0.015 & +0.019 \\
red·B2 & click\_frac & -0.193 & +0.143 & -0.017 & -0.047 \\
red·B2 & finish\_rate & -0.059 & -0.044 & +0.382 & -0.103 \\
red·B2 & n\_steps & -0.126 & +0.057 & -0.137 & -0.054 \\
red·B2 & action\_repeat\_frac & -0.157 & -0.010 & +0.121 & -0.173 \\
WA·B0 & search\_loop & -0.042 & -0.155 & +0.238 & -0.197 \\
WA·B0 & type\_frac & +0.476 & -0.197 & -0.031 & +0.336 \\
WA·B0 & scroll\_frac & -0.247 & +0.155 & -0.191 & -0.135 \\
WA·B0 & selfcorr\_count & +0.106 & -0.053 & +0.088 & +0.078 \\
WA·B0 & click\_frac & -0.255 & +0.111 & -0.139 & -0.170 \\
WA·B0 & finish\_rate & -0.339 & +0.176 & +0.000 & -0.163 \\
WA·B0 & n\_steps & +0.260 & -0.084 & -0.128 & +0.200 \\
WA·B0 & action\_repeat\_frac & +0.157 & +0.012 & -0.217 & +0.170 \\
WA·B1 & search\_loop & +0.081 & -0.200 & -0.270 & -0.119 \\
WA·B1 & type\_frac & -0.016 & -0.271 & -0.109 & -0.275 \\
WA·B1 & scroll\_frac & +0.072 & -0.133 & +0.155 & -0.064 \\
WA·B1 & selfcorr\_count & +0.148 & -0.136 & +0.108 & -0.006 \\
WA·B1 & click\_frac & +0.153 & +0.244 & -0.182 & +0.353 \\
WA·B1 & finish\_rate & +0.121 & -0.204 & -0.021 & -0.083 \\
WA·B1 & n\_steps & +0.071 & +0.058 & +0.010 & +0.121 \\
WA·B1 & action\_repeat\_frac & -0.034 & +0.218 & +0.029 & +0.168 \\
\bottomrule
\end{tabular}
\end{table*}

\begin{table*}[tp]
\centering
\caption{Decision quality versus macro frequency. Does the text axis change per-step decisions more than it changes macro action frequencies? Ratio \textgreater1 means yes. Verdict: generalizes (site\_ok = \{`reddit': True, `classifieds': True, `wa\_reddit': True\}). \textbf{Caution:} \texttt{\_site\_ok} passes a site if any backbone clears 1.0, which is a loose bar: \texttt{WA·B0} is 0.97 and the WA site passes on B1's 2.98. Until 2026-08-03 the verdict function named only the two VWA sites literally and did not consult WA at all. Source: \texttt{axis1\_microbehavior\_with\_wa.json}.}
\label{tab:t31}
\footnotesize
\setlength{\tabcolsep}{3pt}
\begin{tabular}{@{}
  >{\raggedright\arraybackslash}p{(\linewidth - 8\tabcolsep) * \real{0.2000}}
  >{\raggedright\arraybackslash}p{(\linewidth - 8\tabcolsep) * \real{0.2000}}
  >{\raggedright\arraybackslash}p{(\linewidth - 8\tabcolsep) * \real{0.2000}}
  >{\raggedright\arraybackslash}p{(\linewidth - 8\tabcolsep) * \real{0.2000}}
  >{\raggedright\arraybackslash}p{(\linewidth - 8\tabcolsep) * \real{0.2000}}@{}}
\toprule
\begin{minipage}[b]{\linewidth}\raggedright
cell
\end{minipage} & \begin{minipage}[b]{\linewidth}\raggedright
decision effect (mean abs)
\end{minipage} & \begin{minipage}[b]{\linewidth}\raggedright
macro effect (mean abs)
\end{minipage} & \begin{minipage}[b]{\linewidth}\raggedright
ratio
\end{minipage} & \begin{minipage}[b]{\linewidth}\raggedright
\textgreater1?
\end{minipage} \\
\midrule
cls·B0 & 0.1530 & 0.0606 & \textbf{2.52} & yes \\
cls·B1 & 0.1024 & 0.0766 & \textbf{1.34} & yes \\
cls·B2 & 0.3159 & 0.1556 & \textbf{2.03} & yes \\
red·B0 & 0.2761 & 0.1926 & \textbf{1.43} & yes \\
red·B1 & 0.2450 & 0.0863 & \textbf{2.84} & yes \\
red·B2 & 0.3877 & 0.0952 & \textbf{4.07} & yes \\
WA·B0 & 0.2284 & 0.2351 & \textbf{0.97} & \textbf{no} \\
WA·B1 & 0.2588 & 0.0869 & \textbf{2.98} & yes \\
\bottomrule
\end{tabular}
\end{table*}

\begin{table*}[tp]
\centering
\caption{Hallucinated element references. Hallucinated element references, ruleset \texttt{11-intent-text-fallback} over 36 conditions. An action naming an element id that is not in the observation. \textbf{Caution:} \texttt{vision} carries no element-id list at all, so this rule is structurally inapplicable there rather than measuring zero: the same gate-versus-measurement confusion flagged for P2/P4. Source: \texttt{cross\_mode\_failure\_signatures.json}.}
\label{tab:t32}
\footnotesize
\setlength{\tabcolsep}{3pt}
\begin{tabular}{@{}
  >{\raggedright\arraybackslash}p{(\linewidth - 10\tabcolsep) * \real{0.1667}}
  >{\raggedright\arraybackslash}p{(\linewidth - 10\tabcolsep) * \real{0.1667}}
  >{\raggedright\arraybackslash}p{(\linewidth - 10\tabcolsep) * \real{0.1667}}
  >{\raggedright\arraybackslash}p{(\linewidth - 10\tabcolsep) * \real{0.1667}}
  >{\raggedright\arraybackslash}p{(\linewidth - 10\tabcolsep) * \real{0.1667}}
  >{\raggedright\arraybackslash}p{(\linewidth - 10\tabcolsep) * \real{0.1667}}@{}}
\toprule
\begin{minipage}[b]{\linewidth}\raggedright
cell
\end{minipage} & \begin{minipage}[b]{\linewidth}\raggedright
mode
\end{minipage} & \begin{minipage}[b]{\linewidth}\raggedright
episodes
\end{minipage} & \begin{minipage}[b]{\linewidth}\raggedright
failed
\end{minipage} & \begin{minipage}[b]{\linewidth}\raggedright
with hallucinated ref
\end{minipage} & \begin{minipage}[b]{\linewidth}\raggedright
rate of failed
\end{minipage} \\
\midrule
classifieds·B0 & DOM & 224 & 185 & 15 & 8.1\% \\
classifieds·B0 & DOM+\allowbreak{}sprompt & 224 & 180 & 11 & 6.1\% \\
classifieds·B0 & SoM-image & 224 & 189 & 2 & 1.1\% \\
classifieds·B0 & DOM+\allowbreak{}stext & 224 & 189 & 9 & 4.8\% \\
classifieds·B0 & SoM & 187 & 129 & 2 & 1.6\% \\
classifieds·B0 & Vision & 224 & 168 & 0 & 0.0\% \\
reddit·B0 & DOM & 203 & 174 & 8 & 4.6\% \\
reddit·B0 & DOM+\allowbreak{}sprompt & 203 & 178 & 8 & 4.5\% \\
reddit·B0 & SoM-image & 203 & 181 & 1 & 0.6\% \\
reddit·B0 & DOM+\allowbreak{}stext & 203 & 176 & 6 & 3.4\% \\
reddit·B0 & SoM & 203 & 173 & 2 & 1.2\% \\
reddit·B0 & Vision & 203 & 188 & 0 & 0.0\% \\
classifieds·B1 & DOM & 224 & 210 & 20 & 9.5\% \\
classifieds·B1 & DOM+\allowbreak{}sprompt & 224 & 209 & 33 & 15.8\% \\
classifieds·B1 & SoM-image & 224 & 209 & 7 & 3.3\% \\
classifieds·B1 & DOM+\allowbreak{}stext & 224 & 207 & 45 & 21.7\% \\
classifieds·B1 & SoM & 224 & 192 & 0 & 0.0\% \\
classifieds·B1 & Vision & 224 & 196 & 0 & 0.0\% \\
reddit·B1 & DOM & 203 & 191 & 25 & 13.1\% \\
reddit·B1 & DOM+\allowbreak{}sprompt & 203 & 192 & 33 & 17.2\% \\
reddit·B1 & SoM-image & 203 & 191 & 3 & 1.6\% \\
reddit·B1 & DOM+\allowbreak{}stext & 203 & 191 & 24 & 12.6\% \\
reddit·B1 & SoM & 203 & 188 & 5 & 2.7\% \\
reddit·B1 & Vision & 203 & 198 & 0 & 0.0\% \\
classifieds·B2 & DOM & 224 & 221 & 82 & 37.1\% \\
classifieds·B2 & DOM+\allowbreak{}sprompt & 224 & 220 & 114 & 51.8\% \\
classifieds·B2 & SoM-image & 224 & 222 & 43 & 19.4\% \\
classifieds·B2 & DOM+\allowbreak{}stext & 224 & 223 & 75 & 33.6\% \\
classifieds·B2 & SoM & 224 & 219 & 39 & 17.8\% \\
classifieds·B2 & Vision & 224 & 219 & 0 & 0.0\% \\
reddit·B2 & DOM & 203 & 195 & 127 & 65.1\% \\
reddit·B2 & DOM+\allowbreak{}sprompt & 203 & 203 & 155 & 76.4\% \\
reddit·B2 & SoM-image & 203 & 202 & 62 & 30.7\% \\
reddit·B2 & DOM+\allowbreak{}stext & 203 & 199 & 51 & 25.6\% \\
reddit·B2 & SoM & 203 & 201 & 59 & 29.4\% \\
reddit·B2 & Vision & 203 & 199 & 0 & 0.0\% \\
\bottomrule
\end{tabular}
\end{table*}

\section{Audits and provenance}\label{app:audits}

The dispatch-path audit (Table~\ref{tab:t13}), the off-site navigation and container-latency audit (Table~\ref{tab:t29}), the page-change detector correction (Table~\ref{tab:t34}), the evaluator granularity check (Table~\ref{tab:t35}), and the full earned-versus-leaked audit including the first WebArena pass (Table~\ref{tab:t40}).

\begin{table*}[tp]
\centering
\caption{What actually delivered the click. How each action reached the browser. \texttt{Vision} is on the coordinate path by construction (it emits no element ids) so its action success is capped by this harness's coordinate implementation (39\%) rather than by the 89\% the element-id path achieves. That is not a confound to remove (it is what screenshot-only \emph{is}), but the Vision arm measures our grounding code as much as the representation. Separately the element-id fallback share rises with backbone weakness: B0 12\% · B1 35\% · B2 37\% on the text arms: \emph{how often} a run falls back is a model property, the fallback's own 16\% success is ours. No success rate elsewhere is adjusted by this; it bounds external validity. Source: \texttt{dispatch\_path\_audit.json}}
\label{tab:t13}
\footnotesize
\setlength{\tabcolsep}{3pt}
\begin{tabular}{@{}lll@{}}
\toprule
delivery path & actions & action success \\
\midrule
id locator & 8,857 & \textbf{88.9\%} \\
other & 1,530 & \textbf{65.9\%} \\
coord & 2,138 & \textbf{38.6\%} \\
id framework & 4,564 & \textbf{16.1\%} \\
\bottomrule
\end{tabular}
\end{table*}

\begin{table*}[tp]
\centering
\caption{Off-site navigation and container latency. Off-site navigation. Postmill is a link aggregator, so an agent opening a trending thread can walk onto the live public internet; classifieds is self-contained. \textbf{Caution:} Off-site steps are faster, not slower (commercial CDNs beat a Postmill container sharing a host with the agent) so the distortion runs opposite to the intuition. The larger asymmetry is in the last two columns of the on-site medians: reddit's container costs \textasciitilde1.69× what classifieds' does before any agent behaviour enters, which is why no between-site latency number is quotable bare. Source: \texttt{offsite\_navigation\_audit.json}.}
\label{tab:t29}
\footnotesize
\setlength{\tabcolsep}{3pt}
\begin{tabular}{@{}
  >{\raggedright\arraybackslash}p{(\linewidth - 10\tabcolsep) * \real{0.1667}}
  >{\raggedright\arraybackslash}p{(\linewidth - 10\tabcolsep) * \real{0.1667}}
  >{\raggedright\arraybackslash}p{(\linewidth - 10\tabcolsep) * \real{0.1667}}
  >{\raggedright\arraybackslash}p{(\linewidth - 10\tabcolsep) * \real{0.1667}}
  >{\raggedright\arraybackslash}p{(\linewidth - 10\tabcolsep) * \real{0.1667}}
  >{\raggedright\arraybackslash}p{(\linewidth - 10\tabcolsep) * \real{0.1667}}@{}}
\toprule
\begin{minipage}[b]{\linewidth}\raggedright
cell
\end{minipage} & \begin{minipage}[b]{\linewidth}\raggedright
off-site steps
\end{minipage} & \begin{minipage}[b]{\linewidth}\raggedright
off-site episodes
\end{minipage} & \begin{minipage}[b]{\linewidth}\raggedright
median env\_\allowbreak{}step on-site
\end{minipage} & \begin{minipage}[b]{\linewidth}\raggedright
off-site
\end{minipage} & \begin{minipage}[b]{\linewidth}\raggedright
ratio
\end{minipage} \\
\midrule
B0·VWA-cla & 0/20646 (0.00\%) & 0/1344 (0.0\%) & 4,493 ms & --- & --- \\
B0·VWA-red & 478/26425 (1.81\%) & 36/1230 (2.9\%) & 11,349 ms & 10,007 ms & 0.88× \\
B1·VWA-cla & 0/27927 (0.00\%) & 0/1344 (0.0\%) & 5,750 ms & --- & --- \\
B1·VWA-red & 501/29309 (1.71\%) & 54/1230 (4.4\%) & 7,848 ms & 6,349 ms & 0.81× \\
B2·VWA-cla & 58/36529 (0.16\%) & 4/1344 (0.3\%) & 4,656 ms & 14,884 ms & 3.20× \\
B2·VWA-red & 719/33749 (2.13\%) & 49/1230 (4.0\%) & 9,324 ms & 4,850 ms & 0.52× \\
B1·WA-red & 278/14695 (1.89\%) & 40/624 (6.4\%) & 6,809 ms & 6,105 ms & 0.90× \\
B0·WA-red & 123/11703 (1.05\%) & 21/624 (3.4\%) & 6,613 ms & 11,405 ms & 1.72× \\
\bottomrule
\end{tabular}
\end{table*}

\begin{table*}[tp]
\centering
\caption{page\_changed false positives. \texttt{page\_changed} false positives. A step can register a page change that is purely cosmetic; correcting for it raises every mode's no-change rate. The Micro conclusion is unaffected (Vision remains highest in every cell either way) but a router firing on a 2-step no-change streak would trigger 5321 → 5910 times (+11.1\%). Source: \texttt{page\_change\_corrected.json}.}
\label{tab:t34}
\footnotesize
\setlength{\tabcolsep}{3pt}
\begin{tabular}{@{}
  >{\raggedright\arraybackslash}p{(\linewidth - 8\tabcolsep) * \real{0.2000}}
  >{\raggedright\arraybackslash}p{(\linewidth - 8\tabcolsep) * \real{0.2000}}
  >{\raggedright\arraybackslash}p{(\linewidth - 8\tabcolsep) * \real{0.2000}}
  >{\raggedright\arraybackslash}p{(\linewidth - 8\tabcolsep) * \real{0.2000}}
  >{\raggedright\arraybackslash}p{(\linewidth - 8\tabcolsep) * \real{0.2000}}@{}}
\toprule
\begin{minipage}[b]{\linewidth}\raggedright
mode
\end{minipage} & \begin{minipage}[b]{\linewidth}\raggedright
no-change rate observed
\end{minipage} & \begin{minipage}[b]{\linewidth}\raggedright
corrected
\end{minipage} & \begin{minipage}[b]{\linewidth}\raggedright
cosmetic FP steps
\end{minipage} & \begin{minipage}[b]{\linewidth}\raggedright
steps
\end{minipage} \\
\midrule
DOM & 0.3846 & 0.4332 & 1618 & 33277 \\
phantom\_\allowbreak{}prompt & 0.4222 & 0.4681 & 1528 & 33283 \\
phantom\_\allowbreak{}som & 0.3683 & 0.4169 & 1688 & 34741 \\
phantom\_\allowbreak{}text & 0.3331 & 0.3770 & 1518 & 34610 \\
SoM & 0.4071 & 0.4675 & 1870 & 30921 \\
Vision & 0.5675 & 0.6134 & 1569 & 34151 \\
\bottomrule
\end{tabular}
\end{table*}

\begin{table*}[tp]
\centering
\caption{Evaluator granularity. Evaluator granularity over the paper-grade set. Across 7,686 scored episodes in 36 conditions, the evaluator emits 2 distinct values (0.0, 1.0) (645 of them positive) with 0 missing and 0 non-numeric. There is no graded quality target to regress on. That is a property of the benchmark's design rather than of this pipeline, and it is a precondition of every routing negative in this document: a router's training signal can only be as fine-grained as the evaluator, so a partially-completed task is indistinguishable from one the agent never started. Source: \texttt{evaluator\_score\_granularity.json}.}
\label{tab:t35}
\footnotesize
\setlength{\tabcolsep}{3pt}
\begin{tabular}{@{}ll@{}}
\toprule
quantity & value \\
\midrule
conditions resolved & 36 \\
conditions unresolved & 0 \\
episodes scored & 7686 \\
episodes with no score & 0 \\
episodes with a non-numeric score & 0 \\
distinct score values & {[}0.0, 1.0{]} \\
count per value & \{``0.0'': 7041, ``1.0'': 645\} \\
\bottomrule
\end{tabular}
\end{table*}

\begin{table*}[tp]
\centering
\caption{Leaked-success sensitivity. Sensitivity to environmentally-credited successes. \texttt{require\_reset} is a no-op on reddit, so subscriptions accumulate across a run's episodes and a later task can be scored on state an earlier one created. 6 such successes are set to 0 here: the denominator is unchanged, because an attempted-and-unaccomplished task is a 0, not a missing row. 4 of the leaks are on DOM, so removing them helps the fused arm: the direction that disfavours this project's own caution. \textbf{Caution:} The WA cells are unaudited for the same defect. Source: \texttt{leakage\_sensitivity.json}.}
\label{tab:t28}
\footnotesize
\setlength{\tabcolsep}{3pt}
\begin{tabular}{@{}
  >{\raggedright\arraybackslash}p{(\linewidth - 12\tabcolsep) * \real{0.1429}}
  >{\raggedright\arraybackslash}p{(\linewidth - 12\tabcolsep) * \real{0.1429}}
  >{\raggedright\arraybackslash}p{(\linewidth - 12\tabcolsep) * \real{0.1429}}
  >{\raggedright\arraybackslash}p{(\linewidth - 12\tabcolsep) * \real{0.1429}}
  >{\raggedright\arraybackslash}p{(\linewidth - 12\tabcolsep) * \real{0.1429}}
  >{\raggedright\arraybackslash}p{(\linewidth - 12\tabcolsep) * \real{0.1429}}
  >{\raggedright\arraybackslash}p{(\linewidth - 12\tabcolsep) * \real{0.1429}}@{}}
\toprule
\begin{minipage}[b]{\linewidth}\raggedright
cell
\end{minipage} & \begin{minipage}[b]{\linewidth}\raggedright
contrast
\end{minipage} & \begin{minipage}[b]{\linewidth}\raggedright
before
\end{minipage} & \begin{minipage}[b]{\linewidth}\raggedright
95\% CI
\end{minipage} & \begin{minipage}[b]{\linewidth}\raggedright
after
\end{minipage} & \begin{minipage}[b]{\linewidth}\raggedright
95\% CI
\end{minipage} & \begin{minipage}[b]{\linewidth}\raggedright
verdict
\end{minipage} \\
\midrule
red·B0 & SoM − Vision & +7.39 & {[}+2.46, +12.32{]} & +7.88 & {[}+2.96, +12.81{]} & unchanged \\
red·B0 & SoM − DOM & +0.49 & {[}-3.94, +4.93{]} & +0.99 & {[}-3.45, +5.42{]} & unchanged \\
red·B1 & SoM − Vision & +4.93 & {[}+1.48, +8.87{]} & +4.43 & {[}+0.99, +7.88{]} & unchanged \\
red·B1 & SoM − DOM & +1.48 & {[}-1.48, +4.43{]} & +0.99 & {[}-1.48, +3.94{]} & unchanged \\
red·B2 & SoM − Vision & -0.99 & {[}-3.45, +1.48{]} & -0.99 & {[}-3.45, +1.48{]} & unchanged \\
red·B2 & SoM − DOM & -2.96 & {[}-5.91, -0.49{]} & -1.48 & {[}-3.45, +0.49{]} & \textbf{flips} \\
\bottomrule
\end{tabular}
\end{table*}

\begin{table*}[tp]
\centering
\caption{Earned versus leaked successes. Which successes were earned. \texttt{\#sidebar\ \textgreater{}\ section\ \textgreater{}\ ul} is read by 9 VWA reddit tasks; \texttt{require\_reset} is a no-op on reddit so subscriptions accumulate. LEAKED = scored success by an episode that never visited the required forum. On VWA: 6 leaked, 68 earned; Table~\ref{tab:t28} recomputes every contrast with the leaked ones zeroed. WebArena audited 2026-08-03 (first time): 50 scored episodes over its 5 sidebar tasks, 0 leaked. \textbf{Caution:} That zero is a \emph{lower bound}, not a clearance: the test asks whether the episode reached the forum, and an episode can arrive at a forum an earlier one subscribed to, read \texttt{Unsubscribe}, and finish without acting. One such case is hand-confirmed (\texttt{B1}/DOM task 597) and scores \texttt{earned} here; a text heuristic for the pattern was tried and rejected because model self-report cannot separate deliberating from acting. Source: \texttt{reddit\_sidebar\_leakage\_audit\_with\_wa.json}.}
\label{tab:t40}
\footnotesize
\setlength{\tabcolsep}{3pt}
\begin{tabular}{@{}lllll@{}}
\toprule
benchmark & cell · mode & scored successes & of which LEAKED & share \\
\midrule
VWA & B0 · DOM & 5 & 1 & 20.0\% \\
VWA & B0 · SoM-image & 2 & 0 & 0.0\% \\
VWA & B0 · DOM+\allowbreak{}sprompt & 3 & 0 & 0.0\% \\
VWA & B0 · DOM+\allowbreak{}stext & 3 & 0 & 0.0\% \\
VWA & B0 · SoM & 3 & 0 & 0.0\% \\
VWA & B0 · Vision & 2 & 1 & 50.0\% \\
VWA & B1 · DOM & 2 & 0 & 0.0\% \\
VWA & B1 · SoM-image & 4 & 0 & 0.0\% \\
VWA & B1 · DOM+\allowbreak{}sprompt & 2 & 0 & 0.0\% \\
VWA & B1 · DOM+\allowbreak{}stext & 4 & 0 & 0.0\% \\
VWA & B1 · SoM & 3 & 1 & 33.3\% \\
VWA & B2 · DOM & 3 & 3 & 100.0\% \\
VWA & B2 · SoM & 1 & 0 & 0.0\% \\
WA & B0 · DOM & 3 & 0 & 0.0\% \\
WA & B0 · SoM-image & 5 & 0 & 0.0\% \\
WA & B0 · DOM+\allowbreak{}stext & 5 & 0 & 0.0\% \\
WA & B0 · SoM & 3 & 0 & 0.0\% \\
WA & B0 · Vision & 3 & 0 & 0.0\% \\
WA & B1 · DOM & 4 & 0 & 0.0\% \\
WA & B1 · SoM-image & 5 & 0 & 0.0\% \\
WA & B1 · DOM+\allowbreak{}stext & 5 & 0 & 0.0\% \\
WA & B1 · SoM & 3 & 0 & 0.0\% \\
WA & B1 · Vision & 1 & 0 & 0.0\% \\
\bottomrule
\end{tabular}
\end{table*}

\section{Label definitions, nesting, and identifiability}\label{supporting-tables}

Every number in these tables is quoted in the body section that references it. They are placed here because the body has an eight-page limit and these tables document the results rather than making the argument.

\subsection{Observation modes}\label{observation-modes}

The grid \S\ref{sec:setup} refers to. The last column is the modality split Appendix~\ref{screenshot-tier} uses. It is not a cost split, for the reason \S\ref{sec:setup} gives: the image-bearing tier holds both the dearest mode and the cheapest.

\begin{table*}[t]
\centering
\caption{The six observation modes (\S\ref{sec:setup}). The three P-modes keep either the mark legend, the SoM prompt, or both, while removing the per-step screenshot.}
\small
\setlength{\tabcolsep}{4pt}
\begin{tabular}{@{}llll@{}}
\toprule
mode & text payload & prompt family & annotated screenshot \\
\midrule
DOM & accessibility tree & DOM & no \\
SoM & mark legend & SoM & \textbf{yes} \\
Vision & none & vision & \textbf{yes} (unannotated) \\
P-text & mark legend & DOM & no \\
P-prompt & accessibility tree & SoM & no \\
P-SoM & mark legend & SoM & no \\
\bottomrule
\end{tabular}
\end{table*}

\subsection{Where the which-mode label disagrees with measured cost}\label{where-the-which-mode-label-disagrees-with-measured-cost}

Per-cell detail for the 12.5--54.6\% range quoted in \S\ref{sec:lowerbound}, and for the claim that the exact-tie case never occurs.

\begin{table*}[t]
\centering
\caption{Per-cell detail behind \S\ref{sec:lowerbound}. The last column counts labelled tasks on which the fixed priority list selected a mode strictly more expensive than another mode that also succeeded on that task. The exact-tie case, where the list order is the only tiebreaker, occurs in zero rows in every cell.}
\small
\setlength{\tabcolsep}{4pt}
\begin{tabular}{@{}llll@{}}
\toprule
cell & labels & multi-success & list picked a strictly pricier mode \\
\midrule
classifieds · B0 & 97 & 68 (70.1\%) & \textbf{53 (54.6\%)} \\
reddit · B0 & 53 & 36 (67.9\%) & \textbf{23 (43.4\%)} \\
classifieds · B1 & 55 & 29 (52.7\%) & \textbf{26 (47.3\%)} \\
reddit · B1 & 24 & 17 (70.8\%) & \textbf{9 (37.5\%)} \\
classifieds · B2 & 16 & 4 (25.0\%) & \textbf{2 (12.5\%)} \\
reddit · B2 & 15 & 3 (20.0\%) & \textbf{2 (13.3\%)} \\
\bottomrule
\end{tabular}
\end{table*}

\subsection{Effect of nesting the operating-point selection}\label{effect-of-nesting-the-operating-point-selection}

Per-cell detail for the −0.99pp to +1.34pp range quoted in \S\ref{sec:lowerbound}, and for the observation that nesting moves the result in both directions.

\begin{table*}[t]
\centering
\caption{Per-cell detail behind \S\ref{sec:lowerbound}. The naive column selects the threshold and the strongest and cheapest modes from whole-cell outcomes; the nested column re-derives all three inside each outer fold. Nesting moves the result in both directions.}
\small
\setlength{\tabcolsep}{4pt}
\begin{tabular}{@{}llll@{}}
\toprule
cell & naive nesting SR & fully nested SR & Δ \\
\midrule
classifieds · B0 & 25.45\% & 26.79\% & \textbf{+1.34pp} \\
classifieds · B1 & 13.84\% & 14.29\% & +0.45pp \\
classifieds · B2 & 1.34\% & 1.34\% & ±0.00pp \\
reddit · B0 & 13.79\% & 12.81\% & \textbf{−0.99pp} \\
reddit · B1 & 6.40\% & 5.91\% & −0.49pp \\
reddit · B2 & 3.94\% & 3.94\% & ±0.00pp \\
\bottomrule
\end{tabular}
\end{table*}

\subsection{Identifiability of pooled which-mode labels}\label{identifiability-of-pooled-which-mode-labels}

Per-site detail for the conflict rates and modal-agreement figures quoted in Appendix~\ref{pooling-across-backbones}.

\begin{table*}[t]
\centering
\caption{Per-site detail behind Appendix~\ref{pooling-across-backbones}. A conflict is one task on which two cells recorded different oracle modes. In-sample modal agreement is the accuracy of emitting the modal label per task, over all pooled labelled rows (168 on classifieds, 92 on reddit), a row being one (task, backbone) pair on which that backbone solved the task.}
\small
\setlength{\tabcolsep}{4pt}
\begin{tabular}{@{}
  >{\raggedright\arraybackslash}p{(\linewidth - 12\tabcolsep) * \real{0.1429}}
  >{\raggedright\arraybackslash}p{(\linewidth - 12\tabcolsep) * \real{0.1429}}
  >{\raggedright\arraybackslash}p{(\linewidth - 12\tabcolsep) * \real{0.1429}}
  >{\raggedright\arraybackslash}p{(\linewidth - 12\tabcolsep) * \real{0.1429}}
  >{\raggedright\arraybackslash}p{(\linewidth - 12\tabcolsep) * \real{0.1429}}
  >{\raggedright\arraybackslash}p{(\linewidth - 12\tabcolsep) * \real{0.1429}}
  >{\raggedright\arraybackslash}p{(\linewidth - 12\tabcolsep) * \real{0.1429}}@{}}
\toprule
\begin{minipage}[b]{\linewidth}\raggedright
site
\end{minipage} & \begin{minipage}[b]{\linewidth}\raggedright
tasks labelled in ≥2 cells
\end{minipage} & \begin{minipage}[b]{\linewidth}\raggedright
conflicting
\end{minipage} & \begin{minipage}[b]{\linewidth}\raggedright
conflict rate
\end{minipage} & \begin{minipage}[b]{\linewidth}\raggedright
in-sample modal agreement
\end{minipage} & \begin{minipage}[b]{\linewidth}\raggedright
same, on shared tasks only
\end{minipage} & \begin{minipage}[b]{\linewidth}\raggedright
same, exact-vector grouping
\end{minipage} \\
\midrule
classifieds & 54 & 31 & \textbf{57.4\%} & \textbf{79.2\%} & 70.3\% (n=118) & 83.9\% \\
reddit & 25 & 14 & \textbf{56.0\%} & \textbf{83.7\%} & 74.1\% (n=58) & 89.1\% \\
\bottomrule
\end{tabular}
\end{table*}

\emph{We call it agreement and not a Bayes ceiling because it is a resubstitution estimate: it scores the same rows it took the modal label from, so a task labelled by only one backbone is correct by construction. That describes 50 of 168 classifieds rows (29.8\%) and 34 of 92 reddit rows (37.0\%); restricted to tasks two or more backbones label, agreement falls to 70.3\% and 74.1\%. An out-of-sample bound would need leave-one-backbone-out prediction or a shrinkage estimator, and would be lower still. Every number here is therefore an optimistic bound on what a pooled classifier could reach, which is the direction the argument needs.}

\emph{The last column groups by the exact feature vector instead of by task. Rows of one task are not always identical, because three of the five numeric features are read from that backbone's own step-0 observation, so they differ somewhere on 31.5\% of shared classifieds tasks and 80.0\% of shared reddit tasks. We report but do not use that grouping: it leaves 74 of 117 classifieds groups and 69 of 78 reddit groups with a single member, covering 44\% and 75\% of rows, so it is even more inflated than the headline, and a router serving one backbone could not recover backbone identity from that jitter anyway.}

\subsection{The screenshot-modality tier}\label{the-screenshot-modality-tier}

Per-site detail for Appendix~\ref{screenshot-tier}, on the same pooled labelled rows and the same grouping as Appendix~\ref{identifiability-of-pooled-which-mode-labels}. The agreement column is the exception: it is defined only on tasks labelled by two or more backbones, since agreement needs two labels to compare.

\begin{table*}[t]
\centering
\caption{Per-site detail behind Appendix~\ref{screenshot-tier}, over the same solve events. Columns two and three carry the same resubstitution caveat as Appendix~\ref{identifiability-of-pooled-which-mode-labels}, and column three additionally rises for an arithmetic reason: merging six classes into two can only increase a modal share. The claim that backbones agree about the screenshot therefore rests on the last two columns, which measure agreement between two backbones' labels directly rather than against a modal label. Under Appendix~\ref{identifiability-of-pooled-which-mode-labels}'s exact-vector grouping the tier figures are 92.3\% and 97.8\%. No classifier is fitted to this target anywhere in the paper.}
\small
\setlength{\tabcolsep}{4pt}
\begin{tabular}{@{}
  >{\raggedright\arraybackslash}p{(\linewidth - 8\tabcolsep) * \real{0.2000}}
  >{\raggedright\arraybackslash}p{(\linewidth - 8\tabcolsep) * \real{0.2000}}
  >{\raggedright\arraybackslash}p{(\linewidth - 8\tabcolsep) * \real{0.2000}}
  >{\raggedright\arraybackslash}p{(\linewidth - 8\tabcolsep) * \real{0.2000}}
  >{\raggedright\arraybackslash}p{(\linewidth - 8\tabcolsep) * \real{0.2000}}@{}}
\toprule
\begin{minipage}[b]{\linewidth}\raggedright
site
\end{minipage} & \begin{minipage}[b]{\linewidth}\raggedright
which-mode modal agreement
\end{minipage} & \begin{minipage}[b]{\linewidth}\raggedright
tier modal agreement
\end{minipage} & \begin{minipage}[b]{\linewidth}\raggedright
tier agreement across backbones
\end{minipage} & \begin{minipage}[b]{\linewidth}\raggedright
six-way agreement across backbones
\end{minipage} \\
\midrule
classifieds & 79.2\% & \textbf{89.9\%} & \textbf{68.5\%} & 42.6\% \\
reddit & 83.7\% & \textbf{96.7\%} & \textbf{88.0\%} & 44.0\% \\
\bottomrule
\end{tabular}
\end{table*}

\section{Derivations for the four relabelling routes}\label{derivations-for-the-four-relabelling-routes}

\S\ref{sec:lowerbound} and \S\ref{sec:gap} state the outcome of each route. This appendix gives the derivation.

\subsection{Continuous labels}\label{continuous-labels}

The cleanest fix would be to regress on a graded quality signal rather than classify a discrete winner: partial credit turns every episode into a training example regardless of whether it succeeded. VisualWebArena does not provide one. Across the 7,686 scored episodes of our 36 landed conditions the evaluator emits exactly two values, 0.0 and 1.0, in a 7,041 / 645 split, as do the 36 protocol-excluded episodes we do not score. This is a property of the benchmark's evaluation design rather than of our pipeline, and it forecloses the route entirely.

\subsection{Coarser classes}\label{coarser-classes}

\S\ref{sec:lowerbound}'s obstruction is that four of six cells have fewer than ten labelled rows in more than one class. Merging classes does not add rows. The binary collapse of the screenshot tier (Appendix~\ref{screenshot-tier}) does help, but for a different reason, having to do with agreement across backbones rather than with class count.

\subsection{Pooling across backbones}\label{pooling-across-backbones}

Six cells at 15--97 labels become 260 pooled examples and every class clears the minimum-count filter, so supply is solved. Identifiability is not. The features carry no model identity, so two backbones facing the same task on the same site produce near-identical feature vectors, and where they are identical and the oracle labels differ, a classifier is being asked to emit two different answers for one input. The figure reported in Appendix~\ref{identifiability-of-pooled-which-mode-labels} is the accuracy of emitting the modal label per group, scored on the rows the label came from. It is an in-sample bound on what a pooled classifier could reach, not a Bayes ceiling, and Appendix~\ref{identifiability-of-pooled-which-mode-labels} gives the resubstitution caveat.

They are only near-identical, not identical, because \texttt{dom\_\allowbreak{}complexity}, \texttt{text\_\allowbreak{}length} and \texttt{tokens\_\allowbreak{}input\_\allowbreak{}text} are read from the backbone's own step-0 observation rather than from the task config. On 31.5\% of shared classifieds tasks and 80.0\% of shared reddit tasks the rows therefore differ somewhere. Grouping by the exact vector rather than by task raises the figure (Appendix~\ref{identifiability-of-pooled-which-mode-labels}, last column), and we report but do not adopt that number: it leaves most groups with one member, and a group of one is scored perfectly whatever the labels do, so it inflates with feature sparsity rather than tracking identifiability. Every version of the number is an optimistic bound and all of them are far below what a deployable which-mode router needs, which is the only thing the argument turns on.

\subsection{Screenshot tier}\label{screenshot-tier}

The tier label is derived from the same solve events as the which-mode label, by mapping each oracle mode to image-bearing (SoM, Vision) or text-only (DOM, P-text, P-prompt, P-SoM). No new episodes are involved, which is why the figure rises without a single new solve event. Part of that rise is arithmetic (two classes admit a larger modal share than six), so the agreement columns rather than the modal-agreement columns carry the claim. The tier is defined only on tasks that some mode solved, so its denominator is the solved set and not the full task universe. Appendix~\ref{the-screenshot-modality-tier}'s modal-agreement columns run over the whole pooled labelled set, as in Appendix~\ref{identifiability-of-pooled-which-mode-labels}; the two agreement columns are restricted to tasks labelled by two or more backbones, that being the only set on which cross-backbone agreement is defined.

\section{The reddit · B2 saving in detail}\label{the-reddit-b2-saving-in-detail}

\S\ref{sec:lowerbound} reports that the one cell whose cost saving survives Holm correction has an AUROC below chance. The mechanism is tail enrichment rather than a globally ordered score.

Reddit · B2 sends 192 of 203 tasks (95\%) to the cheap mode with no accuracy loss, which is unsurprising in a cell where only 7.4\% of tasks are solvable at all: almost nothing in that 95\% was going to succeed under any mode. The policy therefore differs from the free always-cheapest policy by five percent of the allocation. The 11 tasks it holds back for the strong mode carry four successes that the fixed policy does not collect, against four collected by the fixed policy overall. The permutation null detects that enrichment. A globally ordered score is not required to produce it, which is why the cell's AUROC of 0.483, below both chance and its own best single covariate at 0.711, is consistent with a real saving.

Two properties of that test are worth recording. It runs 10,000 draws and reports the plus-one Monte Carlo estimator (k+1)/(B+1), whose floor is therefore 1.0 × 10⁻⁴, two orders below the tightest Holm threshold of 8.3 × 10⁻³. That matters because at the 200 draws we first used, the floor was 1/201 = 4.98 × 10⁻³, this cell reported exactly it, and whether it could clear the threshold at all was a function of the draw count rather than of the data; at 10,000 draws four of the draws match or beat the observed saving, so p = 5.0 × 10⁻⁴ is measured. Second, the quantity tested is the saving at an operating point selected against whole-cell outcomes, which is not the nested policy of Table 6. Null and observation select that point the same way, so the null absorbs the selection optimism and the comparison is fair, but the point is not one a deployment could occupy, and \S\ref{sec:lowerbound}'s conclusion rests on the nested numbers rather than on this test.

\subsection{Supply and trainability under both label definitions}\label{supply-and-trainability-under-both-label-definitions}

\S\ref{sec:lowerbound} keeps the prior-order label and reports the measured-cost rule as a sensitivity. Supply is identical under both by construction, since each labels exactly the tasks some mode solved, so only the class distribution and through it trainability can move. The relabelled column equals the ``order picked a strictly pricier mode'' column of A.2 exactly, which is the consistency check: a label moves if and only if the order's pick was not the measured cheapest.

\begin{table*}[t]
\centering
\caption{Trainability under the two label definitions. ``Surviving'' counts classes clearing ten training rows in a five-fold split; \textbf{no} marks a cell with fewer than two, where no classifier exists. Four of six cells are untrainable under the reported label and five of six under the measured-cost alternative, so the supply argument does not depend on the choice.}
\small
\setlength{\tabcolsep}{4pt}
\begin{tabular}{@{}
  >{\raggedright\arraybackslash}p{(\linewidth - 8\tabcolsep) * \real{0.2000}}
  >{\raggedright\arraybackslash}p{(\linewidth - 8\tabcolsep) * \real{0.2000}}
  >{\raggedright\arraybackslash}p{(\linewidth - 8\tabcolsep) * \real{0.2000}}
  >{\raggedright\arraybackslash}p{(\linewidth - 8\tabcolsep) * \real{0.2000}}
  >{\raggedright\arraybackslash}p{(\linewidth - 8\tabcolsep) * \real{0.2000}}@{}}
\toprule
\begin{minipage}[b]{\linewidth}\raggedright
cell
\end{minipage} & \begin{minipage}[b]{\linewidth}\raggedright
labels
\end{minipage} & \begin{minipage}[b]{\linewidth}\raggedright
relabelled
\end{minipage} & \begin{minipage}[b]{\linewidth}\raggedright
prior order: surviving classes
\end{minipage} & \begin{minipage}[b]{\linewidth}\raggedright
measured cost: surviving classes
\end{minipage} \\
\midrule
classifieds · B0 & 97 & 53 & 3 (DOM, P-prompt, SoM) & 2 (SoM, Vision) \\
reddit · B0 & 53 & 23 & 1 (DOM) (\textbf{no}) & 1 (P-text) (\textbf{no}) \\
classifieds · B1 & 55 & 26 & 2 (DOM, SoM) & 1 (Vision) (\textbf{no}) \\
reddit · B1 & 24 & 9 & 0 (\textbf{no}) & 0 (\textbf{no}) \\
classifieds · B2 & 16 & 2 & 0 (\textbf{no}) & 0 (\textbf{no}) \\
reddit · B2 & 15 & 2 & 0 (\textbf{no}) & 0 (\textbf{no}) \\
\bottomrule
\end{tabular}
\end{table*}

The single cell that changes, classifieds · B1, loses a class rather than gaining one: the prior-order label keeps DOM and SoM above the threshold, the measured-cost label concentrates enough of those rows onto Vision that only Vision survives. The alternative definition therefore strengthens the negative result, which is a reason to report it and not a reason to adopt it.

\subsection{The best-success mode is not stable across folds}\label{the-best-success-mode-is-not-stable-across-folds}

The nested design of \S\ref{sec:lowerbound} re-selects the best-success mode inside every outer fold, which exposes something the whole-cell version conceals. In reddit · B0 the five outer folds select DOM, DOM, SoM, SoM, DOM. A pipeline that picks one best mode from all realised outcomes is therefore not merely optimistic about its threshold; it reports a mode choice that its own resampling does not reproduce.

\end{document}